\documentclass[11pt]{article}
\pdfoutput=1

\usepackage{mathrsfs}
\usepackage{amsmath,amssymb,amsfonts}
\usepackage{bm}
\usepackage{natbib}
\usepackage[usenames]{color}
\usepackage{adjustbox}
\usepackage{float}
\usepackage{latexsym}
\usepackage[affil-it]{authblk}
\usepackage{enumerate}
\usepackage{mathabx}
\usepackage{multirow} 
\usepackage{enumitem}
\usepackage{algorithm}
\usepackage{algorithmic}
\usepackage[dvipsnames]{xcolor}
\usepackage{graphicx}
\graphicspath{{./figures/}} 
\usepackage[colorlinks,
linkcolor=blue,
anchorcolor=blue,
citecolor=blue
]{hyperref}

\renewcommand{\thefootnote}{\alph{footnote}}
\usepackage[margins]{trackchanges}

\usepackage{mylatexstyle}

\usepackage{chngcntr}
\counterwithout{equation}{subsection}
\counterwithout{equation}{section} 

\usepackage{setspace}
\usepackage[left=1in, right=1in, top=1in, bottom=1in]{geometry}

\usepackage{amsthm}

\usepackage{xcolor}

\definecolor{pinegreen}{rgb}{0.0, 0.47, 0.44}

\newcommand\syb[1]{{\color{blue} #1}} 

\newcommand{\bel}{\begin{eqnarray}\label}
	\newcommand{\eel}{\end{eqnarray}}
\newcommand{\bes}{\begin{eqnarray*}}
	\newcommand{\ees}{\end{eqnarray*}}
\newcommand{\bei}{\begin{itemize}}
	\newcommand{\beiftnt}{\begin{itemize}\footnotesize}
		\newcommand{\eei}{\end{itemize}}

	\newcommand{\convd}{\stackrel{{\rm D}}{\longrightarrow}}

	\def\benu{\begin{enumerate}}
		\def\eenu{\end{enumerate}}

	\def\argmin{\mathop{\rm arg\, min}}
	\def\real{{\mathbb{R}}}

	\def\E{{\mathbb{E}}}

	\def\complex{\mathop{{\rm I}\kern-.58em\hbox{\rm C}}\nolimits}

	\def\diag{\hbox{\rm diag}}

	\def\sgn{\hbox{\rm sgn}}

	\def\Var{\hbox{\rm Var}}

	\def\mathbold{\boldsymbol} 
	
	\def\ba{\mathbold{a}}
	
	\def\bA{\mathbold{A}}\def\calA{{\cal A}}
	\def\hbA{{\widehat{\bA}}}

	\def\bb{\mathbold{b}}
	
	\def\bB{\mathbold{B}}\def\calB{{\cal B}}
	\def\hbB{{\widehat{\bB}}}
	
	\def\bfc{\mathbold{c}}
	
	\def\bfC{\mathbold{C}}\def\calC{{\cal C}}
	\def\hbC{{\widehat{\bfC}}}\def\tbC{{\widetilde{\bfC}}}
	
	\def\bc{\mathbold{c}}
	
	\def\bC{\mathbold{C}}\def\calC{{\cal C}}
	\def\hbC{{\widehat{\bC}}}\def\tbC{{\widetilde{\bC}}}

	\def\bD{\mathbold{D}}\def\calD{{\cal D}}

	\def\bfe{\mathbold{e}}
	\def\tbe{{\widetilde{\bfe}}}
	\def\bE{\mathbold{E}}
	\def\tbE{{\widetilde{\bE}}}
	
	\def\bff{\mathbold{f}}
	
	\def\bF{\mathbold{F}}

	\def\bg{\mathbold{g}}
	
	\def\bG{\mathbold{G}}
	\def\tbG{{\widetilde{\bG}}}
	
	\def\bh{\mathbold{h}}
	
	\def\bH{\mathbold{H}}\def\calH{{\cal H}}

	\def\bi{\mathbold{i}}
	
	\def\bI{\mathbold{I}}

	\def\bj{\mathbold{j}}
	
	\def\bJ{\mathbold{J}}

	\def\bk{\mathbold{k}}
	
	\def\bK{\mathbold{K}}

	\def\bell{\mathbold{\ell}}
	
	\def\bL{\mathbold{L}}
	\def\calL{{\cal L}}
	
	\def\bm{\mathbold{m}}
	
	\def\bM{\mathbold{M}}

	\def\bn{\mathbold{n}}
	
	\def\bN{\mathbold{N}}
	\def\calN{{\cal N}}
	
	\def\bo{\mathbold{o}}
	
	\def\bO{\mathbold{O}}

	\def\bp{\mathbold{p}}
	
	\def\bP{\mathbold{P}}
	\def\calP{{\cal P}}
	
	\def\bq{\mathbold{q}}
	
	\def\bQ{\mathbold{Q}}
	\def\calQ{{\cal Q}}
	
	\def\br{\mathbold{r}}
	
	\def\bR{\mathbold{R}}
	\def\calR{{\cal R}}
	
	\def\bs{\mathbold{s}}
	
	\def\bS{\mathbold{S}}\def\hbS{{\widehat{\bS}}}\def\tbS{{\widetilde{\bS}}}
	\def\calS{{\cal S}}
	
	\def\bt{\mathbold{t}}
	
	\def\bT{\mathbold{T}}

	\def\bu{\mathbold{u}}
	
	\def\bU{\mathbold{U}}

	\def\bv{\mathbold{v}}
	
	\def\bV{\mathbold{V}}

	\def\bw{\mathbold{w}}
	
	\def\bW{\mathbold{W}}\def\tbW{{\widetilde{\bW}}}

	\def\bx{\mathbold{x}}
	
	\def\bX{\mathbold{X}}\def\hbX{{\widehat{\bX}}}

	\def\by{\mathbold{y}}
	
	\def\bY{\mathbold{Y}}

	\def\bz{\mathbold{z}}
	
	\def\bZ{\mathbold{Z}}

	\def\balpha{\mathbold{\alpha}}

	\def\bbeta{\mathbold{\beta}}

	\def\bgamma{\mathbold{\gamma}}

	\def\bGamma{\mathbold{\Gamma}}

	\def\bdelta{\mathbold{\delta}}

	\def\bDelta{\mathbold{\Delta}}

	\def\ep{\varepsilon}
	\def\bep{\mathbold{\ep}}

	\def\bzeta{\mathbold{\zeta}}

	\def\bfeta{\mathbold{\eta}}

	\def\btheta{\mathbold{\theta}}
	
	\def\hbtheta{{\widehat{\btheta}}}
	
	\def\bTheta{\mathbold{\Theta}}

	\def\bkappa{\mathbold{\kappa}}

	\def\lam{\lambda}
	\def\blam{\mathbold{\lam}}

	\def\blambda{\mathbold{\lambda}}

	\def\bLambda{\mathbold{\Lambda}}

	\def\bmu{\mathbold{\mu}}

	\def\bnu{\mathbold{\nu}}

	\def\bxi{\mathbold{\xi}}

	\def\bXi{\mathbold{\Xi}}

	\def\bpi{\mathbold{\pi}}

	\def\bPi{\mathbold{\Pi}}\def\hbPi{{\widehat{\bPi}}}

	\def\brho{\mathbold{\rho}}

	\def\bsigma{\mathbold{\sigma}}\def\hsigma{\widehat{\sigma}}
	\def\tsigma{\widetilde{\sigma}}

	\def\bSigma{\mathbold{\Sigma}}
	\def\tbSigma{{\widetilde{\bSigma}}}

	\def\btau{\mathbold{\tau}}

	\def\bphi{\mathbold{\phi}}

	\def\bPhi{\mathbold{\Phi}}\def\hbPhi{{\widehat{\bPhi}}}
	\def\tbPhi{{\widetilde{\bPhi}}}

	\def\bchi{\mathbold{\chi}}

	\def\bpsi{\mathbold{\psi}}

	\def\bPsi{\mathbold{\Psi}}

	\def\bomega{\mathbold{\omega}}

	\def\bOmega{\mathbold{\Omega}}

	\def\calD{{\cal D}}

	\def\calR{{\mathcal R}}
	\def\calB{{\mathcal B}}

	\def\bbR{{\mathbb{R}}}
	
	\def\bbE{{\mathbb{E}}}
	
	\def\be{{\boldsymbol{e}}}
	\def\calD{{\cal D}}

	\def\tbS{\widetilde{\bS}}

	\def\hsigma{\widehat{\sigma}}
	\def\tsigma{\widetilde{\sigma}}

	\def\bbB{\widebar{\mathbf{B}}}
	\def\bbA{\widebar{\mathbf{A}}}

	\def\bba{\widebar{\mathbf{a}}}

	\def\argmin{\operatorname{argmin} \displaylimits}
	\def\bm{\boldsymbol{\mu}}
	\def\bq{\mathbf{q}}

	\def\mR{\mathcal{R}}

	\def\E{\mathbb{E}}

	\def\T{\top}

	\def\tr{\text{tr}}

	\def\bF{\mathbold{F}}
	\def\tvec{\text{vec}}
	
	\def\calD{{\cal D}}

	\def\bbR{{\mathbb{R}}}
	
	\def\bbE{{\mathbb{E}}}

	\def\bve{\mathbold{\varepsilon}}

\usepackage{xcolor}
\definecolor{MyYellow}{HTML}{FF9C3B}
\definecolor{MyPurple}{HTML}{8286CC}
\definecolor{MyBlack}{HTML}{000000}

\definecolor{MyRed}{HTML}{CB5F5F}
\definecolor{MyGreen}{HTML}{74B156}

\definecolor{red2}{HTML}{db474d}
\definecolor{bluep}{HTML}{535ea9}

\newcommand\redbf[1]{{\bf {\color{MyRed} #1}}}
\newcommand\blackbf[1]{{\bf {\color{MyBlack} #1}}}

\newcommand\pinbf[1]{{\bf {\color{bluep} #1}}}

\usepackage{xr}
\begin{document}
\title{\LARGE Sparsity-Induced Global Matrix Autoregressive Model with Auxiliary Network Data}


\author[a]{\small Sanyou Wu}
\author[, b]{\small Dan Yang  \footnote{Corresponding author \\ Email address:  sanyouwu@connect.hku.hk (Sanyou Wu), dyanghku@hku.hk (Dan Yang), yanxuj@hku.hk (Yan Xu), lfeng@hku.hk (Long Feng)}}
\author[b]{\small Yan Xu}
\author[a]{\small Long Feng}

\affil[a]{\footnotesize Department of Statistics and Actuarial Science, The University of Hong Kong, Pokfulam Road, Hong Kong}
\affil[b]{\footnotesize Faculty of Business and Economics, The University of Hong Kong, Pokfulam Road, Hong Kong}

\date{}

\maketitle
\providecommand{\keywords}[1]{\textbf{\textit{Keywords---}} #1}



\begin{abstract}
Jointly modeling and forecasting economic and financial variables across a large set of countries has long been a significant challenge. Two primary approaches have been utilized to address this issue: the vector autoregressive model with exogenous variables (VARX) and the matrix autoregression (MAR). 
The VARX model captures domestic dependencies, but treats variables exogenous to represent global factors driven by international trade. 
In contrast, the MAR model simultaneously considers variables from multiple countries but ignores the trade network. In this paper, we propose an extension of the MAR model that achieves these two aims at once, i.e., studying both international dependencies and the impact of the trade network on the global economy. Additionally, we introduce a sparse component to the model to differentiate between systematic and idiosyncratic cross-predictability. To estimate the model parameters, we propose both a likelihood estimation method and a bias-corrected alternating minimization version. We provide theoretical and empirical analyses of the model's properties, alongside presenting intriguing economic insights derived from our findings.
\end{abstract}

\newpage

\section{Introduction}
Vector autoregression (VAR) has been an extremely powerful analytic tool for macroeconomic studies  \citep{stock2001vector}. However, in cross-country studies, modeling interdependence and connectedness between economies poses a challenge of dimensionality. Given common movements of growth opportunities or monetary policy shocks across countries, and spillovers in the global capital markets, estimation of VAR becomes increasingly difficult when the number of state variables is large. In this paper, our proposed answer is a model exploiting the innovative framework of matrix autoregression (MAR). This proposal is naturally motivated by the fact that fundamental macroeconomic variables such as gross domestic product (GDP), inflation, interest rates,  and equity prices possess both a time dimension and a country dimension, thus making these observations matrix-valued time series. 
Moreover, as intercountry connectedness in the real economic activities is largely captured by global trade networks, our paper strives to incorporate crucial information from bilateral trade relationships into a matrix-valued time series framework.

In recent years, matrix-valued time series, which consist of a sequence of matrices, have gained increasing popularity across various research domains even outside economics, such as social science, neuroscience, and many others. 
In empirical asset pricing studies, for instance, two- or three-way sorted Fama-French portfolios contain time series of returns unfolded along the dimension of firm characteristics. In social science, dynamic social networks or email correspondence between employees naturally form matrix time series. In neuroscience, medical images of patients over time, such as 2D computed tomography (CT) scans, can be used to monitor disease progression. Additionally, various types of sequential 2D image data, such as satellite image time series, are important sources of information for ecosystem analysis.

The traditional econometric approach to handle matrix-valued time series is to vectorize each matrix into a long vector for each time point then invoke the VAR models. Numerous efforts have been devoted to tackling the challenge of high dimensionality for vector time series, such as the assumption of sparsity \citep{basu2015regularized}, imposition of banded structure \citep{guo2016high}, factor-augmented VAR (FAVAR) \citep{BernankeBoivinPiotr_2005QJE}, or dynamic vector factor models \citep{lam2011estimation, lam2012factor}.
However, in addition to the issue of high dimensionality, VAR of the vectorized sequence also suffers from low accuracy and high computational cost, and has trouble obtaining economic implications. 
More importantly, such VAR models fail to exploit structural information inherent in the rows and columns of matrix-valued observations, leading to the detrimental consequence of ignoring the intercountry connectedness.

Recent evidence has consistently demonstrated that preserving the matrix or tensor structure (extensions from matrices to multi-dimensional arrays) of original data, particularly in the context of time series, offers substantial benefits. In the literature, \citet{chen2021autoregressive} utilize the matrix form in their proposed MAR model, significantly reducing the total number of parameters. This innovation leads to clear advantages in terms of speed, accuracy, and interpretability over the conventional VAR model. Since then, various extensions of the conventional MAR model have been developed, including reduced-rank MAR \citep{xiao2022reduced}, tensor autoregression \citep{li2021multi,wang2024high}, spatio-temporal MAR \citep{hsu2021matrix}, mixture MAR \citep{wu2023mixture}, matrix ARIMA \citep{tsay2024matrix}, etc. Matrix factor models \citep{wang2019factor,chen2020constrained,gao2023two,he2023one,he2024matrix} and tensor factor models \citep{wang2022high,chen2022factor,chen2023statistical,chang2023modelling} have also been extensively studied to reduce the dimensionality of matrix/tensor time series.

To study intercountry economic relationships, it is natural to consider spatial autoregressive (SAR) models \citep{cliff1973spatial}. In SAR models, response observations are influenced by their neighboring observations based on geographic proximity. In cross-country analysis, the foreign trade volume, as a proxy for close economic dependency, can be more useful than geographic proximity. 
Recent extensions of SAR, such as the Spatial Dynamic Panel Data (SDPD) model \citep{yu2008quasi,yu2012estimation}, have garnered significant attention. The SDPD model, a special form of VAR models, consists of three main terms: contemporaneous (lag-0) spatial effect, traditional dynamic (lag-1) effect, and autoregressive (lag-1) spatial effect. The basic SDPD model assumes that the VAR coefficient matrices related to these three terms are products of unknown scalars and a common prespecified linkage matrix. \citet{dou2016generalized} generalize the SDPD model to allow different scalars for each panel or location.
In the literature on spatial-temporal autoregression, \citet{zhu2019portal} aims to identify influential locations by assuming the coefficient matrix as a product of a predetermined matrix and a diagonal matrix. \citet{zhou2017estimating} investigates models with partially observed data. \citet{gao2019banded,ma2023sparse} explore more flexible models with unknown coefficient matrices for contemporaneous and autoregressive terms.


One commonality among the SDPD and spatial-temporal autoregression models mentioned is the inclusion of the contemporaneous (lag-0) spatial effect induced by the trade network. Another research topic that can incorporate the trade network, network autoregression, only allows the autoregressive (lag-1) spatial or network effect \citep{zhu2017network,chen2023community,zhu2023simultaneous}, but excludes cross-sectional (lag-0) spatial network effect. These models also often incorporate group or community structures.

In the presence of trade network, the SDPD, spatial-temporal autoregression, and network autoregression models have been extensively studied. However, these models are designed for vector time series and are therefore only applicable to scenarios involving a single economic variable across multiple countries. Building separate models for different economic variables, on the other hand, neglects the comovement of intrinsically related economic variables.

Inspired by recent developments in MAR and SDPD models, we propose a sparsity-induced global matrix autoregressive (SIGMAR) model to handle multicountry economic fundamental variables simultaneously with valuable information from their trade network. The SIGMAR model integrates two effects: contemporaneous spatial (lag-0) effect and autoregressive (lag-1) effect. 
By including the first effect, our model can capture the interaction of all economic variables comprehensively, thus distinguishing it from the SDPD and network regression models. For the second effect, the lead-lag relationships of all economic variables across all countries are summarized by a Kronecker product plus a residual matrix, which cannot be further decomposed. The use of the Kronecker product has been previously considered in MAR models. In this context, one coefficient matrix describes the general pattern of lead-lag dynamics of economic variables, while the other describes that of countries. Their element-wise product reveals the Granger causality from one variable $j$ of country $i$ at time $t-1$ to another variable $j'$ of country $i'$ at time $t$. We refer to this Kronecker product as ``systematic predictability." The residual matrix, which is another of our contributions to the literature, captures cross-predictability that is not well approximated by the systematic predictability. We refer to this residual matrix as ``idiosyncratic predictability." In particular, we impose sparsity on this residual matrix, not only for mathematical identification purposes, but also for economic reasons. There is already U.S. evidence suggesting that some variables are not well forecast by others \citep{StockWatson_2003JEL}, and it is conceivable that some countries do not share and transmit common technological or monetary shocks as strongly as others. Thus, the sparser this residual matrix is, the better the systematic predictability matrix approximates the lead-lag dynamics of all variables from all countries.

Notably, unlike SDPD models, we deliberately exclude the autoregressive spatial effect, 
partially because of the structure of the Global Vector Autoregression (GVAR) approach \citep{GLPS_2006, pesaran2004modeling,pesaran2009forecasting}.
The GVAR model, extensively employed by econometricians, essentially excludes the autoregressive spatial effect. 
The GVAR model is termed the ``global" VAR because it utilizes separate VAR models for different countries, with each model comprising a domestic term and a ``global" term. The domestic term is simply the conventional autoregressive lag-1 component, and the global term, also known as the star variable or weighted foreign variables, is defined as the product of the trade network and the contemporaneous domestic response variables. By exploiting information from intercountry trade network, 
the global term captures some effect similar to the motivation of the contemporaneous spatial effect. 
Hence, the SIGMAR model \emph{may bear some resemblance} to a particular extension of the GVAR model from vector time series to matrix time series by combining multiple GVAR models from various countries through the use of the Kronecker product and the addition of sparse components. However, mathematically, there are fundamental differences between SIGMAR and such an extension of GVAR. In the implementation of GVAR, the star variables, once constructed by multiplying the trade network with domestic variables, are considered given and exogenous, i.e., almost independent of the noise term. In contrast, in the SIGMAR model, inspired by SAR models, the contemporaneous spatial network term is correlated with the noise, leading to significant methodological distinctions from GVAR.
The relationship between SIGMAR and GVAR will be further elaborated upon in Section \ref{sec:model}.

The quasi-maximum likelihood estimator (QMLE) is proposed for model estimation and is implemented in two steps. In the first step, the QMLE is obtained by maximizing the  quasi-likelihood function without imposing structural restrictions involving the combination of a Kronecker product and a sparse residual matrix. Subsequently, the Kronecker product and the sparse matrix are recovered through a projection method applied to the intermediate result from the first step. The asymptotic properties of the QMLE are investigated, and the stability and accuracy of the recovery using the projection method are also established. 
Moreover, an alternating minimization algorithm is devised to enhance the efficiency of the estimation. Notably, the standard alternating minimization algorithm may lead to biased estimates due to the correlation between explanatory variables and the error term, a phenomenon common in spatial autoregression. To address this issue, we introduce a novel bias-correction procedure to ensure the consistency of our estimators.

Our real data analysis involves gross domestic output, consumption price index, equity price, and long and short term interest rates from ten OECD developed countries and regions, covering a sample period between 1979Q1 and 2019Q4. In a nutshell, we find that time-series movements of equity price and long rate to be most strongly associated with each of their contemporaneous aggregated counterpart, corroborating the finance literature that stock returns or bond yields can be largely described by a parsimonious static factor model. In contrast, short
rate can be predicted by long rate of the same country, or short rate of other countries, implying a nonsynchronous monetary policy cooperation across countries. In a horse race against various extensions of the conventional VAR models, SIGMAR performs the best and produces the smallest forecast errors. It only fares less favorably compared to individual univariate AR model, which however, fails to offer any useful insights about the structure of the global economy. See Section \ref{sec:real} for more interesting economic findings. 

While we propose the SIGMAR to model and forecast global economic and financial variables, we believe our contribution could be beyond the scope of this paper by offering greater potential for a wide range of applications. The unit of analysis is not limited to countries; they can instead be regions, states, industries, sectors, compartments, and so forth. For instance, we could study the housing market and jointly forecast commercial real estate prices of different types, such as offices, apartments, industrial, and retail properties across various metropolitan areas of the U.S. over time, while the network information is embedded in trends of internal immigration due to job relocation, urbanization, and land-use regulation.  Our framework is also useful to study infectious disease. Taking COVID-19 for example, we could consider all states in the U.S. and multiple variables such as the number of infected individuals, the number of recoveries, the number of hospitalizations, and the number of ventilators required. The network in this case could be represented by the traffic volume between different states. Finally, our framework can also be used to study impulse response functions to conduct policy analysis.


The remainder of the paper is organized as follows: Section \ref{sec:model} introduces the model, while Section \ref{sec:mle} presents the model estimation procedure. Section \ref{sec:theory} establishes the theoretical properties of the proposed estimation procedure, and Section \ref{sec:alternating} provides an alternating minimization algorithm with bias correction to further improve the estimation. We conduct simulation studies in Section \ref{sec:simulation} and apply the model to a real data example in the context of the global economy in Section \ref{sec:real}. All technical proofs are provided in the appendix.

\noindent 
{\bf Notations: } For a vector $\bv=(v_1,
\ldots,v_p)^\T$, $\|\bv\|_q=\sum_{1\le j\le p} (|v_j|^q)^{1/q}$ is the $\ell_q$ norm, $\|\bv\|_0$ the number of nonzero entries.  For a matrix $\bM$, we denote
$\|\bM\|_F$ as the Frobenius norm,  $\|\bM\|_1$ as the maximum absolute column sum, $\|\bM\|_{\infty}$ as the maximum absolute row sum, and $\tvec(\bM)$ as the vectorization of $\bM$. 
For any square matrix $\bU$, we use $\rho(\bU)$ to denote its spectral radius, which is defined as the maximum modulus of the (complex) eigenvalues of $\bU$. In addition, we define $\mathbf{1}_{n}$ as the column vector with all entries being 1, use $\bI_n$ to denote  an identity matrix of dimension $n\times n$, and $\otimes$ to denote the Kronecker product.

\section{The SIGMAR models}
\label{sec:model}
\subsection{Review of MAR and GVAR models}
\label{ssec:model-review}
Suppose there are $n$ countries and the $k$ economic and financial variables for country $i$ at time $t$ are denoted by $\bx_{t,i}\in \mathbb{R}^k$. Then a simple VAR model of order one for the $i$-th country itself takes the form
\bel{eq:model-ivar}
\bx_{t,i} = \mathbf{\Phi}_i \bx_{t-1,i}  + \bve_{t,i},\quad i=1,\ldots, n,
\eel
where $\bPhi_i$ is the AR(1) coefficient matrix of size $k\times k$. Since the coefficient matrix $\mathbf{\Phi}_i$ is distinct for each country, from now on, we refer to Model \eqref{eq:model-ivar} as iVAR, short for individual VAR.

It is clear that iVAR Model \eqref{eq:model-ivar} only captures the dependencies among all the variables within country $i$ and fails to model the cross-country impact caused by the interlinked global economy. One ad-hoc remedy to take care of the global inter-dependencies is to first stack all the $k$-dimensional vectors from $n$ countries together, which results in a long vector of size $nk$, denoted by $\by_t = (\bx_{t,1}^\T,
\bx_{t,2}^\T,\ldots,\bx_{t,n}^\T)^\T$, and next formulate a large-scale VAR model for $\by_t$, say,
\bel{eq:model-svar}
\by_{t} = \mathbf{\Phi} \by_{t-1}  + \bve_{t},
\eel
where the coefficient matrix $\mathbf{\Phi}$ is of size $kn\times kn$. Model \eqref{eq:model-svar} will be referred to as the stacked VAR (sVAR) for the rest of this article.
Such an ad-hoc remedy inevitably generates a complex and intimidating high-dimensional problem. Given the limited length of the time series, it is unlikely that estimates for $n^2k^2$ parameters can be obtained. Even if these estimations are achievable, interpreting the large coefficient matrix remains highly complicated, as it necessitates explaining the influence of any variable in any country on any other variable in any other country.

In the literature, two streams of approaches aim to address the shortcomings of the country-specific iVAR Model \eqref{eq:model-ivar} and the global sVAR Model \eqref{eq:model-svar}. These approaches have two primary goals: (i) to model the global effect and (ii) to control model complexity.

The first stream is the MAR model, originally proposed by \citet{chen2021autoregressive}. Let $\bX_t$ denote a matrix of size $k \times n$, where the $i$-th column contains all the variables from country $i$, i.e., $\bx_{t,i}$. In other words, $\bX_t = (\bx_{t,1}, \bx_{t,2}, \ldots, \bx_{t,n})$ stores the information for all variables from all countries at time $t$.
The MAR approach models the dynamics of all countries simultaneously, rather than separately. Specifically, the MAR model of order one is given by:
\bel{eq:model-mar} \bX_t = \bA \bX_{t-1} \bB^\T + \bE_t, \eel
where $\bA$ and $\bB$ are two unknown fixed coefficient matrices of sizes $k \times k$ and $n \times n$, respectively. This model preserves the matrix structure of $\bX_t$ and $\bX_{t-1}$, and is accordingly referred to as the matrix autoregressive model. 

Now, let us explore the connections between the MAR Model \eqref{eq:model-mar} and the sVAR Model \eqref{eq:model-svar}. It is seen that vectorizing the MAR Model \eqref{eq:model-mar} leads to an equivalent expression: 
\bel{eq:model-mar-vec} 
\tvec(\bX_t) = \left(\bB \otimes \bA\right) \tvec(\bX_{t-1}) + \tvec(\bE_t), 
\eel 
where $\otimes$ denotes the matrix Kronecker product and $\tvec$ denotes the vectorization operation. Notably, since $\tvec(\bX_t) = \by_t$, the MAR Model \eqref{eq:model-mar} essentially posits that $\by_t$ follows a large-scale VAR(1) process: $\by_{t} = \left(\bB \otimes \bA\right) \by_{t-1} + \bve_{t}$, but with a specially-structured coefficient matrix $\mathbf{\Phi} = \bB \otimes \bA$. This special structure reduces the original number of coefficients from $k^2n^2$ in the sVAR Model \eqref{eq:model-svar} to $k^2 + n^2$ in the MAR Model \eqref{eq:model-mar-vec}. Note that the Kronecker product $\bB \otimes \bA$ is identifiable, and the two matrices $\bA,\bB$ are only identifiable up to scaling and sign changes. For the rest of the paper, it is assumed that $\|\bA\|_F=1$.

Turning to the comparison between MAR Model \eqref{eq:model-mar} and iVAR Model \eqref{eq:model-ivar}. Note that Model \eqref{eq:model-ivar} is also a special case of  Model \eqref{eq:model-svar}, where $\mathbf{\Phi} = \diag(\mathbf{\Phi}_1, \mathbf{\Phi}_2,\ldots, \mathbf{\Phi}_n)$, which has a total number of parameters $k^2n$. From this regard of number of parameters, MAR Model \eqref{eq:model-mar} is more parsimonious. When $\bB$ is an identity matrix, Model \eqref{eq:model-mar} reduces to large-scale VAR(1) with $\mathbf{\Phi} = \diag(\bA,\bA, \ldots, \bA)$, which essentially assumes all country-specific VAR(1) coefficient matrices $\mathbf{\Phi}_i$s are identical. That is, when $\bB$ is an identity matrix, MAR Model \eqref{eq:model-mar} is more restrictive than iVAR Model \eqref{eq:model-ivar}. However, $\bA$ captures the relationship between different variables whilst $\bB$ captures the relationship between different countries, and the latter global relationship is what iVAR Model \eqref{eq:model-ivar} is short of. The MAR Model \eqref{eq:model-mar} can explain how any variable $j$ from any country $i$ can influence any variable $j'$ from any country $i'$ via the term $a_{j'j}b_{i'i}$. 
See \citet{chen2021autoregressive} for more interpretations of the MAR model and \citet{li2021multi} for extension to more terms of $\bB\otimes \bA$ and to higher order tensor autoregressive model.

For global effect, besides the unknown country interaction via $\bB$ in the MAR model, there is more known network information such as foreign trade or capital flow, which when incorporated into the modeling procedure might offer extra insight on the global economy. However, the current MAR model does not take the network information into consideration.

The second stream of GVAR has a longer history, dated back to the pioneering work by \cite{pesaran2004modeling}, which can incorporate the trade network elegantly. The GVAR model keeps the country-specific model still small-scale while adding the additional so-called foreign variables or the star variables $\bx_{t,i}^*\in \mathbb{R}^k$ to encapsulate the global effect. Specifically, the simplest GVAR model has a VARX*(1,0) model specification for country $i$,
\bel{eq:model-igvar}
\bx_{t,i} = \mathbf{\Phi}_i \bx_{t-1,i}  + \mathbf{\Psi}_i\bx_{t,i}^* + \bve_{t,i},
\eel
where variables $\bx_{t,i}$ are domestic while the star variables $\bx_{t,i}^*$ are foreign. 

The star variables are usually constructed by utilizing data on the bilateral trade network, denoted by $\bW\in \mathbb{R}^{n\times n}$. The matrix $\bW$ is used as weights, and hence row-normalized typically. In the context of the global economy, $w_{ij}$ can be defined as the trade between country $i$ and country $j$ divided by the total trade of country $i$ with all its trading partners, where the diagonal entries are zeros, i.e., $w_{ii} = 0$ for all $i$ by default. The weight matrix $\bW$ is treated as known and fixed. In other application scenarios of GVAR, the weight matrix $\bW$ may be constructed as an adjacency matrix in network analysis \citep{zhu2017network,zhu2023simultaneous,ren2024multi} or may represent the spatial relationships of regions \citep{dou2016generalized, gao2019banded}.

The $i$-th row of the weight matrix $\bW$ is the country-specific weights. Given the network weight matrix $\bW$, the $j$-th star variable for country $i$ is usually defined as $x_{t,ji}^* = \sum_{k=1}^n w_{ik}x_{t,jk}$ in the GVAR literature. This definition is a weighted average of the same variable $j$ from all $n$ countries for the same time period, where the weights $w_{ik}$, $k=1,\ldots, n$, sum up to one and measure how much country $k$ influences country $i$.
Such a definition of star variables in the scalar form has an equivalent expression for all countries in a matrix form
\bel{eq:global}
\bX_t^*=\bX_t\bW^\T,
\eel
where $\bX_t^*$ is still of size $k\times n$, and the $i$-th column of $\bX_t^*$ or $\bx_{t,i}^*$ corresponds to the star variables of country $i$.

Consider an example, which will be used in the real data section as well, where there are $k=5$ variables of interest, including Gross Domestic Product (GDP), Consumer Price Index (CPI), Equity Price Index (EQ), short term interest rate (SR), and long term interest rate (LR). Suppose the first columns of 
$\bX_t$ and $\bX_t^*$ represent USA, Model \eqref{eq:global} suggests that the five star variables of the USA can be constructed as  
\bes
\begin{gathered}
\text {\hspace{-0.2cm}USA \hspace{2.3cm} USA \hspace{2.1cm} UK \hspace{3.1cm} CAN} \\
	\left(\begin{array}{c}
		\text {GDP}^* \\
		\text {CPI}^* \\
		\text {EQ}^* \\
            \text {SR}^* \\
		\text {LR}^*
	\end{array}\right)_{t}=w_{11}\left(\begin{array}{c}
		\text {GDP} \\
		\text {CPI} \\
		\text {EQ} \\
            \text {SR} \\
		\text {LR}
	\end{array}\right)_{t}+w_{12}\left(\begin{array}{c}
		\text {GDP} \\
		\text {CPI} \\
		\text {EQ} \\
            \text {SR} \\
		\text {LR}
	\end{array}\right)_{t}+\cdots+w_{1n}\left(\begin{array}{c}
		\text{GDP} \\
		\text{CPI} \\
		\text {EQ} \\
            \text {SR} \\
		\text{LR}
	\end{array}\right)_{t}. 
\end{gathered}
\ees
Such construction essentially has two implications: (i) the $j$-th star variable depends solely on the $j$-th domestic variable and not on any other variables; (ii) the relationship between the $j$-th star variable and the $j$-th domestic variable is identical to the relationship between the $j'$-th star variable and the $j'$-th domestic variable. In other words, GDP* depends only on the GDP of all countries, not on CPI, EQ, or other variables. Furthermore, the manner in which GDP* depends on the GDP of all countries is exactly the same as how CPI* depends on the CPI of all countries.

Vectorization of the definition of the star variables in Equation \eqref{eq:global} leads to 
\bel{eq:model-gvar-star}
\tvec(\bX_t^*)=   \left(\bW\otimes \bI\right)\tvec(\bX_{t})
\eel
In other words, all the star foreign variables from all countries $\tvec(\bX_t^*)$ depend on all the domestic variables from all countries $\tvec(\bX_{t})$ in a restricted fashion of Kronecker product $\bW\otimes \bI$. 
Combining the GVAR Model \eqref{eq:model-igvar} for all countries and the definition of the star variables for all countries \eqref{eq:global} produces the following joint GVAR model for all countries, which can be used for forecasting purposes,
\bel{eq:model-gvar}
\tvec(\bX_t)=   \diag(\mathbf{\Phi}_1, \mathbf{\Phi}_2,\ldots, \mathbf{\Phi}_n)\tvec(\bX_{t-1}) + \diag(\mathbf{\Psi}_1, \mathbf{\Psi}_2,\ldots, \mathbf{\Psi}_n)\tvec(\bX_t^*) + \tvec(\bE_t),
\eel
where $\tvec(\bX_t^*)$ is defined in \eqref{eq:model-gvar-star}.

It is worth emphasizing that, although it may be tempting to substitute the definition of the star variables \eqref{eq:model-gvar-star} into the joint GVAR Model \eqref{eq:model-gvar}, this approach has not been adopted in the GVAR literature. This is primarily due to the typical assumption of the exogeneity of the star variables. When the definitions of the star variables are substituted, endogeneity becomes inevitable.

In summary, comparing the MAR Model \eqref{eq:model-mar-vec} and the GVAR Model \eqref{eq:model-gvar}, it is seen from the right hand side of the equations that: 1. for the autoregressive lag-one term $\bX_{t-1}$, the MAR model allows both countries and variables to interact, while the GVAR model only incorporates variable interactions and does not permit countries to impact each other; 2. for the contemporaneous lag-zero $\bX_{t-0}$ term, the MAR model does not contain it to capture trade network effect, while the GVAR model only adopts the country interactions, not the variable interactions.


\subsection{SIGMAR models}
\label{ssec:model-sigmar}
Considering the advantages and disadvantages of both the MAR Model \eqref{eq:model-mar-vec} and the GVAR Model \eqref{eq:model-gvar} discussed above, it is natural to explore the following improvement that aims to leverage the strengths of both models simultaneously:
\bel{eq:model-gmar-vec}
\tvec(\bX_t)=  \left(\bB\otimes\bA\right)\tvec(\bX_{t-1}) + (\bW \otimes \bC)\tvec(\bX_{t})  +\tvec(\bE_t),
\eel
where we have introduced a new unknown coefficient matrix $\bC \in \bbR^{k \times k}$ and the other quantities are the same as in Section \ref{ssec:model-review}. 
Model \eqref{eq:model-gmar-vec} will be referred to as the global matrix autoregressive (GMAR) model since it is built upon the MAR model and it incorporates the effect of the global trade network $\bW$. The vectorization version of the GMAR Model \eqref{eq:model-gmar-vec} has an elegant and equivalent matrix form 
\bel{eq:model-gmar}
\bX_t=   \bA\bX_{t-1}\bB^\T + \bC \bX_t \bW^\T + \bE_t
=\bA\bX_{t-1}\bB^\T + \bC \bX_{t}^*+  \bE_t.
\eel
The total number of parameters in GMAR is $2k^2+n^2$, in MAR $k^2+n^2$, both of which are significantly less than that in GVAR $2nk^2$. This significant dimension reduction is crucial in the modeling of global economy given limited length of the time series to preserve sufficient modeling flexibility and possess meaningful interpretability.

Besides the matrix white noise term $\bE_t$, the remaining two terms on the right hand side of the GMAR Model \eqref{eq:model-gmar} have the following interpretations. It is important to point out that both of these two terms model global effect, but from two different perspectives: the first term is autoregressive global effect and the second term is the contemporaneous global effect induced by global trade network.

For the first term $\bA\bX_{t-1}\bB^\T$, it has the same structure as the MAR model, where $(\bA)_{j'j}$ captures the dependence of domestic variable $j'$ at time $t$ on domestic variable $j$ at time $t-1$ and $(\bB)_{i'i}$ captures the dependence of country $i'$ at time $t$ on country $i$ at time $t-1$. In other words, coefficient matrix $\bB$ models the lagged global effect between the countries. 

For the second term $\bC \bX_t \bW^\T = \bC \bX_{t}^*$, it is evident that this term also embodies a global effect, as it incorporates the international trade network $\bW$. It is apparent that the $j$-th star variable at time $t$ can influence the $j'$-th domestic variable at the same time $t$ through $(\bC)_{j'j}$ for all pairs of $j, j'$ and for all countries. The matrix $\bC$ facilitates the interaction of variables for the contemporaneous global effect. The world economy is deeply interconnected despite geopolitical and regional shifts. Consequently, due to the network $\bW$, it is expected that not only do countries have contemporaneous impacts on each other, but economic variables do as well. The GMAR model offers greater versatility than the GVAR model by replacing $\bW \otimes \bI$ with $\bW \otimes \bC$.

Recall that the most general sVAR Model \eqref{eq:model-svar} without considering the global trade network assumes a large-scale autoregressive coefficient matrix $\mathbf{\Phi}\in \mathbb{R}^{kn\times kn}$. Yet, in GMAR Model \eqref{eq:model-gmar-vec}, a special structure of Kronecker product $\bB\otimes\bA$ is assumed. One natural question to raise is whether the GMAR model with $\bB\otimes\bA$ is too restrictive. It is known in the matrix computation literature \citep{van2000ubiquitous,cai2022kopa} that the large matrix $\mathbf{\Phi}$ can be well approximated by the sum of a few terms of Kronecker products $\sum_{l=1}^r\bB_l\otimes\bA_l$, where $r$ can be thought of as the Kronecker product rank, with a similar idea as in the matrix singular value decomposition (SVD). 

However, we will not pursue this direction to enhance the GMAR model. Instead, we focus on the difference between the large matrix $\mathbf{\Phi}$ and its rank-one Kronecker product approximation $\bB \otimes \bA$. Let this difference be represented by the matrix $\bS$, so we have $\mathbf{\Phi} = \bB \otimes \bA + \bS$. To distinguish between these two terms, we assume that $\bS$ is sparse for mathematical identifiability. By applying a certain rearrangement operation, this decomposition of the Kronecker product and sparse components can be reformulated as a matrix decomposition into low-rank and sparse components. The problem of low-rank and sparse matrix decomposition has been intensively studied in the literature, e.g., \citet{candes2011robust, hsu2011robust}. We will pursue this direction for the following economic reasons.

Conditions for all entries of $\bS$ being zero are that, every country has exactly the same lead-lag relationship among all economic variables, up to a scale factor, and also for every variable, countries cross-predict each other in exactly the same manner, also up to a scale factor. We can not easily rule out exceptions to these conditions. If a country consistently has inflation rate above other countries, its inflation rate might have a fairly different effect on future economic growth or asset prices from that in average countries. Or, a country's gdp growth could be negatively autocorrelated while most of the countries exhibit a positive coefficient. Also, country $i$'s short rate may predict country $j$'s short rate positively, but country $i$'s long rate may predict country $j$'s long rate negatively. These exceptions suggest that $\bS$ should contain non-zero elements. Still, the systematic term $\bB\otimes\bA$ should largely capture the cross-predictability, and the idiosyncratic predictability should only occur for a few variables from a few countries. Thus, we set $\bS$ to be sparse.

With the extra sparse term $\bS$ added to the GMAR Model \eqref{eq:model-gmar-vec}, our final proposal for the vectorized version of the Sparsity-Induced Global Matrix AutoRegressive (SIGMAR) model 
is
\bel{eq:model-sigmar-vec}
\tvec(\bX_t)=  \left(\bB\otimes\bA+\bS  \right)\tvec(\bX_{t-1}) + (\bW \otimes \bC)\tvec(\bX_{t})  +\tvec(\bE_t),
\eel
whose matrix form is defined accordingly as
\bel{eq:model-sigmar}
\bX_t=   \bA\bX_{t-1}\bB^\T   + \tvec^{-1}\left[\bS\tvec(\bX_{t-1})\right]+ \bC \bX_t\bW^\T + \bE_t,
\eel
where $\tvec^{-1}$ is the inverse operator of $\tvec$, which converts a long vector of length $kn$ back to a matrix of appropriate size $k\times n$. Here, $\bS \in \bbR^{kn \times kn}$ is an unknown sparse matrix with the number of non-zero entries $s \ll k^2n^2$. We observe that the SIGMAR has a total of $2k^2 + n^2 +s$ parameters to be estimated. Evidently, our model still leads to a substantial dimension reduction relative to sVAR or GVAR when the $\bS$ is highly sparse with small value of $s$. 

Simple algebra further shows that SIGMAR Model \eqref{eq:model-sigmar-vec} can be represented in the form of a VAR model
\bel{eq:model-sigmar-var}
\tvec(\bX_t) = (\bI -  \bW \otimes \bC)^{-1}\left( \bB \otimes \bA   + \bS \right) \tvec(\bX_{t-1}) + \tvec(\widetilde{\bE_t}).
\eel
where $\tvec(\widetilde{\bE_t}) \coloneq(\bI -  \bW \otimes \bC)^{-1} \tvec(\bE_t)$. 

To summarize, comparing the SIGMAR model with the existing models such as iVAR, sVAR, MAR, and GVAR, it has the following advantages: (i) includes the international trade network, whose impact can be evaluated as a result of inclusion; (ii) contains both autoregressive and contemporaneous global effects; (iii) allows variable interaction of the star variables; (iv) captures both systematic and non-systematic effects; (v) maintains the matrix format for clear interpretation; (vi) possesses manageable number of parameters while exhibiting large degree of modeling flexibility.

\subsection{Extensions and connections to other models}
\label{ssec:model-extension}

The very first natural extension of the SIGMAR model is to enlarge the Kronecker product rank from 1 to general $r$ so that the resulting more complicated model could potentially capture more complex patterns in the data
\bel{eq:model-sigmar-rank-r}
\bX_t= \sum_{l=1}^r\bA_l\bX_{t-1}\bB_l^\T +\tvec^{-1}\big(\bS\tvec(\bX_{t-1})\big)
+\bC\bX_{t} \bW^\T + \bE_t.
\eel
For simplicity, we only investigate the rank one $r=1$ case.

Just as the GVAR model could have general lags VARX*($p,q$), the SIGMAR($p,q$) model can be proposed to include more lagged domestic and star terms as well
\bel{eq:model-sigmar-pq}
\bX_t= \sum_{l=1}^p\Big(\bA_l\bX_{t-l}\bB_l^\T +\tvec^{-1}\big(\bS_l\tvec(\bX_{t-l})\big)\Big)
+\bC\bX_{t} \bW^\T
+ \sum_{l = 1}^{q} \bC_l \bX_{t-l}\bW^\T + \bE_t.
\eel
In this article, we focus on the simplest SIGMAR($p=1,q=0$) model. 

There could be multiple relational networks. For the purpose of modeling the global economy, besides the trade flow network $\bW_1$, other networks such as capital flow $\bW_2$, geo-political relationships $\bW_3$, and labor immigration movement $\bW_4$ might also play important roles.
Assuming $m$ networks, SIGMAR could be extended to
\bel{eq:model-sigmar-multiple-w}
\bX_t=   \bA\bX_{t-1}\bB^\T   + \tvec^{-1}\left[\bS\tvec(\bX_{t-1})\right]+ \sum_{l=1}^m\bC_l \bX_t\bW_l^\T + \bE_t,
\eel
so that the effect of multiple networks can be modeled simultaneously. \citet{ren2024multi} considered both spatial and social networks to model Yelp's comment volumes. In this article, we only consider one network.

A few remarks on the comparison and connections between SIGMAR and other models are in order. First, in comparing approaches to macroeconometric modeling, \citet{GLPS_2006} emphasize that the GVAR is a long-run structural cointegrating model, different from the unrestricted VAR without modeling the structure of economy. Our model actually makes the same assumption as that of the conventional VAR and FAVAR, and only considers stationary variables. Our primary goal is rather to provide a novel and parsimonious framework in modeling the joint dynamics of economic variables across countries. We thus leave cointegration error-correction for our future research.


Second, the SIGMAR model bears a resemblance to the SDPD models. The basic SDPD models are formulated as $\by_t \leftarrow \lambda \cdot \bW \by_t$, where the coefficient is a product of an unknown scalar parameter $\lambda$ and a predetermined matrix $\bW$, as detailed in \citet{yu2012estimation}.
Recently, \citet{dou2016generalized} and \citet{gao2019banded} generalized the basic formulation with a scalar $\lambda$ to allow for an unknown diagonal or banded matrix, respectively. However, the assumption that $\bC$ or $\bW\otimes \bC$ is banded is not quite reasonable, since there are no natural orders for the variables or the countries.
Instead, we provide more flexibility to capture the underlying contemporaneous dependence by not imposing any restrictions on the coefficient matrix $\bC$ or $\bW\otimes \bC$. Furthermore, while the SDPD literature typically models vector-valued time series, we model matrix-valued time series.

Third, the SIGMAR model also appears similar to the literature of network regression on the surface \citep{zhu2017network,zhu2023simultaneous,ren2024multi}, but is intrinsically different because of the following fundamental reasons: (i) most of the network regression works focused on vector-valued time series; (ii) they did not have a contemporaneous effect and their network effect was autoregressive lag one; (iii) they did not fully capitalize the special structure of tensor or matrix while we utilize Kronecker product to reduce dimension substantially.

\section{Quasi-maximum likelihood estimation}\label{sec:mle}
To simplify the notations, we denote the underlying true parameters of SIGMAR Model (\ref{eq:model-sigmar-vec}) as $\bA_{0}, \bB_{0}, \bC_{0}$, $\bS_{0}$ and $\sigma_0^2$. Thus Model (\ref{eq:model-sigmar-vec}) can be rewritten as  
\bel{eq:mle-1}
(\bI_{kn} - \bW \otimes \bC_{0}) \tvec(\bX_t)=  \left(\bB_{0} \otimes\bA_{0}+\bS_{0}  \right)\tvec(\bX_{t-1})  +\tvec(\bE_t) , \quad t = 1, \ldots, T,
\eel
where the noise $\{\bE_{t}\}_{ij}$ is $i.i.d.$ across $i$, $j$ and $t$ with mean zero and variance $\sigma_{0}^2$. Please refer to Assumption \ref{assum:4} for more discussion on the necessity of such assumption of the noise.

Denote $\btheta \coloneq \left(\tvec(\bC)^\T, \tvec(\bPhi)^\T, \sigma^2\right)^\T$, where $\bPhi \coloneq \left(\bB \otimes\bA+\bS\right)$, and similarly for the true parameters, $\btheta_{0} \coloneq \left(\tvec(\bC_{0})^\T, \tvec(\bPhi_{0})^\T, \sigma^2_{0}\right)^\T$, where $\bPhi_{0} \coloneq \left(\bB_{0} \otimes\bA_{0}+\bS_{0}\right)$. The log-quasi-likelihood of Model (\ref{eq:mle-1}) is 
\bel{eq:mle-4}
\ln \calL_{T}(\btheta) =  -\frac{knT}{2}\ln 2\pi - \frac{knT}{2}\ln \sigma^2 + T \ln \left|\bI_{kn} - \bW \otimes \bC \right| - \frac{1}{2\sigma^2}\sum_{t = 1}^{T} \bv_t^\T \bv_t,
\eel
where $\bv_t \coloneq (\bI_{kn} - \bW \otimes \bC) \tvec(\bX_t) - \bPhi \tvec(\bX_{t-1})$. Note that the likelihood function $\ln \calL_{T}(\btheta)$ is a function of $\bC, \bPhi$, and $\sigma^2$, not a function of $\bA,\bB,\bC,\bS$, and $\sigma^2$. 
The QMLE, $\hat{\btheta}^{mle}\coloneq \left(\tvec(\hbC^{mle})^\T, \tvec(\hbPhi^{mle})^\T, (\hsigma^{2})^{mle}\right)^\T$ is the estimator obtained by maximizing $\ln \calL_{T}(\btheta)$.

When optimizing the objective function (\ref{eq:mle-4}), a closed-form solution is not available. Instead, the augmented Lagrange method \citep{robinson1972quadratically, ye1988interior}, a widely used approach for general nonlinear optimization problems, can be applied. To achieve faster algorithmic convergence with the augmented Lagrange method, a carefully designed initialization is provided to the algorithm. This initialization is derived from a naive alternating minimization method \emph{without} bias correction.
Section \ref{sec:alternating} details an alternating minimization method that incorporates bias correction. For the efficient computation of the term $|\bI_{kn} - \bW \otimes \bC|$ in the log-likelihood function, the eigenvalues of $\bW$ are precomputed and remain fixed throughout the optimization process.
The effectiveness of the QMLE is assessed through simulation studies presented in Section \ref{sec:simulation}.

In Section \ref{sec:theory}, we will investigate the consistency and asymptotic distribution of $\hbtheta^{mle}$. To establish these asymptotic properties, the first step is to address the identifiability issue of the parameters. Theorem \ref{thm:1} states that, under some mild conditions, the true parameters can be identified as the unique maximizer of $\bbE(\ln \calL_{T}(\btheta))$.
Once identifiability is established, the first and second derivatives of the log-likelihood function (\ref{eq:mle-4}) are derived in Equations (\syb{S74}) and (\syb{S75}) in the appendix. The asymptotic properties of the first derivative of $\ln\calL_T$ at the true parameter value $\btheta_0$ are discussed in Section \syb{S7.3} of the appendix. Consequently, the consistency and asymptotic distribution of $\hbtheta^{mle}$ are established in Theorems \ref{thm:1} and \ref{thm:2}, respectively.
Once a consistent estimate $\hbPhi^{mle}$ is achieved, we can further recover $\bA$, $\bB$, and $\bS$ via a projection method, to be explained in the following subsection.
 
\subsection{Projection of the QMLE}\label{sec:alg-projection}
Note that during the process of solving for $\hbtheta^{mle}$ as described above, the structural constraint on $\bPhi$ is not considered. In this subsection, our goal is to  decompose $\hbPhi^{mle}$  as the summation of a Kronecker product and a sparse matrix by solving the  following optimization problem:
\bel{eq:proj-1}
&& \min_{\bA,\bB, \bS} \|\hbPhi^{mle} - \bB \otimes \bA - \bS\|_{F}^2 ,\cr
&& \text{subject to } \|\tvec(\bS)\|_{1}\leq s.
\eel
If we set  the number of non-zero entries to be zero, i.e. $s = 0$, the  minimization above becomes a projection of $\hbPhi^{mle}$ onto the space of Kronecker products, which is known as the Nearest Kronecker Product (NKP) problem \citep{van2000ubiquitous,cai2022kopa}. The NKP problem has an explicit solution, which is obtained by performing a singular value decomposition after rearranging  $\hbPhi^{mle}$.  The essential idea is that $\bB \otimes \bA$ becomes a rank one matrix after the rearranging operation. Therefore, for $s>0$, we still make use of the rearranging technique to transform problem (\ref{eq:proj-1}) into the task of approximating a matrix by a low-rank matrix and a sparse component.

Thus, let's first define the rearranging operator $\mR$. For any matrix $\bM$ that is a $p_1\times p_2$ array of blocks of the same block size $d_1\times d_2$, let $M_{j,k}^{d_1,d_2}$ be its $(j,k)$-th block, $1\le j\le p_1$, $1\le k\le p_2$. Furthermore, let the operator $\mR_{p_1,p_2,d_1,d_2}: \mathbb{R}^{(p_1d_1)\times (p_2d_2)}\rightarrow \mathbb{R}^{(p_1p_2)\times (d_1d_2)}$ be a mapping such that
\bel{R1:1}
\mR_{p_1,p_2,d_1,d_2}(\bM) = \left[\tvec(\bM_{1,1}^{d_1,d_2}),\ldots, \tvec(\bM_{p_1,1}^{d_1,d_2}),\ldots, \tvec(\bM_{1,p_2}^{d_1,d_2}),\ldots,\tvec(\bM_{p_1,p_2}^{d_1,d_2})\right]^\T.
\eel
When applying the operator $\mR_{p_1,p_2,d_1,d_2}$ to a Kronecker product $\bW \otimes \bC$, where $\bW\in \mathbb{R}^{p_1\times p_2}, \bC \in \mathbb{R}^{d_1\times d_2}$,  it holds that
\bel{R1:2}
\mR_{p_1,p_2,d_1,d_2}(\bW \otimes \bC)=\tvec(\bW)[\tvec(\bC)]^\T.
\eel
The property (\ref{R1:2}) is extremely helpful throughout the estimation process and theoretical analysis.
Recall that $\bPhi_{0} = \bB_{0} \otimes\bA_{0}+\bS_{0}$, and we can apply the operator $\mR_{n,n,k,k}(\cdot)$ on both sides so that 
\bes
\tbPhi_{0}= \tvec(\bB_{0}) \tvec(\bA_{0})^\T + \tbS_{0},
\ees
where $\tbPhi_{0} \coloneq \mR_{n,n,k,k}(\bS_{0})$ $\tbS_{0} \coloneq \mR_{n,n,k,k}(\bS_{0})$. It is evident that the rank-one matrix $ \tvec(\bB_{0}) \tvec(\bA_{0})^\T$ is a low-rank matrix and $\tbS_{0}$ remains sparse after rearranging sparse $\bS_0$. Presuming $\hbPhi^{mle}$ is a consistent estimator of $\bPhi_{0}$, there is a matrix $\bN_0$ with $\|\bN_0\|_F \leq \delta$ for some $\delta>0$, such that $\hbPhi^{mle} = \bPhi_{0} + \bN_0 $.  Denoting ${\tbPhi}  = \mR_{n,n,k,k}(\hbPhi^{mle})$ and $ \widetilde{\bN}_0 \coloneq \mR_{n,n,k,k}({\bN}_0)$, we have
\bes
{\tbPhi} = \tvec(\bB_{0}) \tvec(\bA_{0})^\T + \tbS_{0} + \widetilde{N}_0.
\ees
Therefore, we consider solving the following convex optimization problem
\bel{opt:rpca}
 && \min_{\bL, \tbS} \|\bL\|_{*} + \lambda\|\tvec(\tbS)\|_{1} ,\cr
&& \text{subject to } \|{\tbPhi}  - \bL  - \tbS\|_F \leq \delta,
\eel
where $\|\cdot\|_*$ stands for the nuclear norm. The solution $(\bL^{proj}, \tbS^{proj})$ to the convex program (\ref{opt:rpca}) can be obtained via an alternating direction method \citep{yuan2009sparse}, with its convergence property well-studied in \citet{lin2010augmented, yuan2009sparse}. The choice of $\lambda$ and the details of the implementation can be found in Section \syb{S1} of the appendix.
As a consequence, we can obtain $\tvec(\hbA^{mle})$ by taking the top right singular  vector from $\bL^{proj}$, and $\tvec(\hbB^{mle})$ is calculated as the  top singular value multiplied by the top left singular vector. In addition, the sparse matrix can be recovered after applying the inverse of the rearranging operator, $\hbS^{mle} = \mR_{n,n,k,k}^{-1}(\tbS^{proj})$.
Furthermore, under mild conditions, we show that the estimators above $\hbB^{mle}$, $\hbA^{mle}$ and $\hbS^{mle}$  are consistent in Proposition \ref{prop:rpca}. 

\section{Theoretical analysis}\label{sec:theory}
In this section, we establish the consistency and asymptotic normality of the QMLE for our SIGMAR model. We begin by presenting the following necessary assumptions.

\begin{assumption}\label{assum:3}
$\bW$ is a deterministic known matrix with $\diag(\bW)=\bf{0}$. Moreover, the diagonal elements of $\bW^\top \bW$ are not all the same,  and $\bW + \bW^\T\neq \mathbf{0}_{n\times n}$.
\end{assumption}

The condition imposed on $\bW$ in the first sentence of Assumption \ref{assum:3} is standard for achieving identifiability in vector-valued time series. This condition has been considered in various contexts, including network analysis \citep{ma2020naive}, spatial autoregression \citep{yu2008quasi}, and GVAR \citep{pesaran2004modeling, pesaran2009forecasting}. Our SIGMAR model, which targets matrix-valued time series, is accordingly more complex. Therefore, it is natural to expect additional conditions, as stated in the second sentence of Assumption \ref{assum:3}, to ensure identification due to contemporaneous spatial effects. In practice, these conditions are not difficult to satisfy. For instance, in the global economy data discussed in Section \ref{sec:real}, $\bW$ is constructed as the global trade network with row normalization, ensuring that the entries of $\bW$ are non-negative.

\begin{assumption}\label{assum:1}
Assume that the determinant of $\bI_{kn} - \bW \otimes \bC$ is positive for any $\bC$ in a compact space $\calC$. 
The true parameter $\bC_{0}$ is in the interior of $\calC$.	
Moreover, assume that $\rho\left((\bI_{kn} -  \bW \otimes \bC)^{-1}\left(\bB \otimes \bA  + \bS \right)\right) <1$ for $\bC \in \calC$, $\bB \in \mathcal{B}, \bA \in \calA$ and $\bS \in \calS$, where $\calB, \calA$ and $\calS$ are compact parameter spaces. The true parameters $
	\bB_{0}, \bA_{0}$ and $\bS_{0}$ are in the interior of their respective parameter spaces.
\end{assumption}

 The condition $|\bI_{kn} - \bW \otimes \bC|>0$ over the compact set $\calC$ implies that  both $\bI_{kn} - \bW \otimes \bC$ and its inverse  are uniformly bounded, which is crucial for the stability of spatial econometric models \citep{lee2004asymptotic}. In practice, the QMLE typically avoids the boundary, where $|\bI_{kn} - \bW \otimes \bC| = 0$, provided that $\bC_0$ lies within the interior of $\calC$. This is because the log-likelihood function may diverge to negative infinity as $\bC$ approaches the boundary.
 
Furthermore, the condition $\rho\left((\bI_{kn} -  \bW \otimes \bC)^{-1}(\bB \otimes \bA  + \bS )\right) <1$ in Assumption \ref{assum:1} is  a standard condition for ensuring the stationarity of $\bX_t$. Such an assumption has been considered in the spatial econometrics literature, as seen in works such as \citet{yu2008quasi}, \citet{yu2012estimation}, \citet{dou2016generalized}, and \citet{gao2019banded}.  Moreover, the compactness requirement for parameter spaces ($\calC, \calB, \calA$ and $\calS$) is  necessary to establish the uniform convergence of the log likelihood function.

\begin{assumption}\label{assum:4}
	The noise $\{\bE_{t}\}_{ij}$, $1 \leq i \leq k, 1 \leq j \leq n$ and $1 \leq t\leq T $, are $i.i.d.$ noise with $\E(\{\bE_{t}\}_{ij})=0$, $\Var(\{\bE_{t}\}_{ij})=\sigma_{0}^2$ and $\bbE|\{\bE_{t}\}_{ij}|^{4+\gamma} < \infty$ for some $\gamma >0 $.
\end{assumption}

Assumption \ref{assum:4} is a regular condition for the error term in spatial econometrics \citep{yu2008quasi,yu2012estimation}.
We note that the independent and equal variance assumption is crucial for addressing identification issues in this paper.  
Specifically, consider a simple process $(\bI_{kn} - \bW \otimes \bC_0)\tvec(\bX_t) = \tvec(\bE_t)$ for a period $t$. Under Assumption \ref{assum:4}, the identifiability of $(\bC_0,\sigma_0^2)$ in $\bbE\calL(\bC,\sigma^2)$ is equivalent to the identifiability in the precision matrix of $\tvec(\bX_{t})$, given by $1/\sigma_0^2\cdot (\bI_{kn} - \bW \otimes \bC_0)^\T(\bI_{kn} - \bW \otimes \bC_0)$. Thus, we can prove that $(\bC_0,\sigma_0^2)$ is identifiable, with detailed proof provided in Appendix \syb{S6.1}. On the other hand, in the general case where $\mathrm{cov}(\tvec(\bE_t))$ is an invertible matrix $\bSigma$, the precision matrix of $\tvec(\bX_t)$ takes the form $(\bI_{kn} - \bW \otimes \bC)^{\T} \bSigma^{-1} (\bI_{kn} - \bW \otimes \bC)^{}$. In this case, $\bSigma$ and $\bC$ are not identifiable due to such matrix product structure.  
Furthermore, this equal variance condition is also considered in network analysis \citep{zhu2023simultaneous} and  structural equation modeling \citep{peters2014identifiability, chen2019causal}.

{Define the parameter space as $\bTheta \coloneq \{\btheta=\left(\tvec(\bC)^\T, \tvec(\bPhi)^\T, \sigma^2\right)\big| \bPhi=\bB\otimes\bA+\bS, \bC \in \calC, \bB \in \calB, \bA \in \calA, \bS \in \calS, \sigma^2 <\infty \}$.} 
Now we are ready to state our main theorem on the consistency of QMLE.

\begin{theorem}\label{thm:1}

Under Assumptions \ref{assum:3} - \ref{assum:4}, we have $\btheta_{0}$ is a unique global maximizer of $\bbE\calL(\btheta)$, and the QMLE converges in probability to the target, 
    \bel{eq:theory-3}	\left(\tvec(\hbC^{mle})^\T, \tvec(\hbPhi^{mle})^\T, (\hsigma^{2})^{mle}\right) \overset{p}{\rightarrow} \left(\tvec(\bC_{0})^\T, \tvec(\bPhi_{0})^\T, \sigma^2_0\right) .
	\eel
\end{theorem}

\begin{remark}
    The proof of Theorem \ref{thm:1} relies on the global identification of $\btheta_0$, and the following two properties:  (i) $T^{-1} \left(\ln\calL_T(\btheta) - \bbE\ln \calL_T(\btheta)\right) $ converges  in probability to zero uniformly for $\btheta \in \bTheta$; (ii) $T^{-1}\bbE\calL_T(\btheta)$ is uniformly  equicontinuous.
\end{remark}

Having established the consistency of the estimator, we now turn our attention to the asymptotic distribution of QMLE.
The asymptotic distribution of $\hbtheta^{mle}$ can be derived from the Taylor expansion of $ 
 \partial \calL_{T}(\hbtheta^{mle})/\partial \btheta$ around $\btheta_0$. 
 At $\btheta_0$, the first-order derivative of $\ln\calL_T(\btheta)$  is given in equation (\syb{S76}) of the appendix.
If we further define the variance matrix of  $\frac{1}{\sqrt{T}} \frac{\partial \calL_{T}(\btheta_{0})}{\partial \btheta}$ as
\bel{dis-2}
\Psi_{\btheta_{0},T} \coloneq  \bbE \left(\frac{1}{\sqrt{T}} \frac{\partial \calL_{T}(\btheta_{0})}{\partial \btheta}  \cdot \frac{1}{\sqrt{T}} \frac{\partial \calL_{T}(\btheta_{0})}{\partial \btheta^\T} \right).
\eel
Then the asymptotic distribution of $\frac{1}{\sqrt{T}} \frac{\partial \calL_{T}(\btheta_{0})}{\partial \btheta}$ can be derived from the central limit theorem for martingale difference arrays. The results are stated in the following proposition.
\begin{proposition}\label{prop:2}
	Under Assumptions \ref{assum:3} - \ref{assum:4}, 
	\bel{eq:theory-4}
	\frac{1}{\sqrt{T}} \frac{\partial \ln \calL_{T}(\btheta_{0})}{\partial \btheta} 
	\overset{d}{\Rightarrow} \calN(0,\Psi_{\btheta_{0}} ),
	\eel
    where $\Psi_{\btheta_{0}} \coloneq \lim_{T \rightarrow \infty} \Psi_{\btheta_{0},T}$.
\end{proposition}
By Taylor expansion, we have
\bel{eq:theory-5}
\sqrt{T} 	(\hbtheta^{mle} -  \btheta_{0} ) = \left(-\frac{1}{{T}} \frac{\partial^2 \ln \calL_{T}(\bar{\btheta})}{\partial^2 \btheta}\right)^{-1} \cdot  \left(\frac{1}{\sqrt{T}} \frac{\partial \ln \calL_{T}(\btheta_{0})}{\partial \btheta}\right),
\eel
where  $\bar{\btheta}$ lies between $\btheta_{0}$ and $\hbtheta^{mle}$. Moreover, under Assumptions \ref{assum:3} - \ref{assum:4}, we further have $	\left(	-\frac{1}{{T}} \frac{\partial^2 \ln \calL_{T}(\bar{\btheta})}{\partial {\btheta} \partial \btheta^\T} - \left(	-\frac{1}{{T}} \frac{\partial^2 \ln \calL_{T}({\btheta}_{0})}{\partial{\btheta} \partial \btheta^\T} \right)\right) = \|\hbtheta^{mle} - \btheta_{0}\|_2 \cdot O_{p}(1)$, and $	\left(	-\frac{1}{{T}} \frac{\partial^2 \ln \calL_{T}({\btheta}_{0})}{\partial{\btheta} \partial \btheta^\T} \right) - \Xi_{\btheta_{0},T} = O_{p}(\frac{1}{\sqrt{T}})$ 
where $\Xi_{\btheta_{0},T} $ is the information matrix defined as
\bel{def-xi}
\Xi_{\btheta_{0},T}	\coloneq - \bbE \left(\frac{1}{{T}} \frac{\partial^2 \calL_{T}(\btheta_{0})}{\partial  \btheta \partial \btheta^\T} \right) .
\eel
Thus we have $-\frac{1}{{T}} \frac{\partial^2 \ln \calL_{T}(\bar{\btheta})}{\partial \btheta \partial \btheta^\T} =  \Xi_{\btheta_{0},T} + O_p(\frac{1}{\sqrt{T}})$.
Furthermore, denote $\Xi_{\btheta_{0}} \coloneq \lim_{T \rightarrow \infty} \Xi_{\btheta_{0},T}$.
Combining with Proposition \ref{prop:2}, we have the following theorem for the distribution of $\hbtheta^{mle}$. 
\begin{theorem}\label{thm:2}
Under Assumptions \ref{assum:3} - \ref{assum:4}, 	\bel{eq:theory-6}
	\sqrt{T}(\hbtheta^{mle} -  \btheta_{0} ) \overset{d}{\Rightarrow} \calN(0, \Xi_{\btheta_{0}}^{-1} \Psi_{\btheta_{0}} \Xi_{\btheta_{0}}^{-1}).
	\eel
Additionally, if $\tvec(\bE_t) \sim \calN(0,\sigma_{0}^2 \cdot \bI_{kn})$ for $t = 1,\ldots, T$, we further have 
\bel{eq:theory-7}
\sqrt{T}(\hbtheta^{mle} -  \btheta_{0} ) \overset{d}{\Rightarrow} \calN(0, \Xi_{\btheta_{0}}^{-1} ).
\eel
\end{theorem}
The explicit expressions of $\Xi_{\btheta_{0}}^{-1}$ and $ \Psi_{\btheta_{0}}$ are provided in Section \syb{S7.3} and \syb{S7.4} of the appendix. Finally, we prove the estimation consistency for $\bA_0$, $\bB_0$, and $\bS_0$ with our projection method.
\begin{proposition}\label{prop:rpca}
Under Assumptions \ref{assum:3} - \ref{assum:4} and certain regularity conditions (deferred to Appendix \syb{S2}) on the sparsity patterns of $\bS_0$, $\tvec(\bA_0)$ and $\tvec(\bB_0)$, 
the estimations $\hbB^{mle}, \hbA^{mle}$,  and $\hbS^{mle}$ as the solution to the convex program (\ref{opt:rpca}) satisfies 	
\bel{eq:theory-9}
	\|\hbB^{mle} - \bB_{0} \|_F + \| \hbA^{mle} - \bA_{0}\|_F + \|\hbS^{mle} - \bS_{0}\|_F  = O_{p}(1/\sqrt{T}).
	\eel
\end{proposition}

Proposition \ref{prop:rpca} guarantees the estimation consistency of the target matrices $\bA_0$, $\bB_0$, and $\bS_0$ using the projection method. To separate $\bS_0$ from $\bL_{0} = \tvec(\bB_{0}) \tvec(\bA_{0})^\T$, a regularity condition on the sparsity pattern of $\bS_0$, $\bA_0$, and $\bB_0$ is required. We refer to the appendix \syb{S2} for a detailed discussion. 
Similar regularity conditions have been discussed in the robust principal component analysis literature \citep{hsu2011robust, candes2011robust}. In practice, the relationships among economic variables (reflected by $\bA_0$) and among countries (captured by $\bB_0$) suggest the validity of a non-sparse assumption. Furthermore, the sparsity pattern assumption on $\tbS_{0}$ requires that the nonzero entries of $\tbS_{0}$ are not concentrated in a single row or column.

\section{Estimation enhancement via bias correction}\label{sec:alternating}
We have demonstrated the consistency of the QMLE in Section \ref{sec:theory}. However, the log-likelihood function in (\ref{eq:mle-4}) does not incorporate the structural information of $\bPhi$. Although the projection approach enables us to compute $\bA$, $\bB$, and $\bS$ following the QMLE step, it may not yield optimal estimation efficiency. In this section, we explore an alternative method using alternating minimization to enhance the accuracy of our estimates. Additionally, we introduce a bias correction step to further improve the results.

To account for the sparseness of $\bS$, we consider the following $\ell_1$-regularized minimization problem 
\bel{eq:8}
 \min_{\bA,\bB, \bC,\bS} \frac{1}{T}\sum_{t=1}^T\|\bX_t - \bC\bX_{t} \bW^\T-\bA\bX_{t-1}\bB^\T   -\tvec^{-1}\left[\bS\tvec(\bX_{t-1})\right]\|_F^2  + \lambda \|\tvec(\bS)\|_{1}, 
\eel
where $\lambda$ is a regularization parameter to control the sparsity level. 
Given an appropriate initialization, the optimization problem can be solved by an alternating minimization algorithm (AMA). 

In AMA, we alternatingly update $\bA$, $\bB$, $\bS$, and $\bC$ while keeping others fixed. For example, updating $\bA$ given $\hbB$, $\hbS$, and $\hbC$ can be obtained by ordinary least square estimation with closed-form solution: 
\bes
\hbA = \left(\sum_t \bX_t ^{ab}\hbB \bX_{t-1}^\T\right) \left(\sum_{t} \bX_{t-1}\hbB^\T \hbB \bX_{t-1}^\T \right) ^{-1},
\ees
where $\bX_t^{ab} \coloneq \bX_t - \hbC\bX_t\bW^\T - \tvec^{-1}\left[\hbS\tvec(\bX_{t-1})\right] $. 
Similarly, $\bB$ can be obtained by
\bes
\hbB = \left(\sum_t (\bX_t^{ab})^\T \hbA \bX_{t-1} \right) \left(\sum_{t} \bX_{t-1}^\T\hbA^\T \hbA \bX_{t-1} \right) ^{-1}.
\ees
To update $\bS$ given $\hbA, \hbB$ and $\hbC$, we consider the following Lasso-type problem
\bes
\min_{\bS}\sum_{t=1}^{T}\|\tvec(\bX^s_{t}) - \bS\tvec(\bX_{t-1}) \|^2_2 + \lambda \|\tvec(\bS)\|_{1},
\ees
where $\bX^s_{t} \coloneq \bX_t - \hbC \bX_t \bW^\T - \hbA \bX_{t-1} \hbB^\T$. Here, we choose the regularization parameter $\lambda$ by Bayesian Information Criteria. 

Now we consider the update of $\bC$. Given $\hbA$, $\hbB$ and $\hbS$, a straightforward estimate for $\bC$ is 
\bel{eq:bc-3}
\hbC^{lse} = \left(\frac{1}{T}\sum_{t=1}^{T} \bX_t^c \bW \bX_t^\T \right)\left( \frac{1}{T}\sum_{t=1}^{T}\bX_t\bW^\T \bW \bX_t^\T \right)^{-1},
\eel
where 
$\bX_t^c \coloneq \bX_t - \tvec^{-1}\left[\hbPhi\tvec(\bX_{t-1})\right]$.
However, we shall note that such an estimation for $\bC$ may not be consistent. This occurs due to the correlation between $\bX_t \bW^\T$ and $\bE_t$. 
Such an issue of inconsistency also arises in other spatial-temporal models such as \citet{dou2016generalized}, \cite{gao2019banded}, and \cite{ma2023sparse}. To address this problem, the generalized Yule-Walker equation has been considered in the literature. We refer to \citet{dou2016generalized} for more details on the generalized Yule-Walker equation. In this paper, we consider an alternative approach of bias-correction to address this issue.

To introduce the bias-corrected estimator, we first define some additional notations. Let $\bGamma_w =   \frac{1}{T}\sum_{t=1}^{T}\bX_t\bW^\T \bW \bX_t^\T$, and error-related term
$\tvec(\tbE_t) \coloneq (\bI_{kn} - \bW \otimes \bC_0)^{-1} \tvec(\bE_t)$. The covariance matrix of this term is ${{\tbSigma}} \coloneq (1/T)\sum_{t=1}^{T} (\tbE_t \otimes \tbE_t)$, which can be estimated based on observations. 
Then, the inconsistency of $\hbC^{lse}$ can be suggested by
\begin{align}\label{eq:bc-10}
(\hbC^{lse} - \bC_{0})\bGamma_w  =  \tbSigma_{w} - \bC_{0}\tbSigma_{w^2} + 
\tvec^{-1}\left(\mathbf{Q}_w\tvec(\bPhi_{0} - \hbPhi)\right) + o_p(1).
\end{align}
where the definitions of $\tbSigma_w$, $\tbSigma_{w^2}$, and $\mathbf{Q}_w$, along with the detailed procedure for deriving equation \eqref{eq:bc-10},  can be found in Appendix \syb{S3}. Specifically, $\tbSigma_w$ and $\tbSigma_{w^2}$ solely depend on $\tbSigma$ and $\bW$.  Moreover, when $\hat{\bPhi}$ is a consistent estimator of $\bPhi_0$, the third term on the right-hand side of (\ref{eq:bc-10}) becomes negligible. This motivates us to propose the following bias-corrected estimator:
\bel{eq:bc-12}
\hbC^{bc}  = (\hbC^{lse}\bGamma_w  -  \tbSigma_{w})(\bGamma_w -\tbSigma_{w^2})^{-1}  .
\eel
We summarize the AMA with the bias-correction in Algorithm \ref{alg:1} below.

\begin{algorithm}
	\renewcommand{\algorithmicrequire}{\textbf{Input:}}
	\renewcommand{\algorithmicensure}{\textbf{Output:}}
	\caption{Alternating Minimization Algorithm with bias-correction for SIGMAR}
	\label{alg:1}
	\begin{algorithmic}[1]
		\REQUIRE $\bX_t$, $t=1,\ldots, 	T$.
		\STATE Initialization: the elements of $\hbA^{(0)}$ and $\hbB^{(0)}$ are independently  generated from $\calN(0,1)$, and $\hbS^{(0)}$ is set to the zero matrix.
		\FOR{j in $0,1,2,
        \ldots, J-1$}
		\STATE {\text{Updating $\bC$:} \\
			\quad $\hbC^{lse} = \argmin_{\bC} 1/T\sum_{t=1}^{T}  \|  \bX_t - \tvec^{-1}\left[\hbPhi^{(j)}\tvec(\bX_{t-1})\right]- \bC\bX_{t} W^\T\|_F^2$. \\
            \quad Bias-correction: 
			\quad $[\hbC^{bc}]^{(j+1)} \leftarrow (\hbC^{lse}\bGamma_w  -  \tbSigma_{w})(\bGamma_w -\tbSigma_{w^2})^{-1}$.
            }
		\STATE {Updating $\bA,\bB$: \\
			\quad Denote $\bX_t^{ab} = \bX_t -[\hbC^{bc}]^{(j+1)}  \bX_t\bW^\T - \tvec^{-1}\left[\hbS^{(j)}\tvec(\bX_{t-1})\right]$. \\
			\quad $\hbA^{(j+1)} \leftarrow \left(\sum_t \bX_t ^{ab}\hbB^{(j)} \bX_{t-1}^\T\right) \left(\sum_{t} \bX_{t-1}(\hbB^{(j)})^\T \hbB^{(j)} \bX_{t-1}^\T \right) ^{-1}.$ \\
			\quad $\hbB^{(j+1)} \leftarrow \left(\sum_t (\bX_t^{ab} )^\T \hbA^{(j+1)} \bX_{t-1} \right) \left(\sum_{t} \bX_{t-1}^\T( \hbA^{(j+1)})^\T \hbA^{(j+1)} \bX_{t-1} \right) ^{-1}$.\\
                \quad Normalization:  $\hbB^{(j+1)} \leftarrow \hbB^{(j+1)}\times \|\hbA^{(j+1)}\|_F$, $\hbA^{(j+1)} \leftarrow \hbA^{(j+1)}/ \|\hbA^{(j+1)}\|_F$.
		}
		\STATE {Updating $\bS$: \\
			\quad Denote $\bX^s_{t} = \bX_t - [\hbC^{bc}]^{(j+1)}  \bX_t \bW^\T - \hbA^{(j+1)} \bX_{t-1} (\hbB^{(j+1)})^\T$.\\
			\quad  $\hbS^{(j+1)} \leftarrow \min_{\bS}\sum_{t=1}^{T}\|\tvec(\bX^s_{t}) -  \bS\tvec(\bX_{t-1}) \|^2_F + \lambda \|\bS\|_{\tvec(1)}$.
            }
		\ENDFOR
		\STATE \textbf{return} $\hbA^{(J)}$, $\hbB^{(J)}$, $\hbS^{(J)}$ and $[\hbC^{bc}]^{(J)}$.
	\end{algorithmic}  
\end{algorithm}

\section{Simulations}\label{sec:simulation}
In this section, we conduct a simulation study under SIGMAR Model \eqref{eq:model-sigmar},
where each entry of $\bE_{t}$ is independent and drawn from the standard normal distribution. We investigate different settings for various choices of matrix dimensions $k$ and $n$, as well as the length of the time series $T$.

Specifically, we consider $(k,n) $ equal to $(3,4)$, $(4,6)$ and $(5,10)$ and four different time lengths: $T = 100, 500, 1000, 2000$.
For given dimensions $k$ and $n$, the $\bA_{0}$, $ \bB_{0}$, and $\bC_{0}$ are {randomly generated from a standard normal distribution} and then rescaled such that $\rho(\bA_{0})\rho(\bB_{0}) = 0.6$ and $\rho(\bC_{0}) = 0.6$. The location of the non-zero entries of $\bS$ are randomly selected and assigned a value of either $0.15$ or $-0.15$. Additionally, the number of non-zero entries of $\bS$ is set to  10, 20, and 30, corresponding to the three sets of dimensions of $(k,n)$, respectively. Moreover, we randomly generate the weight matrix $\bW$ from a standard uniform distribution, set the diagonal entries to zero, and normalize each row such that the sum of the entries in each row  equals one. Hence, $\bW$ satisfies the requirements in Assumption \ref{assum:3}. The coefficients generated as described above satisfy Assumption \ref{assum:1}.  For a particular simulation setup with multiple repetitions, the coefficient matrices $\bA_{0}, \bB_{0}, \bC_{0}$ and $\bS_{0}$ remain fixed.

For each configuration, we repeat the experiment 200 times, and use the following relative error to assess the estimation performance
\bes
\|\hbC - \bC_{0}\|_F/\|\bC_0\|_F \text{ and } \|\hbPi - \bPi_{0}\|_F/\|\bPi_0\|_F,
\ees
where $\bPi_0\ \coloneq (\bI_{kn} - \bW \otimes \bC_{0})^{-1}(\bB_0 \otimes \bA_0 + \bS_0)$. 
Furthermore, to evaluate the selection performance of $\bS$, we record the False Positive Rate (FPR) and True Positive Rate (TPR). Specifically, denote $\bI(\cdot)$ be the indicator function and define
\bes
\text{FPR} && = \sum_{j= 1}^{kn} \sum_{\ell= 1}^{kn} \frac{\bI(\{\hbS\}_{j\ell} \neq 0)\bI(\{\bS_{0}\}_{j\ell}  = 0)}{\bI(\{\bS_{0}\}_{j\ell}  = 0)}, \cr
\text{TPR} && = \sum_{j= 1}^{kn} \sum_{\ell= 1}^{kn} \frac{\bI(\{\hbS\}_{j\ell} \neq 0)\bI(\{\bS_{0}\}_{j\ell}  \neq 0)}{\bI(\{\bS_{0}\}_{j\ell}  \neq  0)}.
\ees

We denote our estimation methods by QMLE and bias-corrected alternating minimization  as QMLE and BC, respectively. Additionally, we implement sVAR \citep{stock2001vector}  and MAR \citep{chen2021autoregressive},  as benchmarks for performance comparison. For different dimensional settings and time lengths, we calculate the FPR, TPR and relative errors of $\hbC$ and $\hbPi$, and report the mean and standard errors  in Table \ref{table:sim-1}. It is important to note  that sVAR and MAR do not consider the contemporaneous global term $\bC$ and the sparse term $\bS$, and therefore only the relative error of $\hbPi$ is reported for these two methods.

It is evident that both QMLE and BC demonstrate competitive performance in terms of estimation and variable selection. Specifically, with respect to variable selection, we can see that our QMLE and BC are able to achieve higher TPR and lower FPR with longer time series.  Additionally, for relatively small dimensions $(k,n) = (3,4)$, QMLE attains a higher TPR compared to BC.  Furthermore, BC generally exhibits a smaller FPR compared to QMLE in most scenarios. While QMLE achieves a slightly smaller FPR under the setting $(k,n) = (5,10)$ and $T = 100$, this comes at a cost of a lower TPR. It is worth noting that  under this limited time length scenario (i.e., $T = 100$), BC achieves a TPR exceeding  $73\%$, thereby endorsing its applicability in the followup real data analysis.

In terms of the estimation of $\bC_0$, we can observe that the relative error and standard error of our QMLE and BC methods decrease as the time length increases. Specifically, QMLE demonstrates superior performance when $(k,n) = (3,4)$, indicating that it is the preferred choice in small-dimension settings. On the other hand, the BC method performs better than QMLE as the dimensions increase, showing a slightly faster convergence. 

Regarding the estimation performance of $\bPi_0$, it is observed that sVAR, QMLE, and BC all exhibit consistent properties, i.e., they demonstrate smaller estimation errors with larger sample sizes, whereas MAR does not. The consistency of sVAR is expected because the SIGMAR model is a special case of the sVAR model. Therefore, sVAR estimation is not completely off track, although it is less efficient compared to our QMLE and BC methods. The inconsistency of the MAR method is also anticipated, as the MAR model only includes the autoregressive term and not the contemporaneous one, making it incompatible with the data-generating model SIGMAR. Notably, BC demonstrates the best performance in most cases, with QMLE ranking second.
Furthermore, for sVAR, when the dimensions are relatively small, such as $(k,n) = (3,4)$ or $(4,6)$, and the sample size is large ($T = 2000$), it is not so disadvantageous compared to our methods.

\begin{table}[H]
	\centering
	\caption{The FPR and TPR of $\hbS$, as well as relative errors of $\hbC$ and $\hbPi$ for settings with varying dimensions and time lengths. The mean and standard errors (in parentheses), based on 200 repetitions, are reported. The best and second-best are marked in  {\color{bluep}\bf purple} and {\bf bold},  respectively.}
	\setlength{\tabcolsep}{1mm}{
		\begin{tabular}{*{12}{c}}
			\hline 
			\vspace{-0.15in} \\
			\multicolumn{4}{c}{$\qquad \qquad \qquad \|\hbC - \bC_{0}\|_F/\|\bC_{0}\|_F$}& \multicolumn{2}{c}{ FPR}& \multicolumn{3}{c}{TPR}  \\ 
			$(k,n)$ & $T$ & QMLE & BC & QMLE & BC & QMLE & BC \\
			
			
			&100&\pinbf{0.269}(0.110)&$\bf{0.701}$(0.347)&$\bf{0.197}$(0.041)&\pinbf{0.157}(0.032)&\pinbf{0.698}(0.102)&$\bf{0.686}$(0.098)&\\
			(3,4)&500&\pinbf{0.165}(0.058)&$\bf{0.311}$(0.116)&$\bf{0.121}$(0.037)&\pinbf{0.058}(0.019)&\pinbf{0.792}(0.071)&$\bf{0.787}$(0.057)&\\
			&1000&\pinbf{0.192}(0.083)&$\bf{0.228}$(0.071)&$\bf{0.152}$(0.033)&\pinbf{0.060}(0.019)&\pinbf{0.845}(0.053)&$\bf{0.829}$(0.049)&\\\bigskip
			&2000&\pinbf{0.157}(0.024)&$\bf{0.187}$(0.044)&$\bf{0.145}$(0.030)&\pinbf{0.050}(0.018)&\pinbf{0.846}(0.050)&$\bf{0.826}$(0.044)&\\

%
&100&$\bf{0.381}$(0.074)&\pinbf{0.314}(0.094)&$\bf{0.210}$(0.018)&\pinbf{0.178}(0.018)&$\bf{0.684}$(0.079)&\pinbf{0.704}(0.081)&\\
(4,6)&500&$\bf{0.368}$(0.075)&\pinbf{0.151}(0.040)&$\bf{0.198}$(0.015)&\pinbf{0.138}(0.014)&$\bf{0.933}$(0.041)&\pinbf{0.991}(0.018)&\\

&1000&$\bf{0.296}$(0.048)&\pinbf{0.119}(0.031)&$\bf{0.190}$(0.011)&\pinbf{0.075}(0.011)&$\bf{0.988}$(0.018)&\pinbf{0.999}(0.005)&\\\bigskip
&2000&$\bf{0.287}$(0.036)&\pinbf{0.102}(0.023)&$\bf{0.191}$(0.011)&\pinbf{0.039}(0.008)&$\bf{0.997}$(0.010)&\pinbf{1.000}(0.002)&\\

			
&100&$\bf{0.483}$(0.063)&\pinbf{0.381}(0.103)&\pinbf{0.097}(0.007)&$\bf{0.149}$(0.008)&$\bf{0.677}$(0.088)&\pinbf{0.739}(0.077)&\\
(5,10)&500&$\bf{0.422}$(0.024)&\pinbf{0.167}(0.052)&$\bf{0.064}$(0.005)&\pinbf{0.058}(0.005)&$\bf{0.991}$(0.017)&\pinbf{0.993}(0.014)&\\
&1000&$\bf{0.410}$(0.045)&\pinbf{0.111}(0.023)&$\bf{0.082}$(0.005)&\pinbf{0.067}(0.005)&\pinbf{1.000}(0.000)&$\pinbf{1.000}$(0.000)&\\\bigskip
&2000&$\bf{0.416}$(0.021)&\pinbf{0.080}(0.018)&$\bf{0.077}$(0.005)&\pinbf{0.050}(0.004)&\pinbf{1.000}(0.000)&$\pinbf{1.000}$(0.000)&\\

			\multicolumn{8}{c}{$ \|\hbPi - \bPi_{0}\|_F/\|\bPi_0\|_F$}&  \\
			$(k,n)$ & $T$ & sVAR  & MAR  & QMLE & BC \\
			&100&0.335(0.036)&0.210(0.010)&\pinbf{0.138}(0.016)&$\bf{0.153}$(0.051)&\\
			(3,4)&500&0.141(0.014)&0.197(0.004)&$\bf{0.089}$(0.008)&\pinbf{0.081}(0.007)&\\
			&1000&0.099(0.010)&0.195(0.003)&$\bf{0.079}$(0.008)&\pinbf{0.065}(0.005)&\\\bigskip
			&2000&$\bf{0.069}$(0.006)&0.195(0.002)&0.073(0.004)&\pinbf{0.059}(0.004)&\\


&100&0.594(0.041)&0.602(0.023)&$\bf{0.309}$(0.035)&\pinbf{0.241}(0.035)&\\
(4,6)&500&0.232(0.014)&0.583(0.007)&$\bf{0.162}$(0.012)&\pinbf{0.114}(0.010)&\\
&1000&0.162(0.009)&0.580(0.005)&$\bf{0.144}$(0.039)&\pinbf{0.090}(0.008)&\\\bigskip
&2000&$\bf{0.114}$(0.007)&0.580(0.004)&0.125(0.030)&\pinbf{0.078}(0.005)&\\

			
&100&0.927(0.039)&0.379(0.010)&$\bf{0.236}$(0.022)&\pinbf{0.219}(0.028)&\\
(5,10)&500&0.301(0.008)&0.355(0.004)&$\bf{0.139}$(0.009)&\pinbf{0.089}(0.010)&\\
&1000&0.207(0.006)&0.352(0.003)&$\bf{0.123}$(0.013)&\pinbf{0.062}(0.004)&\\\bigskip
&2000&0.144(0.004)&0.351(0.002)&$\bf{0.119}$(0.007)&\pinbf{0.044}(0.003)&\\
			\hline
		\end{tabular}
	}
	\label{table:sim-1}
\end{table}

\section{Real data analysis}\label{sec:real}
In this section, we conduct real data analysis to demonstrate the superiority of our model and effectiveness of our method. 
We consider logarithm of the following five quarterly state variables: consumer price index $\text{CPI}_{i t}$ of country $i$ in the $t$-th quarter, real (CPI-adjusted) nominal $\text{GDP}_{i t}$,  long- and short-term nominal interest rate $\text{LR}_{it}$ and $\text{SR}_{it}$. They are the main variables of interest in the fundamental equations of dominating New Keyesian models in the macroeconomics literature. Finally, we also include real (CPI-adjusted) $\text{EQ}_{it}$ as the real equity price index, as forward-looking asset prices are potentially useful to predict output and inflation.\footnote{We thank \cite{GVARdata} for making their data public. We have also updated trade weights after 2016.} 

We focus our analysis on ten industrialized countries and regions: Australia (AUS), Canada (CAN), the Euro Area (EUR), Japan (JPN), Norway (NOR), New Zealand (NZL), Sweden (SWE), Switzerland (CHE), the United Kingdom (GBR), and the United States of America (USA).\footnote{The Euro Area includes eight countries: Austria, Belgium, Finland, France, Germany, Italy, the Netherlands, and Spain.} Consequently, in this analysis, the dimension of $\bX_t$ is $(k,n) = (5,10)$. The entire sample period spans from the first quarter of 1979 (1979Q1) to the fourth quarter of 2019 (2019Q4).
Before fitting the models, we perform the following preprocessing steps:
We take the first-order difference to ensure stationarity.
We demean each univariate time series for every variable from every country.
To make all five variables comparable, we normalize the variance of each variable across all countries and all quarters to 1. We refer to the transferred variables as GDP growth ($\Delta GDP$), inflation ($\Delta CPI$), capital gain ($\Delta EQ$) and change in short rate and long rate ($\Delta SR$ and $\Delta LR$).

The original observed trade network data are not static over time; they are recorded annually. To construct a fixed and meaningful $\bW$ matrix in our model (\ref{eq:model-sigmar}), the trade flow matrix for each year is first row-normalized to have a row sum of 1, with diagonal entries being zero. For in-sample estimation based on data from $t=1,\ldots, T$, and for out-of-sample forecasting for time $T+1$, the matrix $\bW$ is calculated as the average of the row-normalized trade flows over the most recent available three-year period up to time $T$. For example, when the in-sample period is from 1979Q1 to 2008Q3, and trade flow data are not available until 2008Q4, $\bW$ is computed as the average of the trade flows from 2005 to 2007. This method of constructing $\bW$ is commonly adopted in the GVAR literature.

\begin{figure}
    \centering
    \includegraphics[width=1\linewidth]
    {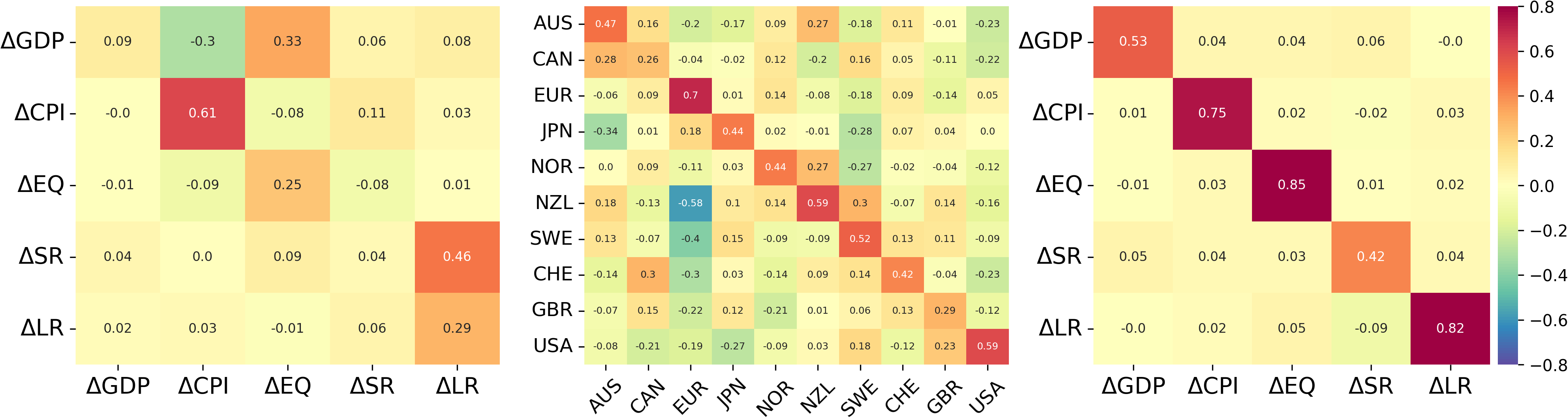}
    \caption{Heatmaps of the estimated coefficient matrices via BC for the period 1979Q1 to 2019Q4. From left to right: $\hbA$, $\hbB$, and $\hbC$.}
    \label{fig:hmap-abc}
\end{figure}

\begin{figure}
	\centering
	\includegraphics[width=1\columnwidth]
    {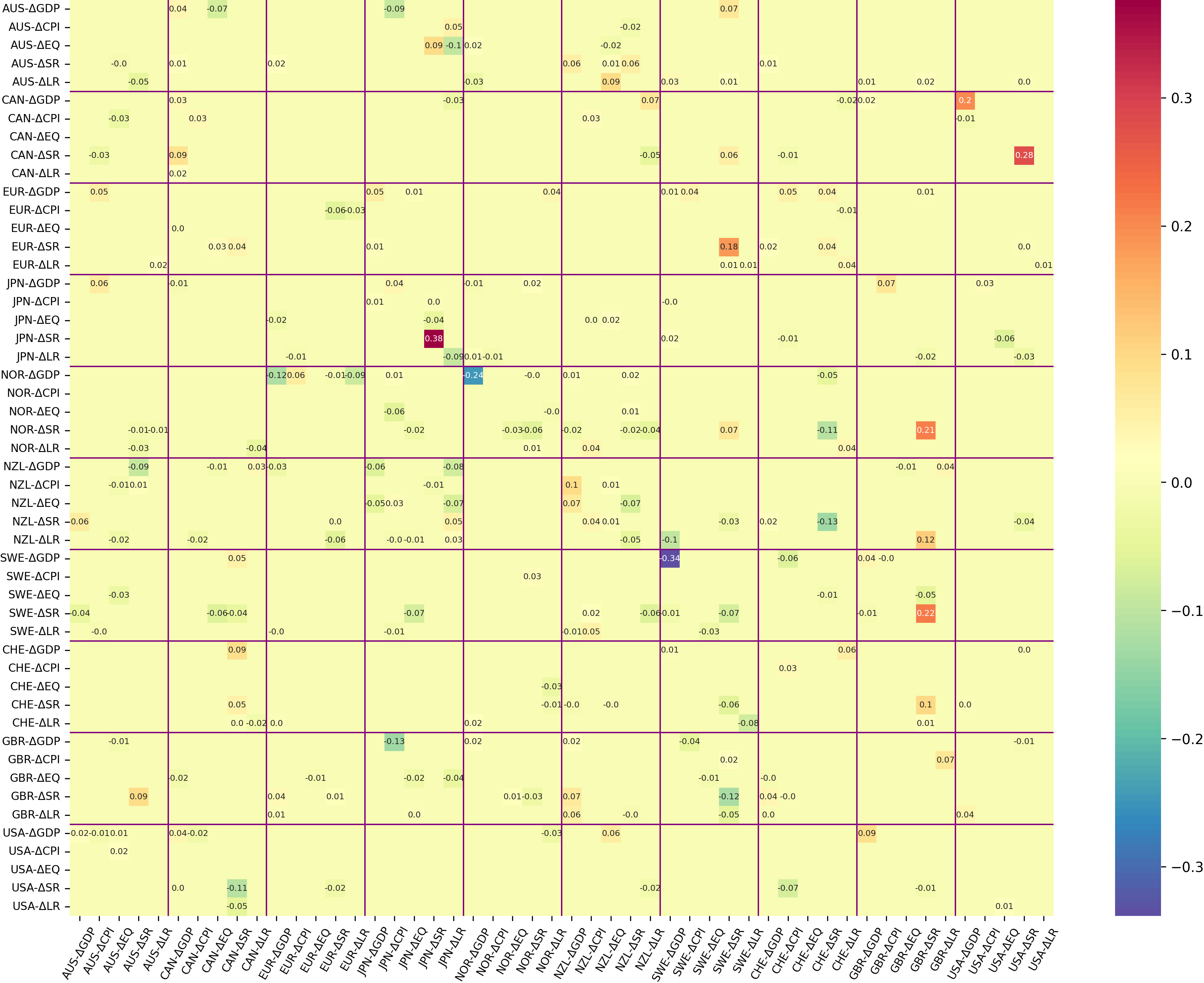}
	\caption{Heatmap of the estimated coefficient matrix $\hbS$ via BC for the period 1979Q1 to 2019Q4. All of the entries with no numbers are exactly zeros. }
	\label{fig:heatmp_s1}
\end{figure}

In what follows, we undertake two tasks to investigate the properties of our methodology. First, we fit our SIGMAR model to the entire sample period and use our BC method to obtain coefficient estimates. Figures \ref{fig:hmap-abc} and \ref{fig:heatmp_s1} present the heatmaps of the BC estimates for all unknown coefficients in the SIGMAR model. The results from QMLE are quite similar and are therefore omitted. We then discuss the economic implications of these estimations. Second, we compare the rolling forward forecasting capabilities of our procedure with those of existing methods.

Figure \ref{fig:hmap-abc} reveals a general pattern of the lead-lag and contemporaneous relationship of five state variables across ten countries. We first focus on $\hbC$. Diagonal elements of this matrix are all positive. While the coefficient estimates are only 0.42 for short rate change and 0.53 for GDP growth, they are much larger for the other variables: 0.75 for inflation, 0.85 for capital gain, and 0.82 for long rate change. In comparison, off-diagonal elements have much smaller magnitude, ranging from $-$0.09 to 0.06. These patterns suggest that contemporaneously these variables are strongly affected by their country-aggregate counterparts, but rather weakly by other variables. 

While the contemporaneous relationship from the trade network is summarized in $\hbC$, the joint lead-lag dynamics of all economic variables across all countries are described by $\hbA$ and $\hbB$. Recall that although $\bB \otimes \bA$ is identifiable, $\bA$ and $\bB$ are only identifiable up to sign and scaling. To make them fully determined, we impose constraints that the Frobenius norm of $\hbA$ is one and that most of the diagonal entries of these two matrices are positive. The heatmaps of $\hbA$ and $\hbB$ in Figure \ref{fig:hmap-abc} satisfy these constraints.
Overall, these variables can all be positively predicted by their own lags, as indicated by the positivity of all the diagonal entries of $\hbA$ and $\hbB$. In particular, we find that inflation has the largest autoregressive coefficient of 0.61, while the short rate appears to be weakly predictable by its past values.

We first examine the predictability of GDP growth through the first row of the heatmap of $\hbA$. Other than the lagged GDP growth, we also find that real capital gain strongly predicts higher real GDP growth. Stock market price has long been known to be a leading indicator of economic activity in the U.S. \citep{Fama_1990JF}, though the literature has found weaker predictive power of stock market return (with dividend), especially after controlling for lagged output growth \footnote{For a comprehensive review, see \citep{StockWatson_2003JEL}}. Our findings suggest if we use a variation of return, i.e., without dividend, some novel evidence in support of predictability across the major OECD developed countries would emerge. We also find that GDP growth can be strongly negatively predicted by inflation. Short rate, in contrast, seems to have weak predictive power. Although it is well known that in the U.S., Fed raises (cuts) rates to cool (stimulate) economic growth, our finding is still consistent with the literature that change in short rate has little marginal predictive content after controlling for other predictors. As we also find that the predictive power of long rate is not strong either, combined, change in term spread (the difference between these two variables), would not be able to strongly predict GDP growth, consistent with the literature on the U.S. GDP predictability \citep{AngPiazzesiWei_2005JoE}. 

Turning to inflation (the second row of the heatmap of $\hbA$), we find that its autoregressive coefficient is as high as 0.61, implying that it is likely to be always economically large after multiplying diagonal elements of matrix $\hbB$. Literature agrees that there is little or no marginal information content in the nominal interest rate term structure for future inflation. However, we find that in real terms, increase in short rate still leads to a higher inflation. Next, turning to capital gain (the third row of the heatmap of $\hbA$), we also find that it can only be weakly predicted by lower inflation and short rate, which is reminiscent of fragile evidence of in-sample return predictability by short term yield, long term yield, and inflation, as in \cite{GoyalWelch_2008RFS}. 

Finally, regarding the two real term structure variables (the last two rows of the heatmap of $\hbA$), long rate strongly predicts short rate with a coefficient of 0.46, while short rate predicts long rate only with a coefficient of 0.06. Combining the weak autocorrelation of short rate itself, this again suggests that the term spread should be able to predict short rate generally. We also do not detect predictability by other variables, though the literature suggests otherwise.\footnote{\cite{Monench_2008JoE} and \cite{DieboldRudebuschArouba_2006JoE} provide some evidence of yield curve predictability with more complicated econometric models.}

The middle panel of Figure \ref{fig:hmap-abc} reports the lead-lag relationship 
across different countries that holds \emph{systematically for all} economic variables. 
The diagonal of the matrix $\hbB$ shows all positive values, suggesting that the general pattern of $\hbA$ exists everywhere when predicting domestic variables, albeit with significant cross-country differences. The pattern is most significant in Euro Area while substantially attenuated in Canada. Turning to off-diagonals, we observe sharp differences in the cross-predictability. For example, U.S. variable cannot predict Japan variables systematically and only weakly Euro Area variables, which is somewhat surprising given the significant role of U.S. in the world economy. 
Similarly, Norway and Switzerland variables are not very useful in predicting other countries either. In contrast, Canada variable predicts higher variable values of Switzerland, and Sweden variable also positively predicts Newzealand. Conversely, Euro Area variable values significantly predicts lower values of New Zealand, Switzerland, and Sweden. In fact, most of the estimates (8 out of 9) for Euro area are negative. 

Parsimony in our model is achieved through the Kronecker product of $\hbA$ and $\hbB$. Nevertheless, any uncaptured marginal cross-predictability can still be recovered via our sparse estimation of the residual matrix, $\hbS$, as illustrated in the heatmap in Figure \ref{fig:heatmp_s1}. The key insights from the heatmap become more apparent after re-ordering the rows and columns to cluster the same variables from different countries together, and by focusing on the entries with large magnitudes. Such an alternative version of heatmap of $\hbS$ can be found as Figure \syb{S1}, and the corresponding chord diagram can be found as Figure \syb{S2}, both in the appendix. 
We have several interesting findings.

First, we document ample residual cross-predictability $\hbS$ beyond the general pattern described by $\hbB\otimes \hbA$. For example, there are most lead-lag relationships about change in short rate. Namely, U.S., U.K., Switzerland, Sweden, Canada short rates lead that of Canada, Norway and Sweden, Norway and New Zealand, U.K. and Euro Area, U.S. short rates, respectively. And Sweden and U.K. are the countries that both lead and lag other countries in change of short rate. This evidence implies a sequential cooperation of monetary policy across these OECD countries. In particular, although in general short rate cannot predict itself systematically for all countries as shown in $\hbA$, Japan makes a strong exception with an estimated coefficient of 0.38. 
Secondly, perhaps due to geographic adjacency, U.S. GDP growth positively predicts that of Canada, while U.S. short rate also leads that of Canada. More interestingly, we find that the positive residual GDP growth and short rate cross-predictability from U.S. to Canada as described in $\hbS$ is dominating the systematic cross-predictability such that the negative sign reported in $\hbB$ can be completely reversed. In fact, we continue to detect this pattern in other cases, and identify at least seven of reversed sign with the magnitude of residual cross-predictability above 0.1 (see Figure \syb{S3} in the appendix for the plot). This observation highlights the importance of estimating $\hbS$, as $\hbA$ and $\hbB$ may fail to capture significant residual cross-predictability.  Thirdly, Australia seems quite a loner--there is only weak association with other countries. In other words, the general pattern summarized by $\hbA$ and $\hbB$ describe Australia quite well. Finally, change in stock price is hardly associated with other variables in terms of the residual cross-predictability.  

We now summarize our findings. Time-series movements of changes in long-term risky assets price, i.e., stock price and long rate of a single country, seem to be more associated with their contemporaneous aggregate counterparts, while not much predicted by other variables, nor by their past values of a different country. Such weak leag-lag relationships can be found in Figure \syb{S4} of the appendix, which provides the heatmap of total cross-predictability $\hbB \otimes \hbA +\hbS$. Moreover, if we interpret that the bulk of contemporaneous cross-sectional variation of stock price or long rate is each approximately captured by a static single-factor model, our finding may be reminiscent of an International CAPM under the financial market integration across countries. Still, there is a major difference. In factor models such as CAPM, the factor (world market) is the same for all countries. In our model, however, each country has its own unique world market factor, distinct from those of other countries.

In contrast, short rate of each country is only weakly associated with their aggregated counterpart concurrently, nor by its own lags (with only exception of Japan). However, there is a strong lead-lag relationship of short rates across countries. Furthermore, short rates are strongly predicted by long rates of all countries, in particular, high long rates of the same country always lead to high short rates. The joint evidence from interest rate term structure implies a cooperative yet nonsynchronous global monetary policy. Finally, the case for inflation and GDP growth is more ambiguous. Other than the effect of global counterpart and their own lags, there is also some evidence for cross-predictability by other variables from other countries.

To measure the forecasting performance of the model, we compute the mean squared forecast error (MSFE) for the $i$-th economic variable as follows
\bel{def:rmsfe}
\text{MSFE}(i,T_{test}) =  {T_{test}^{-1} \sum_{t = T}^{T+T_{test}-1} \frac{1}{n} \cdot  \|\bX_{t+1}[i] - \hbX_{t+1|t}[i]\|_2^2}, \quad i = 1,\cdots, k, 
\eel
where $T_{test}$ denotes the number of times the model is fitted during the rolling forecasting, $\bX_{t+1}[i]$ refers to the $i$-th row of $\bX_{t+1}$ for all countries, and  $\|\bX_{t+1}[i] - \hbX_{t+1|t}[i]\|_2^2$ is defined as the forecast error in terms of ground truth $\bX_{t+1}[i]$ and the corresponding one-quarter-ahead prediction $\hbX_{t+1|t}[i]$ conditional on the information up to time $t$.

We then compare SIGMAR with stacked VAR (sVAR),  individual country's VAR (iVAR), individual country's VAR with star variables (iVARX),  univariate AR (iAR), and matrix autoregression (MAR). Particularly, the forecasting with the iVARX involves two steps. First, the coefficients for the country-specific iVARX model in (\ref{eq:model-igvar}) are estimated. Second, the joint iVARX model for all countries is constructed as in (\ref{eq:model-gvar}), which could be utilized for forecasting after plugging in the estimated coefficients. In addition, we explore two variants of our SIGMAR model: one variant excludes the contemporaneous global term but keeps the sparse term, referred to as SMAR; the other variant does not incorporate the sparse term but keeps the contemporaneous global term as in Model (\ref{eq:model-gmar}), referred to as GMAR. Finally, the QMLE and BC estimates for the SIGMAR model are also included.
The in-sample rolling window has a fixed length of 120 quarters, which starts from 1979Q1-2008Q4. The out-of-sample testing period spans 2009Q1 to 2019Q4. 

Table \ref{table:real} reports the MSFEs of the nine methods mentioned above. Methods with the least, second least, and third least MSFEs are highlighted in \redbf{red}, {\color{bluep} \bf bluish-purple}, and {\bf bold}, respectively. We first note that iAR model typically performs the best, with the smallest MSFE for 4 out of 5 variables, except for $\Delta GDP$. This is not surprising as iAR is the simplest model and it is documented in the literature before that the forecasting performance of iAR could be the best for many applications. However, since iAR model assumes each variable from each country is only related to the lagged values of itself in the same country, it is completely silent on the state of the entire economy, not to mention the interconnectedness with other economies. 
In contrast, SIGMAR carefully models the joint dynamics of all economic variables across all countries while maintaining a parsimony with a relatively small number of model parameters. For example, SIGMAR (QMLE) generates the smallest MSFE of 0.8 for $\Delta GDP$, and SIGMAR (BC) produces a similar MSFE of 0.801. In contrast, iAR model delivers a much larger MSFE of 0.836. Comparing other variables where iAR takes the first place, SIGMAR (QMLE)  ranks the second in $\Delta CPI$ and the third in $\Delta LR$, while SIGMAR (BC) also ranks the third in $\Delta CPI$. The other variations of MAR or SIGMAR only fare slightly worse: MAR ranks the second in $\Delta SR$ and the third in $\Delta EQ$, and GMAR ranks the second in $\Delta LR$ and the third in $\Delta GDP$ and $\Delta SR$. Finally, sVAR, iVAR and iVARX models are overshadowed in these horse races with much larger MSFEs. The only exception is that iVAR ranks the second in predicting $\Delta EQ$. 

Taken together, SIGMAR models achieve a better trade-off by still dwindling forecast errors, yet offering useful insights about the structure of the global economy. Indeed, we find that the general pattern of these rolling window estimates is fairly similar to that of the whole sample, as described in the last few paragraphs. 

\begin{table}
\centering
\begin{adjustbox}{max width=\textwidth}
\begin{tabular}{cccccccccc}
\toprule
    & \textbf{sVAR}      & \textbf{iVAR}     & \textbf{iVARX}    & \textbf{iAR}    & \textbf{MAR} & \textbf{SMAR} & \textbf{GMAR} & \textbf{SIGMAR(QMLE)} & \textbf{SIGMAR(BC)} \\ 
    \midrule
$\Delta$\textbf{GDP}  & 1.287(0.015) & 0.803(0.010) & 0.839(0.010) & 0.836(0.014) & 0.847(0.010) & 0.820(0.009) & \blackbf{0.814(0.011)} & \redbf{0.800(0.011)}  & \pinbf{0.801(0.010)} \\ 
$\Delta$\textbf{CPI}  & 0.880(0.009) & 0.337(0.003) & 0.382(0.005) & \redbf{0.282(0.003)} & 0.333(0.004) & 0.331(0.004) & 0.367(0.005) & \pinbf{0.323(0.004)} & \blackbf{0.326(0.005)} \\ 
$\Delta$\textbf{EQ} & 1.557(0.028) & \pinbf{0.640(0.011)} & 1.553(0.044) & \redbf{0.601(0.010)} & \blackbf{0.671(0.012)} & 0.692(0.012) & 0.842(0.019) & 0.762(0.014) & 0.756(0.013) \\ 
$\Delta$\textbf{SR}  & 0.599(0.018) & 0.151(0.004) & 0.250(0.008) & \redbf{0.091(0.004)} & \pinbf{0.094(0.003)} & 0.142(0.004) & \blackbf{0.117(0.003) }& 0.129(0.004) & 0.131(0.004) \\ 
$\Delta$\textbf{LR} & 1.016(0.017) & 0.449(0.008) & 0.687(0.011) & \redbf{0.417(0.008)} & 0.470(0.009) & 0.533(0.011) & \pinbf{0.422(0.009)} & \blackbf{0.436(0.008)} & 0.443(0.008) \\ 
\bottomrule
\end{tabular}
\end{adjustbox}
\caption{The rolling-forward MSFEs for five economic variables and their standard errors by nine estimators. The best, second-best, and third-best are highlighted in \redbf{red}, {\color{bluep} \bf bluish-purple}, and {\bf bold}, respectively.}
\label{table:real}
\end{table}

\section{Conclusion}

In this paper, we propose an innovative approach to modeling matrix-valued time series: SIGMAR. It incorporates the impact of the trade network in studying the contemporaneous dependencies of fundamental economic variables across countries, and also parsimoniously decomposes their lead-lag relationships to a systematic and an idiosyncratic cross-predictability.  We propose both a likelihood estimator and a bias-corrected iterative estimator. Our empirical work involves a dataset of 10 OECD developed countries and regions from 1979Q1 to 2019Q4. Our model offers useful insights into the structure of global economy, and also delivers the smallest forecast errors in comparison with competing VAR models.

We see several extensions of our work. First, the model could introduce more lagged domestic and foreign variables, allowing for more intricate temporal dynamics. Second, multi-relational networks could be utilized simultaneously to capture a wider array of interactions across different types of dependencies. Third, when the number of economic variables or countries becomes even larger, we can further impose sparse structures on the parameter matrices to facilitate dimensionality reduction. Lastly, while our current model only considers stationary variables, a natural extension would be to further incorporate cointegrating relationship in studying the long-run equilibrium structure of the global economy.

\section*{Acknowledgments}
We gratefully acknowledge the support from the United States NSF Grant IIS-1741390 (Yang), the GRF sponsored by the RGC in Hong Kong, No. 17301620 (Yang),  No. 17301123 (Feng), No. 21313922 (Feng) and Hong Kong CRF C7162-20G (Yang).

\newpage
\bibliography{sigmarRef}
\bibliographystyle{ims}
\end{document}


  \begin{center}
    {\LARGE\bf Supplementary Materials to  \\``Sparsity-Induced Global Matrix Autoregressive Model with Auxiliary Network Data''}
\end{center}

\noindent In the supplementary material, we first 
provide  the implementation steps for the convex optimization problem \eqref{opt:rpca} of the projection method in \ref{supp:proj-implement} and present the specific conditions used in Proposition \ref{prop:rpca} in \ref{supp:proj-proof}. Next, we outline the detailed procedure for deriving the bias-correction estimator in Section \ref{supp:bc}. Then, we show some additional results of real data  analysis in \ref{supp:realdata}. Afterwards, we define two rearrangement operators in \ref{supp-sec-1}.  Following this, we present the detailed proofs for Theorem \ref{thm:1}, Proposition \ref{prop:2} and Theorem \ref{thm:2} in \ref{supp:proof-thms}. Furthermore, we show some theoretical properties related to  QMLE and two lemmas used in the preceding proofs in \ref{supp:proof-related}.

\setcounter{equation}{0}
\def\theequation{S\arabic{equation}}
\setcounter{figure}{0}
\def\thefigure{S\arabic{figure}}
\setcounter{table}{0}
\def\thetable{S\arabic{table}}
\setcounter{section}{0}
\def\thesection{S\arabic{section}}

\section{Implementation of projection}\label{supp:proj-implement}
The augmented Lagrangian function of (\ref{opt:rpca}) is 
\bel{ALM}
\ell(\bL, \tbS, \bY) = \|\bL\|_{*} + \lambda \| \tvec(\tbS)\|_{1} + \langle \bY, \tbPhi - \bL -\tbS \rangle + \frac{\mu}{2}\|\tbPhi - \bL -\tbS \|_F^2,
\eel
where $\bY \in \bbR^{n^2 \times k^2}$ is the multiplier of the linear constraint; $\mu >0$ is a penalty parameter for the violation of the linear constraint. 

Define  $\calS_{\tau} : \bbR \rightarrow \bbR$ as the shrinkage operator $\calS_{\tau}(x) = \text{sgn}(x) \max (|x| - \tau,0)$, and extend it to matrices by applying it to each element. It is easy to show that 
\bel{alm:1}
\argmin_{\tbS} \ell(\bL, \tbS, \bY) = \calS_{\lambda \mu^{-1}} (\tbPhi - \bL  + \mu^{-1}\bY ).
\eel
Similarly, for matrix $\bM$, we introduce the soft-thresholding operator $\calD_{\tau}$ defined as $\calD_{\tau}(\bM) = \bU \calS_{\tau}(\bSigma)\bV^\T$, where $\bM = \bU \bSigma \bV^\T$ is any singular value decomposition. According to Theorem 2.1 in \citet{cai2010singular}, we have 
\bel{alm:2}
\argmin_{\bL} \ell(\bL, \tbS, \bY) = \calD_{\mu^{-1}} (\tbPhi - \tbS + \mu^{-1}\bY ).
\eel
Thus, the alternating directions algorithm \citep{yuan2009sparse} for (\ref{ALM})  is summarized as follows.
\begin{algorithm}[H]
	\renewcommand{\algorithmicrequire}{\textbf{Input:}}
	\renewcommand{\algorithmicensure}{\textbf{Output:}}
	\caption{Alternating Directions for (\ref{ALM})}
	\label{alg:2}
	\begin{algorithmic}[1]
		\STATE {\textbf{Initialization:}  $\tbS^{(0)} = \bY^{(0)}  = \mathbf{0}_{kn}$.}
		\STATE {\textbf{While:} \text{not converged} \textbf{do}\\
        \quad \text{compute} $\bL^{(j+1)} = \calD_{\mu^{-1}} (\tbPhi - \tbS^{(j)} + \mu^{-1}\bY^{(j)} )$ \\
        \quad \text{compute} $\tbS^{(j+1)} = \calS_{\lambda \mu^{-1}} (\tbPhi - \bL^{(j+1)}  + \mu^{-1}\bY^{(j)} )
        $ \\
        \quad \text{compute} $\bY^{(j+1)} = \bY^{(j)} + \mu (\tbPhi - \bL^{(j+1)} -\tbS^{(j+1)})$ }
        \STATE{\textbf{end while}}
        
		\STATE \textbf{return: } $\bL^{proj}$, $\tbS^{proj}$
	\end{algorithmic}  
\end{algorithm}
Specifically, we simply choose $\lambda = 1/\sqrt{kn}$ and $\mu = k^2n^2/4\|\tvec(\tbPhi)\|_{1}$, as suggested by \citet{yuan2009sparse}. As a consequence, we can obtain $\tvec(\hbA^{mle})$ by taking the top right singular  vector from $\bL^{proj}$, and $\tvec(\hbB^{mle})$ is calculated as the  top singular value multiplied by the top left singular vector. In addition, $\hbS^{mle} = \mR^{-1}(\tbS^{proj})$.
\section{Regularity conditions in Proposition \ref{prop:rpca}}\label{supp:proj-proof}
To begin, it should be noted that the true matrices $\bA_0$ and $\bB_0$ are subject to sign change; specifically, $(-\bA_0,-\bB_0)$ also satisfies $(-\bB_0) \otimes (-\bA_0) = \bB_0 \otimes \bA_0$. For convenience, we define $\|\hbB^{mle} - \bB_{0} \|_F^2$ and $\| \hbA^{mle} - \bA_{0}\|_F^2$ to the smallest corresponding errors, i.e., $\| \hbB^{mle} - \bB_{0}\|_F^2 \coloneq \|\hbB^{mle}\|_F^2 + \|\bB_0\|_F^2  - 2|\langle\hbB^{mle}  , \bB_0\rangle|$ and $\| \hbA^{mle}- \bA_{0}\|_F^2 \coloneq \|\hbA^{mle}\|_F^2  + \|\bA_0\|_F^2  - 2|\langle\hbA^{mle}  , \bA_0\rangle|$.

We first define a quantity to measure the maximum number of non-zero entries in any row or column of $\tbS_{0}$ as follows: 
\bel{eq:alpha}
\alpha(\mu) \coloneq \max\{\mu \|\text{sgn}(\tbS_{0})\|_1, \mu^{-1}\|\text{sgn}(\tbS_{0})\|_{\infty}\},
\eel
where $\sgn(\cdot)$ denotes the sign function applied element-wise to the matrix and $\mu>0$ is a balancing parameter to accommodate disparity between the number of rows and columns. Next, we define $\beta(\mu)$ to measure the sparseness of  $\tvec(\bbB_0)$ and $\tvec(\bA_0)$, where $\tvec(\bbB_0) \coloneq \tvec(\bB_0)/\|\bB_0\|_F$,
\bel{eq:beta}
\beta(\mu) \coloneq \mu^{-1}\|\tvec(\tvec(\bbB_{0})\tvec(\bbB_{0})^\T)\|_{\infty} + \mu\|\tvec(\tvec(\bA_{0})\tvec(\bA_{0})^\T)\|_{\infty} + \|\tvec(\bbB_{0})\|_{\infty}\|\tvec(\bA_{0})\|_{\infty}.
\eel
	Assume that $\alpha(\mu)$ and $\beta(\mu)$, as defined in (\ref{eq:alpha}) and (\ref{eq:beta}), comply with  
    \bel{eq:theory-8}
\alpha(\mu) \cdot \|\tvec(\tvec(\bbB_{0})\tvec(\bA_{0})^\T)\|_{\infty} \leq \frac{1}{41}	\text{ and }  \alpha(\mu) \beta(\mu) \leq \frac{3}{41},
	\eel
 for some $\mu >0$. By setting $\lambda = (15/82)/\alpha(\mu)$, the $\hbB^{mle}, \hbA^{mle}$,  and $\hbS^{mle}$  extracted from the solution to the convex program (\ref{opt:rpca}) satisfies 	\bel{eq:theory-9-supp}
	\|\hbB^{mle} - \bB_{0} \|_F + \| \hbA^{mle} - \bA_{0}\|_F + \|\hbS^{mle} - \bS_{0}\|_F  = O_{p}(1/\sqrt{T}).
	\eel

The proof of equation (\ref{eq:theory-9-supp}) can be found in the proof of Theorem 3 in \citet{hsu2011robust}.
\section{Estimation enhancement via Bias Correction}\label{supp:bc}
Recall that $\bPhi_0\coloneq \bB_0 \otimes \bA_0 + \bS_0$ and our model is
\bel{eq:bc-1}
\bX_t=   \bC_{0}\bX_{t} \bW^\T+ \tvec^{-1}\left[\bPhi_0\tvec(\bX_{t-1})\right]+ \bE_t.
\eel
Given $\hbPhi$, the updating of $\bC$ can be divided into two steps, the first step is to  derive $\hbC^{lse}$ directly by solving the following least square problem:
\bel{obj:C0_lse}
\hbC_0^{lse} \coloneq \argmin_{\bC} \frac{1}{T}\sum_{t=1}^{T}  \|{\bX}_t^c  - \bC_0\bX_{t} \bW^\T\|_F^2,
\eel
where 
\bel{eq:bc-2}
{\bX}_t^c && \coloneq \bX_t - \tvec^{-1}\left[\hbPhi\tvec(\bX_{t-1})\right] \cr
&&= \bC_{0}\bX_{t} \bW^\T+ \bE_t + \tvec^{-1}\left[(\bPhi_{0} - \hbPhi_0)\tvec(\bX_{t-1})\right].
\eel
The solution to (\ref{obj:C0_lse}) has a closed form and can be expressed as follows:
\bel{eq:bc-3-supp}
\hbC_0^{lse} && = \frac{1}{T}\sum_{t=1}^{T} {\bX}_t^c \bW \bX_t^\T \left( \frac{1}{T}\sum_{t=1}^{T}\bX_t \bW^\T \bW \bX_t^\T \right)^{-1} \cr
&& = \bC_{0}   +  \frac{1}{T}\sum_{t=1}^{T} \bE_t \bW \bX_t^\T \left( \frac{1}{T}\sum_{t=1}^{T}\bX_t \bW^\T \bW \bX_t^\T \right)^{-1} \cr
&& \qquad + \frac{1}{T}\sum_{t=1}^{T} \tvec^{-1}\left[(\bPhi_{0} - \hbPhi)\tvec(\bX_{t-1})\right]  \bW\bX_t^\T \left( \frac{1}{T}\sum_{t=1}^{T}\bX_t \bW^\T \bW \bX_t^\T \right)^{-1}.
\eel
Moreover, denoting 
\bel{eq:bc-4}
\bGamma_w \coloneq   \frac{1}{T}\sum_{t=1}^{T}\bX_t \bW^\T \bW \bX_t^\T,
\eel
and 
\bel{eq:bc-5}
\tvec(\tbE_t) && = \bG_{0}^{-1} \tvec(\bE_t)\cr
&& \overset{\Delta}{=} \tbe_t.
\eel
we have 
\bel{eq:bc-6}
&& \left(\hbC_0^{lse} - \bC_{0} \right) \bGamma_w \cr
&& = \frac{1}{T}\sum_{t=1}^{T} \tvec^{-1}\left[(\bPhi_{0} - \hbPhi_1)\tvec(\bX_{t-1})\right]  \bW\left[\tvec^{-1}\left(\bG_{0}^{-1}\bPhi_{0} \tvec(\bX_{t-1}) + \tbe_t \right) \right]^\T \cr
&&  \quad +   \frac{1}{T}\sum_{t=1}^{T} \bE_t \bW \left[\tvec^{-1}\left(\bG_{0}^{-1}\bPhi_{0} \tvec(X_{t-1}) + \tbe_t \right)\right]^\T.
\eel
Take the vectorization operation $\tvec(\cdot)$ on both side, we achieve 
\bel{eq:bc-7}
(\bGamma_w^\T \otimes \bI_{k})\tvec(\hbC_0^{lse} - \bC_{0} ) \!\!\!
&=& \!\! \frac{1}{T}\sum_{t=1}^{T} (\tbE_t \otimes \bE_t)\tvec(\bW)  + {\frac{1}{T}\sum_{t=1}^{T}(\bZ_{t-1} \otimes \bE_t ) \tvec(\bW)}  \cr 
&& \!\!\!\!\!\!\! +\frac{1}{T}\sum_{t=1}^{T}   \!\!\left(\tvec(\bX_{t-1})^\T \! \otimes\!  (\bZ_{t-1}\bW^\T\!\! + \tbE_{t}\bW^\T) \! \otimes  \!\bI_{k} \!\right)\!\tvec(\bPhi_{0} - \hbPhi) ,
\eel
where $\tvec(\bZ_{t-1}) \coloneq \bG_{0}^{-1}\bPhi_{0} \tvec(\bX_{t-1})$.
Moreover, (\ref{eq:bc-5}) implies $\bE_t = \tbE_t - \bC_{0} \tbE_t W^\T$. Thus, we get
\bel{eq:bc-8}
\tbE_t \otimes \bE_t   &&=  \tbE_t \otimes  \tbE_t -  \tbE_t \otimes  \bC_{0} \tbE_t \bW^\T  \cr 
&& = \tbE_t \otimes  \tbE_t 
- (\bI_k \otimes \bC_{0}) (\tbE_t \otimes  \tbE_t) (\bI_n \otimes \bW^\T) .
\eel
Substituting (\ref{eq:bc-8}) to (\ref{eq:bc-7}),  and denoting  ${{\tbSigma}} = \frac{1}{T}\sum_{t=1}^{T} (\tbE_t \otimes \tbE_t)$, we obtain 
\bel{eq:bc-9}
(\bGamma_w^\T \otimes \bI_{k})\tvec(\hbC_0^{lse} - \bC_{0})  &&= 
{ \frac{1}{T}\sum_{t=1}^{T}   \left(\tvec(\bX_{t-1})^\T \otimes  (\bZ_{t-1}\bW^\T)  \otimes  \bI_{k} \right)\tvec(\bPhi_{0} - \hbPhi)} \cr
&& \quad + \left( \tbSigma - (\bI_k \otimes \bC_{0}) \tbSigma (\bI_n \otimes \bW^\T) \right) \tvec(\bW)  + o_p(1) .
\eel
where this equality holds since that every non-zero element in the $\bZ_{t-1}\otimes \bE_t$ and $\tvec(\bX_{t-1})^\T \otimes \tbE_{t}$ are both involves the product of $\bX_{t-1}$ and $\bE_{t}$, which is assumed to  be  0 due to the independence  of $\bX_{t-1}$ and $\bE_{t}$. Moreover, for simplicity, let
$\tvec(\tbSigma_w) \coloneq \tbSigma \tvec(\bW)$ and $\tvec(\tbSigma_{w^2})\coloneq  \tbSigma \tvec(\bW^\T \bW) $, rewriting (\ref{eq:bc-9}) as matrix version, we have 
\bel{eq:bc-10-supp}
(\hbC_0^{lse} - \bC_{0})\bGamma_w  &&=   
\tvec^{-1}\left({ \frac{1}{T}\sum_{t=1}^{T}   \left(\tvec(\bX_{t-1})^\T \otimes  (\bZ_{t-1}\bW^\T)  \otimes  \bI_{k} \right)\tvec(\bPhi_{0} - \hbPhi)}\right) \cr
&& \quad + \tbSigma_{w} - \bC_{0}\tbSigma_{w^2}  + o_p(1).
\eel
Since ${\bGamma}_w$, ${\tbSigma}$ are consistent estimators of $\bbE\bGamma_w$ and $\bbE\tbSigma$ respectively, we can observe that  (\ref{eq:bc-10-supp}) is only dependent on $\hbC_0^{lse}$ and $\hbPhi$, hence  the bias correction estimator $ \bC_{0}^{bc}$ of $\bC_{0}$, can be derived in one step, that is 
\bel{eq:bc-11}
(\hbC^{lse} - \bC_{0})\bGamma_w  =  \tbSigma_{w} - \bC_{0}\tbSigma_{w^2} + 
\tvec^{-1}\left(\mathbf{Q}_w\tvec(\bPhi_{0} - \hbPhi)\right) + o_p(1).
\eel
where $\bQ_w \coloneq \frac{1}{T}\sum_{t=1}^{T}   \left(\tvec(\bX_{t-1})^\T \otimes  (\bZ_{t-1}\bW^\T)  \otimes  \bI_{k} \right)$.
In addition, we can choose  $\hbA^{mle}, \hbB^{mle}$ and $\hbS^{mle}$ as the initialization  in Algorithm \ref{alg:1}, which together form a consistent estimator of $\bPhi_{0}$. Hence, the  term involves $\hbPhi - \bPhi_0$ in (\ref{eq:bc-11}) is of order $o_{p}(1)$. Accordingly, the estimation for $ \bC_{0}^{bc}$ can be expressed as follows
\bel{eq:bc-12-supp}
\hbC_{0}^{bc}  = (\hbC_0^{lse}\bGamma_w  -  \tbSigma_{w})(\bGamma_w -\tbSigma_{w^2})^{-1}  .
\eel
\section{Additional results of real data  analysis}\label{supp:realdata}
\begin{figure}[H]
	\centering
	\includegraphics[width=0.95\columnwidth]{figures/hmap_S1.png}
	\caption{Heatmap of the estimated coefficient matrix $\hbS$ via BC for the period 1979Q1 to 2019Q4. The rows and columns are re-ordered to cluster the same variables from different countries together. Only the entries with the top 15 largest magnitudes are annotated with numbers.}
	\label{suppfig:s1}
\end{figure}
\begin{figure}[H]
	\centering
	\includegraphics[width=0.95\columnwidth]{figures/chord_S_spec.png}
	\caption{A chord diagram illustrating the largest 15 relationships in $\hbS$ of the BC estimate for the period 1979Q1 to 2019Q4. Each coordinate in the chord diagram corresponds to an economic variable of a specific country. The connections between coordinates are directed,  with the arrowheads  indicating the  direction of the relationship.  {\color{MyRed}\bf{Red}} links represent positive effects, while  {\color{bluep}\bf{bluish-purple}} links indicate negative effects. The thickness of the lines corresponds to the absolute values of $\hbS$.}
	\label{suppfig:chord_s}
\end{figure}

\begin{figure}[H]
	\centering
	\includegraphics[width=0.95\columnwidth]{figures/hmap_Sflip_trans_allS.png}
	\caption{A heatmap of $\hbS^{flip}$, where  $\hbS^{flip}$ is obtained from  $\hbS$ by retaining the values of entries of $\hbS$, where the entries satisfy $\sgn(\hbB \otimes \hbA) \neq \sgn(\hbB \otimes \hbA +\hbS)$, and setting other entries of $\hbS$ to zero. Here, $\sgn(\cdot)$ denotes the sign function. Note that $\hbB, \hbA$ and $\hbS$ are estimated using BC for the period 1979Q1 to 2019Q4. The rows and columns are re-ordered to cluster the same variables from different countries together. }
	\label{suppfig:hmap_s_flip}
\end{figure}

\begin{figure}[H]
	\centering
	\includegraphics[width=0.95\columnwidth]{figures/hmap_BAS_all.png}
	\caption{A heatmap of $\hbB \otimes \hbA +\hbS$ via BC for the period 1979Q1 to 2019Q4.  The rows and columns are re-ordered to cluster the same variables from different countries together.}
	\label{suppfig:hmap_bas}
\end{figure}

\section{Techniques for rearrangement operators}\label{supp-sec-1}
For any matrix $\bM$ that is a $p_1\times p_2$ array of blocks of the same block size $d_1\times d_2$, let $M_{j,k}^{d_1,d_2}$ be the $(j,k)$-th block, $1\le j\le p_1$, $1\le k\le p_2$. Further let the operator $\mR_1: \mathbb{R}^{(p_1d_1)\times (p_2d_2)}\rightarrow \mathbb{R}^{(p_1p_2)\times (d_1d_2)}$ be a mapping such that
\bel{R1:1-supp}
\mR_1(\bM)=\left[\tvec(M_{1,1}^{d_1,d_2}),\ldots, \tvec(M_{p_1,1}^{d_1,d_2}),\ldots, \tvec(M_{1,p_2}^{d_1,d_2}),\ldots,\tvec(M_{p_1,p_2}^{d_1,d_2})\right]^\T.
\eel
When applying the operator $\mR_1$ to a Kronecker product $\bW \otimes \bC$, here $d_1= d_2 = k$ and $p_1= p_2 = n$, it holds that
\bel{R1:2-supp}
\mR_1(\bW \otimes \bC)=\tvec(\bW)[\tvec(\bC)]^\T.
\eel
Note that $\calR_{1}(\cdot)$ is equivalent to $\calR(\cdot)$ as defined in the manuscript.
The property (\ref{R1:2-supp}) would be of great use in our analysis.
Similarly, we define the rearrangement operator $\calR_2: \bbR^{d_1d_2 \times p_1p_2} \rightarrow \bbR^{p_1d_1 \times p_2d_2}$ be a mapping such that
\bel{R2:1}
\mR_2(\bM)=\left[\tvec(\bM_{1,1}^{d_1,p_1}),\ldots, \tvec(\bM_{d_2,1}^{d_1,p_1}),\ldots, \tvec(\bM_{1,p_2}^{d_1,p_1}),\ldots,\tvec(\bM_{d_2,p_2}^{d_1,p_1})\right]^\T.
\eel
Moreover, when applying the operator $\mR_2$ to a Kronecker product $\bE_{t}\otimes \bE_{t}$ (notably, $d_1= d_2 = k$ and $p_1= p_2 = n$), it still holds that
\bel{R2:2}
\mR_2(\bE_{t}\otimes \bE_{t})=\tvec(\bE_{t})[\tvec(\bE_{t})]^\T.
\eel
According to these two definitions, we have the fact that 
\bel{R12:1}
\mR_2[\mR_1(\bW \otimes \bC)^\T]^\T = \bW \otimes \bC.
\eel
That is for any matrix $\bM \in \bbR^{p_1d_1 \times p_2d_2}$, we have
\bel{R12:2}
\mR_1(\bM^\T)^\T = \mR_2^{-1}(\bM).
\eel

\section{Proofs for theorems and propositions}\label{supp:proof-thms}

     
\subsection{Proof of Theorem \ref{thm:1}}\label{supp-proof-1}
\begin{proof}
    The proof of Theorem \ref{thm:1} can be divided into the following three parts:  (i) $T^{-1} \left(\ln\calL(\btheta) - \bbE\ln \calL(\btheta)\right) $ converges  in probability to zero uniformly for $\btheta \in \bTheta$; (ii) $T^{-1}\bbE\calL(\btheta)$ is uniformly  equicontinuous;  and (iii) global identification of $(\bC_0, \bPhi_0, \sigma_{0}^2)$. 
    
    Firstly, we aim to prove
		\bel{pf-part1-1}
		T^{-1} \left(\ln\calL(\btheta) - \bbE\ln \calL(\btheta)\right) = o_{p}(1).
		\eel
	Recall that 
	\bes
	T^{-1} \ln \calL(\btheta) = -\frac{kn}{2}\ln 2\pi - \frac{kn}{2}\ln \sigma^2 +  \ln |\bG| - \frac{1}{2\sigma^2}\frac{1}{T}\sum_{t = 1}^{T} \bv_t^\T\bv_t.
	\ees
	Then  we have
	\bel{pf-part1-2}
	T^{-1} \left(\ln \calL(\btheta) - \bbE\ln \calL(\btheta)\right)  && = 
	- \frac{1}{2\sigma^2}
	\left(\frac{1}{T}\sum_{t = 1}^{T} \bv_t^\T\bv_t  -  \bbE  \bv_t^\T\bv_t \right).
	\eel
	By $\bv_t = \bG_0 \tvec(\bX_t) - \bPhi \tvec(\bX_{t-1})$ and $\be_{t} = \bG_{0} \tvec(\bX_t) - \bPhi_{0} \tvec(\bX_{t-1}) = \tvec(\bE_{t})$, we obtain  
	\bel{pf-part1-3}
	\bv_t = \be_{t}  + (\bG - \bG_{0})\tvec(\bX_t) - (\bPhi - \bPhi_{0})\tvec(\bX_{t-1}),
	\eel
	and  
	\bel{pf-part1-4}
	\bv_t^\T\bv_t \! &=& \! \be_t^\T \be_t + 2\tr((\bG - \bG_{0})\tvec(\bX_t)\be_t^\T) -2 \tr((\bPhi - \bPhi_{0})\tvec(\bX_{t-1}) \be_t^\T) \cr
    &+&  \!\! \!\! \tr((\bG - \bG_{0})\tvec(\bX_t)\tvec(\bX_t)^\T(\bG - \bG_{0})^\T) 
	 + \tr((\bPhi - \bPhi_{0})\tvec(\bX_{t-1}) \tvec(\bX_{t-1}) ^\T(\bPhi - \bPhi_{0})^\T) \cr 
	&-&  \!\! \!\!  2 \tr((\bG - \bG_{0})\tvec(\bX_t)\tvec(\bX_{t-1})^\T  (\bPhi - \bPhi_{0})^\T).
	\eel
	By using the fact that $\be_t = \bG_{0} \tvec(\bX_t) - \bPhi_{0} \tvec(\bX_{t-1})$,  the second and third term in (\ref{pf-part1-4}) can be rewritten as 
	\bel{pf-part1-5}
	\tr((\bG_0 - \bG_{0})\tvec(\bX_t)\be_t^\T) &=& \tr((\bG_0 - \bG_{0})\tvec(\bX_t)\tvec(\bX_t)^\T \bG_{0}^\T) \cr
	&&  - \tr((\bG_0 - \bG_{0})\tvec(\bX_{t})\tvec(\bX_{t-1})^\T \bPhi_{0}^\T),
	\eel
	and 
	\bel{pf-part1-6-1}
	\tr((\bPhi - \bPhi_{0})\tvec(\bX_{t-1}) \be_t^\T)
	&&= \tr((\bPhi - \bPhi_{0})\tvec(\bX_{t-1})\tvec(\bX_t)^\T \bG_{0}^\T) \cr
	&& \quad - \tr((\bPhi - \bPhi_{0})\tvec(\bX_{t-1}) \tvec(\bX_{t-1})^\T \bPhi_{0}^\T).
	\eel
	On the other hand, by the ergodic theorem for the Law of Large Numbers, it follows that 
	\begin{itemize}
		\item[$(i)$] $	\frac{1}{T}\sum_{t = 1}^{T} \tr\left(\be_t^\T \be_t\right)   -  \bbE \tr\left(\be_t^\T \be_t\right)   \overset{p}{\rightarrow} 0$.
        \item[$(ii)$] 	$\frac{1}{T}\sum_{t = 1}^{T} \tr\left(\Delta\bG \tvec(\bX_t)\tvec(\bX_t)^\T (\Delta\bG)^\T\right)    - \bbE\tr\left(\Delta\bG\tvec(\bX_t)\tvec(\bX_t)^\T(\Delta\bG)^\T\right) \overset{p}{\rightarrow} 0 $.
        \item[$(iii)$] 		$\frac{1}{T}\sum_{t = 1}^{T}  \tr\left(\Delta\bPhi\tvec(\bX_{t-1}) \tvec(\bX_{t-1}) ^\T(\Delta\bPhi)^\T\right) 	
		-  \bbE \tr\left(\Delta\bPhi\tvec(\bX_{t-1}) \tvec(\bX_{t-1}) ^\T(\Delta\bPhi)^\T\right)  \overset{p}{\rightarrow} 0$.
        \item[$(iv)$]$	\frac{1}{T}\sum_{t = 1}^{T}\tr\left(\Delta\bG\tvec(\bX_t)\be_t)^\T\right) - \bbE  \tr\left(\Delta\bG\tvec(\bX_t)\be_t^\T\right) \overset{p}{\rightarrow} 0$.
        \item[$(v)$]$	\frac{1}{T}\sum_{t = 1}^{T} \tr(\Delta\bPhi \tvec(\bX_{t-1}) \tvec(\bE_t)^\T) - \bbE \tr\left(\Delta\bPhi\tvec(\bX_{t-1}) \tvec(\bE_t)^\T\right) \overset{p}{\rightarrow} 0$.
        \item[$(vi)$] $\frac{1}{T}\sum_{t = 1}^{T} \tr\left(\Delta\bG\tvec(\bX_t)\tvec(\bX_{t-1})^\T  (\Delta\bPhi)^\T\right) - \bbE   \tr\left(\Delta\bG \tvec(\bX_t)\tvec(\bX_{t-1})^\T  (\Delta\bPhi)^\T\right)\overset{p}{\rightarrow} 0$.
	\end{itemize}
    where $\Delta\bG \coloneq \bG - \bG_0$ and $\Delta\bPhi \coloneq \bPhi - \bPhi_0$. Therefore, with $\sigma^2 < \infty$, 
	\bel{pf-part1-6}
	T^{-1} \left(\calL(\btheta) - \bbE\calL(\btheta)\right)  \overset{p}{\rightarrow} 0.
	\eel
Secondly, we aim to  show $T^{-1}\bbE\calL(\btheta)$ is uniformly equicontinuous in $\btheta$. In other words, we need to prove that $\ln|\bG|$ and $\bbE \bv_t^\T\bv_t$ are both uniformly equicontinuous. For $\ln |\bG|$, we have 
	\bel{pf-part2-G}
	\partial \ln |\bG| && = \partial \ln |\bI_{kn} - \bW \otimes \bC| \cr
	&& =  - \tr\left((\bI_{kn} - \bW \otimes \bC)^{-1} \bW \otimes \partial \bC\right).
	\eel
Since both $\bW$ and $(\bI_{kn} - \bW \otimes \bC)^{-1}$ are bounded matrices, it follows that $ \ln |\bG| $ is uniformly equicontinuous. Then, we consider reorganizing the formulation of $\bv_t$ and $\bbE \bv_t^\T\bv_t$ as follows
	\bel{pf-part2-1}
	\bv_t &&= \bG \tvec(\bX_t) - \bPhi \tvec(\bX_{t-1}) \cr 
	&& = (\bG\bG_{0}^{-1} - \bI_{kn}) \bPhi_{0} \tvec(\bX_{t-1}) + (\bPhi_{0} - \bPhi)  \tvec(\bX_{t-1}) + \bG\bG_{0}^{-1}\be_t,
	\eel
    and 
	\bel{pf-part2-2}
 && \bbE \bv_t^\T\bv_t  =    2 \bbE \tr\left([ (\bG\bG_{0}^{-1} - \bI_{kn}) \bPhi_{0} \tvec(\bX_{t-1}) + (\bPhi_{0} - \bPhi)  \tvec(\bX_{t-1})]\times [\bG\bG_{0}^{-1}\be_t]^\T\right) \cr 
	&&\qquad \qquad + \sigma_{0}^2 \tr([\bG_{0}^{-1}]^\T\bG^\T \bG\bG_{0}^{-1}) \cr
    &&\qquad \qquad + \bbE  \|(\bG\bG_{0}^{-1} - \bI_{kn}) \bPhi_{0} \tvec(\bX_{t-1}) + (\bPhi_{0} - \bPhi)  \tvec(\bX_{t-1})\|_2^2 .
	\eel
	The first term  of (\ref{pf-part2-2}) tends to 0 when $T$ goes to infinity due to the natural assumption that $\tvec(\bE_t)$ is independent of $\tvec(\bX_{t-1})$. The second term of (\ref{pf-part2-2}) is a polynomial function of $\bC$ as the fact that 
	\bel{pf-part2-3}
	\bG \bG_{0}^{-1} &&= (\bI_{kn} - \bW \otimes \bC) (\bI_{kn} - \bW \otimes \bC_{0})^{-1} \cr
	&& = \bG_{0}^{-1} - (\bW \otimes \bC)\bG_{0}^{-1}\cr
	&& = (\bG_{0} + \bW \otimes \bC_{0})\bG_{0}^{-1} - (\bW \otimes \bC)\bG_{0}^{-1} \cr
	&& = \bI_{kn} + (\bW \otimes (\bC_{0} - \bC))\bG_{0}^{-1}.
	\eel
Thus, 	the last term  of (\ref{pf-part2-2}) can be rewritten as 
	\bel{pf-part2-4}
 \bbE  \|(\tvec(\bX_{t-1})^\T \otimes \bI_{kn} ) \tvec(\bPhi_{0} - \bPhi) + (\bW \bZ_{t-1}^\T \otimes \bI_{k})\tvec(\bC_{0} - \bC)\|_2^2,
	\eel
	where $\tvec(\bZ_{t-1}) : = \bG_{0}^{-1} \bPhi_{0} \tvec(\bX_{t-1})$. Hence, it equals to 
	\bel{pf-part2-5}
	\left((\tvec(\bPhi_{0} - \bPhi)^\T,\tvec(\bC_{0} - \bC)^\T\right) \bbE\calH_{T}\left((\tvec(\bPhi_{0} - \bPhi)^\T,\tvec(\bC_{0} - \bC)^\T\right),
	\eel
        where 
    \bel{pf-part2-6}
    \calH_{T} \coloneq \frac{1}{T} \sum_{t = 1}^{T} \left((\tvec(\bX_{t-1})^\T \otimes \bI_{kn} ), (\bW \bZ_{t-1}^\T \otimes \bI_{k})\right)^\T \left((\tvec(\bX_{t-1})^\T \otimes \bI_{kn} ), (\bW \bZ_{t-1}^\T \otimes \bI_{k})\right).
    \eel
	which obviously  uniformly equicontinuous. In summary, combining the result of (\ref{pf-part2-G}), $T^{-1}\bbE\calL(\btheta)$ is uniformly equicontinuous for $\btheta \in \bTheta$.

Thirdly, we need to establish the global identification of $(\bC_0, \bPhi_0, \sigma_{0}^2)$. Before that, let's rewritten some notations. We have
    $\bbE \sum_{t = 1}^{T}\be_{t}^\T \be_{t} = kn T\sigma^2_0$
    and 
    \bel{pf-part3-1}
	\bbE \ln \calL_{T}(\btheta_{0}) =  -\frac{knT}{2}\ln 2\pi - \frac{knT}{2}\ln \sigma^2_0 + T \ln |\bG_{0}| - \frac{knT}{2}.
	\eel
    Besides, according to (\ref{pf-part2-3}), we get
\bel{pf-part3-norm}
	&& \|\bG \tvec(\bX_t) - \bPhi \tvec(\bX_{t-1})) \|_2^2 \cr
	&& =  \|(\bPhi_{0} - \bPhi) \tvec(\bX_{t-1})) + \bG\tvec(\bX_t)  - \bPhi_{0}  \tvec(\bX_{t-1}))  \|_2^2 \cr
	&& =  \|(\bPhi_{0} - \bPhi) \tvec(\bX_{t-1})) + (\bG\bG_{0}^{-1} - \bI_{kn})  \bPhi_{0}  \tvec(\bX_{t-1})) + \bG(\tvec(\bX_t) - \bG_{0}^{-1} \bPhi_{0} \tvec(\bX_{t-1})) ) \|_2^2 \cr
	&&= \|(\tvec(\bX_{t-1})^\T \otimes \bI_{kn} ) \tvec(-\Delta\bPhi) + (W \bZ_{t-1}^\T \otimes \bI_{k})\tvec(-\Delta\bC)  +
	\bG_0(\tvec(\bX_t) - \bG_{0}^{-1} \bPhi_{0} \tvec(\bX_{t-1}))  \|_2^2, \nonumber \\
	\eel
    where $\Delta \bPhi \coloneq  \bPhi - \bPhi_{0}$,  $\Delta \bC \coloneq  \bC - \bC_{0}$ and $\tvec(\bZ_{t-1}) \coloneq \bG_{0}^{-1} \bPhi_{0} \tvec(\bX_{t-1})$.
Then we arrive at
	\bel{pf-part3-2}
	&& \frac{1}{T}\bbE \ln \calL_{T}(\btheta) - \frac{1}{T}\bbE \ln \calL_{T}(\btheta_{0}) \cr
	&& = - \frac{kn}{2}(\ln \sigma^2 - \ln \sigma^2_0) + \ln |\bG|  - \ln |\bG_{0}|  - \left(\frac{1}{2\sigma^2 }\frac{1}{T} \sum_{t = 1}^{T}\bbE \bv_{t}^\T \bv_{t} - kn \right)\cr
	&& \coloneq U_{1,T}(\bC,\sigma^2)  -  \frac{1}{2\sigma^2 } U_{2,T}(\bC, \bPhi) + o(1)
	\eel
	where 
	\bel{pf-part3-3}
	U_{1,T}(\bC,\sigma^2) \!=\! - \frac{kn}{2}(\ln \sigma^2 - \ln \sigma^2_0) + \ln |\bG|  - \ln |\bG_{0}| - \frac{1}{2\sigma^2} \left(\sigma^2_{0}\tr([\bG_{0}^{-1}]^\T \bG^\T \bG \bG_{0}^{-1}) - kn \sigma^2 \right)\!,
	\eel
	and 
	\bel{pf-part3-4}
	U_{2,T}(\bC, \bPhi) &&=  \frac{1}{2\sigma^2 }\frac{1}{T} \sum_{t = 1}^{T}\bbE \{ \left[(\tvec(\bX_{t-1})^\T \otimes \bI_{kn} ) \tvec(\bPhi_{0} - \bPhi) + (\bW \bZ_{t-1}^\T \otimes \bI_{k})\tvec(\bC_{0} - \bC)\right]^\T \cr
 && \qquad \left[(\tvec(\bX_{t-1})^\T \otimes \bI_{kn} ) \tvec(\bPhi_{0} - \bPhi) + (\bW \bZ_{t-1}^\T \otimes \bI_{k})\tvec(\bC_{0} - \bC)\right]  \}.
	\eel
	We can verify the last equation in  (\ref{pf-part3-2}) holds by referring to  equation (\ref{pf-part2-3}) and (\ref{pf-part3-norm}). Consequently, our goal now is to demonstrate the global identification of  $(\bC_0, \sigma^2_0)$ in  $U_{1,T}(\bC,\sigma^2)$ and $(\bC_0, \bPhi_0)$ in $U_{2,T}(\bC, \bPhi)$, respectively. 

	In the following, we first prove $U_{1,T}(C,\sigma^2) \leq 0$. Consider the process
\bel{pf-part3-5}
	\bG \tvec(\bX_t)=  \tvec(\bE_t),
	\eel
	for a period $t$. The log likelihood of this process is 
\bel{pf-part3-6}
	\ln \ell_{t}(\bC, \sigma^2) = -\frac{kn}{2}\ln 2\pi - \frac{kn}{2}\ln \sigma^2 +  \ln |\bG| - \frac{1}{2\sigma^2} (\bG\tvec(\bX_t))^\T \bG\tvec(\bX_t).
	\eel
	Let $\bbE(\cdot)$ be the expectation operator for $\tvec(\bX_t)$ based on this process. It follows that 
\bel{pf-part3-7}
	&& \bbE \ln \ell_{t}(\bC, \sigma^2) - \bbE \ln \ell_{t}(\bC_{0}, \sigma^2_{0}) \cr
	&&= - \frac{kn}{2}(\ln \sigma^2 - \ln \sigma^2_0) + \ln |\bG|  - \ln |\bG_{0}| - \frac{1}{2\sigma^2} \left(\sigma^2_{0}\tr([\bG_{0}^{-1}]^\T \bG^\T \bG \bG_{0}^{-1})  - kn \sigma^2 \right).
	\eel
By $\ln x \leq 2(\sqrt{x} -1)$ for $x>0$, we have
\bel{pf-part3-8}
	\bbE\ln \ell_{t}(\bC, \sigma^2) -  \bbE \ln \ell_{t}(\bC_{0}, \sigma^2_{0}) &&= \bbE \ln \ell_{t}(\bC, \sigma^2)/\ell_{t}(\bC_{0}, \sigma^2_{0}) \cr
	&& \leq 2 (\bbE \sqrt{\ell_{t}(\bC, \sigma^2)/\ell_{t}(\bC_{0}, \sigma^2_{0}) }-1) \cr
	&& = 2 \int \left(\sqrt{\ell_{t}(\bC, \sigma^2)/\ell_{t}(\bC_{0}, \sigma^2_{0}) }-1\right) \ell_{t}(\bC_{0}, \sigma^2_{0})   d\tvec(\bX_t) \cr
	&& = 2 \left(\int \sqrt{\ell_{t}(\bC, \sigma^2)\ell_{t}(\bC_{0}, \sigma^2_{0}) } d\tvec(\bX_t) - 1\right) \cr  
	&& = - \int \left( \sqrt{\ell_{t}(\bC, \sigma^2) }  - \sqrt{\ell_{t}(\bC_{0}, \sigma^2_{0}) } \right)^2 d\tvec(\bX_t)\cr
	&& \leq 0.
\eel
	Thus, $U_{1,T}(\bC,\sigma^2) \leq 0$ for any $(\bC,\sigma^2)$ and the equality holds if and only if $\ln \ell_{t}(\bC, \sigma^2) = \ln \ell_{t}(\bC_{0}, \sigma^2_{0})$.
    
	Next, we show that $\ln \ell_{t}(\bC, \sigma^2) = \ln \ell_{t}(\bC_{0}, \sigma^2_{0})$ if and only if  $(\bC, \sigma^2) = (\bC_{0}, \sigma^2_{0})$, i.e., $(\bC_{0}, \sigma^2_{0})$ is the unique maximizer. If $(\bC_{0}, \sigma^2_{0})$ is not identified, then there exists $(\tbC_{0}, \tsigma^2_{0}) \neq (\bC_{0}, \sigma^2_{0})$ such that $\ln \ell_{t}(\bC_{0}, \sigma^2_{0}) = \ln \ell_{t}(\tbC_{0}, \tsigma^2_{0})$. Take the second derivative with respect to $\tvec(\bX_{t})$ in (\ref{pf-part3-6}), we have
	\bel{pf-part3-8-2}
	\frac{1}{2\sigma_{0}^2} \bG_{0}^\T \bG_{0}  = \frac{1}{2\tsigma_{0}^2} \tbG_{0}^\T \tbG_{0}. 
	\eel
	Specifically, that is 
	\bel{eq:thm-prf-1}
	&& \frac{1}{2\sigma_{0}^2} (\bI_{kn} - \bW \otimes \bC_{0})^\T (\bI_{kn} - \bW \otimes \bC_{0}) \cr
	&& = \frac{1}{2\sigma_{0}^2} \left(\bI_{kn} -\bW \otimes \bC_{0} - \bW^\T \otimes \bC_{0}^\T + \bW^\T \bW\otimes  \bC_{0}^\T \bC_{0}\right) \cr
	&& =  \frac{1}{2\tsigma_{0}^2} \left(\bI_{kn} -\bW \otimes \tbC_{0} - \bW^\T \otimes \tbC_{0}^\T + \bW^\T \bW\otimes  \tbC_{0}^\T \tbC_{0}\right).
	\eel
	Let us consider the block diagonal matrix of size $k \times k$ in the equation (\ref{eq:thm-prf-1}).  That is 
	\bel{pf-part3-9}
	&& \frac{1}{2\sigma_{0}^2} \left(\bI_{k} -\bW_{jj}  \bC_{0} - \bW_{jj} \bC_{0}^\T + \sum_{i = 1}^{n}\bW_{ij}^2  \bC_{0}^\T \bC_{0}\right)  \cr 
	&& =  \frac{1}{2\sigma_{0}^2} \left(\bI_{k} + \sum_{i = 1}^{n}\bW_{ij}^2  \bC_{0}^\T \bC_{0}\right) \cr
	&& = \frac{1}{2\tsigma_{0}^2} \left(\bI_{k} + \sum_{i = 1}^{n}\bW_{ij}^2  \tbC_{0}^\T \tbC_{0}\right),
	\eel
	where the first equation holds because of the diagonal entry of $\bW$ is zero, i.e., $\bW_{jj} = 0$. Moreover, there exists $j' \neq j$ such that $\sum_{i = 1}^{n}\bW_{ij'}^2 \neq \sum_{i = 1}^{n}\bW_{ij}^2 $, we obtain 
	\bel{eq:thm-prf-2}
	1/\sigma_{0}^2 = 1/\tsigma_{0}^2 \text{  and   } \bC_{0}^\T \bC_{0}/\sigma_{0}^2   = \tbC_{0}^\T \tbC_{0}/\tsigma_{0}^2.
	\eel
	Hence, $\sigma_{0}^2  = \tsigma_{0}^2 $ and $\bC_{0} ^\T\bC_{0} = \tbC_{0}^\T \tbC_{0}$. We then consider the off block diagonal matrix of (\ref{eq:thm-prf-1}), i.e, $k \neq j$, we have 
	\bel{pf-part3-10}
	&& \frac{1}{2\sigma_{0}^2} \left(-\bW_{kj}  \bC_{0} - \bW_{jk} \bC_{0}^\T + \sum_{i = 1}^{n}\bW_{ik}\bW_{ij}  \bC_{0}^\T \bC_{0}\right)  \cr 
	&& = \frac{1}{2\tsigma_{0}^2} \left(-\bW_{kj}  \tbC_{0} - \bW_{jk} \tbC_{0}^\T + \sum_{i = 1}^{n}\bW_{ik}\bW_{ij}  \tbC_{0}^\T \tbC_{0}\right).
	\eel
	Combining with (\ref{eq:thm-prf-2}), we get 
	\bel{pf-part3-11}
	\bW_{kj}  \bC_{0} + \bW_{jk} \bC_{0}^\T  = \bW_{kj}  \tbC_{0} + \bW_{jk} \tbC_{0}^\T.
	\eel
	If $\bC_{0}$ is symmetric and $\bW + \bW^\T\neq \mathbf{0}_{n\times n}$, or if $\bC_{0}$ is asymmetric, we have $\bC_{0} = \tbC_{0}$. Thus, $(\bC_{0},\sigma_{0}^2)$ is the unique maximizer of $U_{1,T}(\bC,\sigma^2) $. 
	
	On the other hand, $U_{2,T}(\bC, \bPhi)$ is a quadratic function in terms of $\bPhi$ and $\bC$. If $\lim_{T\rightarrow \infty}\bbE \calH_{T} $ is nonsingular,  where $\calH_{T}$ define in (\ref{pf-part2-6}),
	then  $U_{2,T}(\bC, \bPhi) > 0 $ whenever  $(\bC, \bPhi) \neq (\bC_{0}, \bPhi_{0})$. Otherwise, if $\lim_{T\rightarrow \infty}\bbE \calH_{T} $ is singular, $\bC$ and $\bPhi$ cannot be identified solely  from $U_{2,T}(\bC, \bPhi)$. However, since $(\bC_{0},\sigma_{0}^2)$ is the unique maximizer of $U_{1,T}(\bC,\sigma^2) $, given $\bC_0$, $\bPhi_0$ can still be identified from  $U_{2,T}(\bC, \bPhi)$.  Therefore,  $(\bC, \phi, \sigma^2)$ is uniquely and globally identified. 
    
    In addition, combining $	T^{-1} \left(\calL(\btheta) - \bbE\calL(\btheta)\right)  \overset{p}{\rightarrow} 0$ with the  uniform equicontinuity of $T^{-1}\bbE\calL(\btheta)$   for $\btheta \in \bTheta$, 
	we invoke the results of Theorem 4.1.1 of \citep{amemiya1985advanced}, to obtain
    \bel{pf-part3-12}
    \left(\tvec(\hbC^{mle}), \tvec(\hbPhi^{mle}), (\hsigma^{2})^{mle}\right) \overset{p}{\rightarrow} \left(\tvec(\bC_{0}), \tvec(\bPhi_{0}), \sigma^2_0\right) .
    \eel
\end{proof}
\subsection{Proof of Proposition \ref{prop:2}}\label{supp:pf-prop-2}
\begin{proof}
	Recall that 
	\bel{pf-prop1-1}
	\frac{1}{\sqrt{T}} \frac{\partial \ln \calL_{T}(\btheta_{0})}{\partial \btheta} = \begin{pmatrix}
		\frac{1}{\sigma_{0}^2}\frac{1}{\sqrt{T}}\sum_{t = 1}^{T} \left((\bZ_{t-1}W^\T \otimes \bI_{k}) \be_{t} +   (\tbE_{t}W^\T \otimes \bI_{k}) \be_{t} - \sigma_{0}^2 \calR_1(\bG_{0}^{-\T})^\T  \tvec(W)\right)\\
		\frac{1}{\sigma^2_0}\frac{1}{\sqrt{T}}\sum_{t = 1}^{T} (\tvec(\bX_{t-1}) \otimes \bI_{kn}) \be_{t} \\	\frac{1}{2\sigma^4_0}\frac{1}{\sqrt{T}}\sum_{t = 1}^{T}  \left(\be_{t}^\T\be_{t} - kn \sigma^2_0\right)
	\end{pmatrix},\nonumber \\
	\eel
    where  $\tvec(\bZ_{t-1}) : = \bG_{0}^{-1} \bPhi_{0} \tvec(\bX_{t-1})$ and $\tbE_{t} \coloneq \tvec^{-1}(\bG_0 \bE_t)$. Given that $\bW$ is a constant matrix as stated in Assumption \ref{assum:3} and both $\bG_0$ and $\bPhi_0$ are bounded,  it is straightforward to verify that (\ref{pf-prop1-1}) satisfies the conditions of Lemma 13 in \citep{yu2008quasi}. Consequently, 
	by applying the CLT of the martingale difference array of  Lemma 13 in \citep{yu2008quasi}, we have 
	\bel{pf-prop1-2}
	\frac{1}{\sqrt{T}} \frac{\partial \ln \calL_{T}(\btheta_{0})}{\partial \btheta} 
	\overset{d}{\Rightarrow} \calN(0,\Psi_{\btheta_{0}} ),
	\eel
    where $ \Psi_{\btheta_{0}} \coloneq \lim_{T \rightarrow \infty}  \Psi_{\btheta_{0},T}$ and $\Psi_{\btheta_{0},T}$ is provided in (\ref{psi}).
\end{proof}
\subsection{Proof of Theorem \ref{thm:2}}\label{proof:thm2}
\begin{proof}
	According to the Taylor expansion, 
	\bel{pf-thm2-1}
	\sqrt{T} 	(\hbtheta^{mle} -  \btheta_{0} ) = \left(-\frac{1}{{T}} \frac{\partial^2 \ln \calL_{T}(\bar{\btheta})}{\partial \btheta \partial \btheta^\T}\right)^{-1} \cdot  \left(\frac{1}{\sqrt{T}} \frac{\partial \ln \calL_{T}(\btheta_{0})}{\partial \btheta}\right),
	\eel
	where  $\bar{\btheta}$ lies between $\btheta_{0}$ and $\hbtheta^{mle}$. As 
\bel{pf-thm2-2}
	-\frac{1}{{T}} \frac{\partial^2 \ln \calL_{T}(\bar{\btheta})}{\partial \btheta \partial \btheta^\T} \!=  \! \!\underset{P_1}{\underbrace{-\frac{1}{{T}} \frac{\partial^2 \ln \calL_{T}(\bar{\btheta})}{\partial \btheta \partial \btheta^\T} \!- \! \left( \!-\frac{1}{{T}} \frac{\partial^2 \ln \calL_{T}({\btheta}_{0})}{\partial \btheta \partial \btheta^\T} \right)}}   \! +  \! \underset{P_2}{\underbrace{\left(	-\frac{1}{{T}} \frac{\partial^2 \ln \calL_{T}({\btheta}_{0})}{\partial \btheta \partial \btheta^\T}  \! -  \!\Xi_{\btheta_{0},T} \right)}}  \!+\!\Xi_{\btheta_{0},T}.
	\eel
	We will show that the first term $P_1$ in (\ref{pf-thm2-2}) has the same order as $\|\hbtheta^{mle} - \btheta_{0}\|_2$, as established in Lemma \ref{lemma:6}, 
\bel{pf-thm2-3}
	\left(	-\frac{1}{{T}} \frac{\partial^2 \ln \calL_{T}(\bar{\btheta})}{\partial \btheta \partial \btheta^\T} - \left(	-\frac{1}{{T}} \frac{\partial^2 \ln \calL_{T}({\btheta}_{0})}{\partial \btheta \partial \btheta^\T} \right)\right) = \|\hbtheta^{mle} - \btheta_{0}\|_2 \cdot O_{p}(1).
	\eel
Meanwhile, by Lemma \ref{lemma:7},  the second term $P_2$ in (\ref{pf-thm2-2}) converge to zero as follows:
\bel{pf-thm2-4}
	\left(	-\frac{1}{{T}} \frac{\partial^2 \ln \calL_{T}({\btheta}_{0})}{\partial {\btheta} \partial {\btheta}^\T} - \Xi_{\btheta_{0},T} \right) = O_{p}(\frac{1}{\sqrt{T}}).
	\eel
	Thus,  (\ref{pf-thm2-2}) can be rewritten as
\bel{pf-thm2-5}
	-\frac{1}{{T}} \frac{\partial^2 \ln \calL_{T}(\bar{\btheta})}{\partial {\btheta} \partial {\btheta}^\T} = \|\hbtheta^{mle} - \btheta_{0}\|_2 \cdot O_{p}(1) + O_{p}(\frac{1}{\sqrt{T}}) + \Xi_{\btheta_{0},T}.
	\eel
Then, as shown in Theorem \ref{thm:1}, we have $\|\hbtheta^{mle} - \btheta_{0}\|_2  = o_{p}(1)$ . Moreover, with $\Xi_{\btheta_{0},T}$ is nonsingular in the limit,  we can obtain $ 	-\frac{1}{{T}} \frac{\partial^2 \ln \calL_{T}(\bar{\btheta})}{\partial {\btheta} \partial {\btheta}^\T}$ is invertible  for large $T$ and 
	\bel{pf-thm2-6}
	\left(-\frac{1}{{T}} \frac{\partial^2 \ln \calL_{T}(\bar{\btheta})}{\partial {\btheta} \partial {\btheta}^\T}\right)^{-1}	 = O_p(1).
	\eel
	Thus, combining (\ref{pf-thm2-6}) and the result of Proposition \ref{prop:2}, we get
	\bel{pf-thm2-7}
	\sqrt{T} 	(\hbtheta^{mle} -  \btheta_{0} ) = O_p(1),
	\eel
	which implies that  
\bel{pf-thm2-8}
	(\hbtheta^{mle} -  \btheta_{0} ) = O_p(\frac{1}{\sqrt{T} }).
	\eel
	Furthermore, according to (\ref{pf-thm2-1}), we have
	\bel{pf-thm2-9}
	\sqrt{T} 	(\hbtheta^{mle} -  \btheta_{0} ) = \left( \Xi_{\btheta_{0},T} + O_p(\frac{1}{\sqrt{T}})\right)^{-1} \cdot  \left(\frac{1}{\sqrt{T}} \frac{\partial \ln \calL_{T}(\btheta_{0})}{\partial \btheta}\right).
	\eel
	Using the fact that 
	\bel{pf-thm2-10}
	\left( \Xi_{\btheta_{0},T} + O_p(\frac{1}{\sqrt{T}})\right)^{-1}  = \left( \Xi_{\btheta_{0},T} \right)^{-1}  + O_p(\frac{1}{\sqrt{T}}).
	\eel
	Therefore, by (\ref{pf-thm2-9}) and (\ref{pf-thm2-10}), we arrive at
		\bel{pf-thm2-11}
	\sqrt{T} 	(\hbtheta^{mle} -  \btheta_{0} ) = \Xi_{\btheta_{0},T} ^{-1} \cdot  \left(\frac{1}{\sqrt{T}} \frac{\partial \ln \calL_{T}(\btheta_{0})}{\partial \btheta}\right) + O_p(\frac{1}{\sqrt{T}}) \cdot  \left(\frac{1}{\sqrt{T}} \frac{\partial \ln \calL_{T}(\btheta_{0})}{\partial \btheta}\right).
	\eel
	As $\Xi_{\btheta_{0}} = \lim\limits_{T \rightarrow \infty}\Xi_{\btheta_{0},T}$ exists, and  using the result of the Proposition \ref{prop:2},  we achieve
	\bel{pf-thm2}
	\sqrt{T}(\hbtheta^{mle} -  \btheta_{0} )\overset{d}{\Rightarrow} \calN(\mathbf{0}, \Xi_{\btheta_{0}}^{-1} \Psi_{\btheta_0}\Xi_{\btheta_{0}}^{-1} ).
	\eel
\end{proof}

\section{Some theoretical properties related to  QMLE}\label{supp:proof-related}
\subsection[The first and second derivatives of determinate]{The first and second derivatives of $\ln |\mathbf{I}_{kn} - \bW\otimes \mathbf{C}|$}

For simplicity, we denote $\bG = \bI_{kn} - \bW \otimes \bC$ and $\bG_0 = \bI_{kn} - \bW \otimes \bC_0$. The first derivative with respect to $\tvec(\bC) $ is 
\bes
\partial \ln |\bG| && = \partial \ln |\bI_{kn} - \bW \otimes \bC| \cr
&& =  - \tr\left((\bI_{kn} - \bW \otimes \bC)^{-1} \bW \otimes \partial \bC\right) \cr
&& = - \langle \bG^{-\T},  \bW \otimes \partial \bC \rangle \cr
&& = - \langle \calR_1(\bG^{-\T}),  \tvec(\bW) \tvec(\partial \bC)^\T \rangle \cr
&& = -   \tvec(\bW)^\T \calR_1(\bG^{-\T}) \tvec(\partial \bC).  
\ees
Hence, 
\bel{deriv: c0-1}
\frac{\partial \ln |\bG| }{\partial \tvec(\bC)  } &&= - \calR_1(\bG^{-\T})^\T  \tvec(\bW).
\eel
Moreover, as the fact that
\bes
\partial \bG^{-\T} = (\bG^{-1} \cdot (\bW \otimes \partial \bC) \cdot  \bG^{-1})^\T,
\ees
we consider treating $(\bG^{-1})$ as a $n \times n$ array of blocks of the same block size $k \times k$ and let $(\bG^{-1})_{[i,j]}$ be the $(i,j)$-th block, $1 \leq i \leq n, 1 \leq j \leq n$. This allows us to compute the $(i,j)$-th block of $(\bG^{-1} \cdot (W \otimes \partial \bC) \cdot  \bG^{-1})^\T$ is 
\bes
\sum_{\tilde{i}}^{n} \sum_{\tilde{j}}^{n} \bW_{\tilde{i},\tilde{j}} [(\bG^{-1})_{[\tilde{i},i]}]^\T \cdot \partial \bC^\T \cdot [(\bG^{-1})_{[j,\tilde{j}]}]^\T.
\ees
Thus, 
\bes
\partial \calR_1(\bG^{-\T})^\T \tvec(W) &&= \sum_{i = 1}^{n} \sum_{j = 1}^{n} \sum_{\tilde{i} = 1}^{n} \sum_{\tilde{j} = 1}^{n} W_{\tilde{i},\tilde{j}} W_{i,j}     \left((\bG^{-1})_{[j,\tilde{j}]} \otimes [(\bG^{-1})_{[\tilde{i},i]}]^\T \right) \partial \tvec( \bC^\T) \cr
&& = \sum_{i = 1}^{n} \sum_{j = 1}^{n} \sum_{\tilde{i} = 1}^{n} \sum_{\tilde{j} = 1}^{n} W_{\tilde{i},\tilde{j}} W_{i,j}     \calP \left((\bG^{-1})_{[j,\tilde{j}]} \otimes [(\bG^{-1})_{[\tilde{i},i]}]^\T \right) \partial \tvec( \bC).
\ees
where $\calP(\cdot)$ is a { column-wise permutation operator} that ensures the validity of the last equality. Therefore, we have
\bel{deriv: c0-2}
\frac{\partial^2 \ln |\bG| }{\partial \tvec(\bC) \tvec( \bC)^\T } &&= 
\frac{\partial \calR_1(\bG^{-\T})^\T \tvec(W)}{\partial \tvec( \bC)^\T } \cr
&&= \sum_{i,j,\tilde{i},\tilde{j}} W_{\tilde{i},\tilde{j}} W_{i,j}     \calP \left((\bG^{-1})_{[j,\tilde{j}]} \otimes [(\bG^{-1})_{[\tilde{i},i]}]^\T \right)^\T \cr
&& \overset{\Delta}{=}  \tbW_{\tvec(\bC)} .
\eel
\subsection[The first and second derivatives of log-likelihood]{The first and second derivatives of $\ln \mathcal{L}_{T}({\btheta})$}\label{sec:8-3}
Recall the log-likelihood is 
\bes
\ln \calL_{T}(\btheta) = -\frac{knT}{2}\ln 2\pi - \frac{knT}{2}\ln \sigma^2 + T \ln |\bG| - \frac{1}{2\sigma^2}\sum_{t = 1}^{T} \bv_t^\T\bv_t.
\ees
We can rewrite $\bv_t$ as followings
\bel{vt}
\bv_t &&= \bG \tvec(\bX_t) - \bPhi \tvec(\bX_{t-1}) \cr
&& = \tvec(\bX_t) - (\bW \otimes \bC)\tvec(\bX_t) - \bPhi\tvec(\bX_{t-1})\cr
&& = \tvec(\bX_t) - (\bW \bX_t ^\T\otimes \bI_{k})\tvec(\bC) - (\tvec(\bX_{t-1})^\T \otimes \bI_{kn})\tvec(\bPhi).
\eel
The first derivative of $\ln \calL_{T}(\btheta)$ is
\bel{deriv:1}
\frac{1}{\sqrt{T}} \frac{\partial \calL_{T}(\btheta)}{\partial \btheta} = \begin{pmatrix}
	\frac{1}{\sigma^2}\frac{1}{\sqrt{T}}\sum_{t = 1}^{T} \left((\bX_{t}\bW^\T \otimes \bI_{k}) \bv_t -\sigma^2  \calR_1((\bG^{-1})^\T)^\T  \tvec(\bW) \right)\\
	\frac{1}{\sigma^2}\frac{1}{\sqrt{T}}\sum_{t = 1}^{T} (\tvec(\bX_{t-1})^\T \otimes \bI_{kn})^\T \bv_{t} \\	\frac{1}{2\sigma^4}\frac{1}{\sqrt{T}}\sum_{t = 1}^{T}  \left(\bv_t^\T\bv_t - kn \sigma^2\right)
\end{pmatrix}.
\eel
Moreover, the second order derivative is 
\bel{deriv:2}
&& \frac{1}{{T}} \frac{\partial^2 \calL_{T}(\btheta)}{\partial \btheta \partial \btheta^\T} = \frac{-1}{T} \times \cr
&&   \begin{pmatrix}
	\calQ_{1,1},&	\frac{1}{\sigma^2}\sum_{t = 1}^{T} (\tvec(\bX_{t-1}) \otimes \bI_{kn})(\bW \bX_{t}^\T \otimes \bI_{k})  , & *\\
	*	,&	\frac{1}{\sigma^2}\sum_{t = 1}^{T} \tvec(\bX_{t-1})\tvec(\bX_{t-1})^\T  \otimes \bI_{kn},& * \\	
	\frac{1}{\sigma^4}\sum_{t = 1}^{T} \left((\bX_{t}\bW^\T \otimes \bI_{k}) \bv_t \right) ,&  \frac{1}{\sigma^4}\sum_{t = 1}^{T} (\tvec(\bX_{t-1}) \otimes \bI_{kn}) \bv_{t}  ,&  \calQ_{3,3}
\end{pmatrix},\cr
&& \quad
\eel
where 
\bes
 \calQ_{1,1} = 	\frac{1}{\sigma^2}\sum_{t = 1}^{T} \left(\bX_{t}\bW^\T \bW \bX_t ^\T \otimes \bI_{k}  + \sigma^2 \tbW_{\tvec(\bC)}\right)   ,
\ees
and 
\bes
\calQ_{3,3} = - \frac{knT}{2\sigma^4} + \frac{1}{\sigma^6}\sum_{t = 1}^{T} \bv_t^\T\bv_t.
\ees
Among that, $\tbW_{\tvec(\bC)}$ is defined in (\ref{deriv: c0-2}).
\subsection[The first and second derivatives of log-likelihood function at true value]{The first and second derivatives of $\ln \mathcal{L}$ at $\mathbold{\theta}_0$}\label{proof:deriv2}
Note that when $\btheta = \btheta_{0}$,  we have $\be_{t} \coloneq \bG_{0} \tvec(\bX_t) - \bPhi_{0} \tvec(\bX_{t-1}) = \tvec(\bE_{t})$. Then the first derivative of $\ln \calL(\btheta_{0})$ becomes 
\bel{deriv:3}
\frac{1}{\sqrt{T}} \frac{\partial \calL_{T}(\btheta_{0})}{\partial \btheta} = \begin{pmatrix}
	\frac{1}{\sigma_{0}^2}\frac{1}{\sqrt{T}}\sum_{t = 1}^{T} \left((\bZ_{t-1}\bW^\T \otimes \bI_{k}) \be_{t}+   (\tbE_{t}\bW^\T \otimes \bI_{k}) \be_{t} - \sigma_{0}^2 \calR_1(\bG_{0}^{-\T})^\T  \tvec(\bW)\right)\\
	\frac{1}{\sigma^2}\frac{1}{\sqrt{T}}\sum_{t = 1}^{T} (\tvec(\bX_{t-1}) \otimes \bI_{kn}) \be_{t} \\	\frac{1}{2\sigma^4}\frac{1}{\sqrt{T}}\sum_{t = 1}^{T}  \left(\be_{t}^\T\be_{t} - kn \sigma^2\right)
\end{pmatrix}. \nonumber \\
\eel
where $\tvec(\widetilde{\bE_t}) \coloneq(\bI -  \bW \otimes \bC_0)^{-1} \tvec(\bE_t)$. Subsequently, we will derive the information matrix  $\Xi_{\btheta_{0},T} \coloneq - \bbE \left(\frac{1}{{T}} \frac{\partial^2 \calL_{T}(\btheta_{0})}{\partial  \btheta \partial \btheta^\T} \right) $. For notational purpose, we define $\tvec(\bZ_{t-1}) : = \bG_{0}^{-1} \bPhi_{0} \tvec(\bX_{t-1})$, then our true model can be rewritten as  
\bel{A4-2}
\tvec(\bX_{t}) &&= \bG_{0}^{-1}\bPhi_{0} \tvec(\bX_{t-1}) + \bG_{0}^{-1} \tvec(\bE_{t})\cr
&& = \tvec(\bZ_{t-1})  + \tvec(\tbE_{t}).
\eel
We aim to  simplify the second-order derivative of $\ln\calL(\btheta)$ in  (\ref{deriv:2}), which assists in deriving the expression for $\btheta = \btheta_0$. Firstly, we observe that $\calQ_{1,1}$ in (\ref{deriv:2}) is 
\bel{A4-3}
 \bX_{t}\bW^\T \bW \bX_t ^\T &=& (\bZ_{t-1} + \tbE_{t})\bW^\T \bW (\bZ_{t-1} + \tbE_{t})^\T  \cr
 &=& \bZ_{t-1}\bW^\T \bW \bZ_{t-1} ^\T \!+\! \bZ_{t-1} \bW^\T \bW\tbE_{t}^\T \!+ \! \tbE_{t} \bW^\T \bW \bZ_{t-1}^\T +  \tbE_{t} \bW^\T \bW \tbE_{t}^\T\!.
\eel
Among equation (\ref{A4-3}), the last term can be rewritten as follows
\bel{A4-4}
\bbE \tbE_{t} \bW^\T \bW \tbE_{t}^\T &&= \bbE \tvec^{-1} \tvec\left( \tbE_{t} \bW^\T \bW \tbE_{t}^\T\right) \cr
&& = \bbE \tvec^{-1}\left( (\tbE_{t} \otimes \tbE_{t})\tvec(\bW^\T \bW)\right) \cr
&& =  \bbE \tvec^{-1}\left( \calR_2^{-1}(\bG_{0}^{-1}\be_{t} \be_{t}^\T (\bG_{0}^{-1})^\T) \tvec(\bW^\T \bW)\right) \cr
&& =   \sigma_{0}^2 \tvec^{-1}\left( \calR_2^{-1}(\bG_{0}^{-1} (\bG_{0}^{-1})^\T) \tvec(\bW^\T \bW)\right)  \cr 
&& =  \sigma_{0}^2 \tvec^{-1}\left( \calR_1((\bG_{0}^{-1})^\T\bG_{0}^{-1} )^\T \tvec(\bW^\T \bW)\right).
\eel
Secondly, the block matrix $\calQ_{1,2}$ in (\ref{deriv:2}) simplifies to 
\bel{A4-5}
&&\bbE (\tvec(\bX_{t-1}) \otimes \bI_{kn}) (\bW \bX_{t}^\T \otimes \bI_{k}) \cr 
&&= \bbE (\tvec(\bX_{t-1}) \otimes \bI_{kn})\left(\bW\bZ_{t-1}^\T \otimes  \bI_{k} + \bW\tbE_{t}^\T \otimes  \bI_{k}\right) \cr 
&& =  \bbE (\tvec(\bX_{t-1}) \otimes \bI_{kn})(\bW\bZ_{t-1}^\T \otimes  \bI_{k} ) +  \bbE (\tvec(\bX_{t-1}) \otimes \bI_{kn})(\bW\tbE_{t}^\T \otimes  \bI_{k})\cr
&& = \bbE (\tvec(\bX_{t-1}) \otimes \bI_{kn})(\bW\bZ_{t-1}^\T \otimes  \bI_{k} ).
\eel
The last equality holds because that every non-zero element in the second term involves the expected product of $\bX_{t-1}$ and $\bE_{t}$, which is assumed to  be  0 due to the independence  of $\bX_{t-1}$ and $\bE_{t}$. 

Thirdly,  the block matrix $\calQ_{1,3}$ in (\ref{deriv:2}) becomes
\bel{A4-6}
(\bX_{t}W^\T \otimes \bI_{k}) \bv_t &&=  (\bZ_{t-1}W^\T \otimes \bI_{k}) \bv_t +   (\tbE_{t}W^\T \otimes \bI_{k}) \bv_t.
\eel
Replacing $\bv_t$ with $\be_t$ in the last term, we arrive at 
\bel{A4-7}
\bbE (\tbE_{t}W^\T \otimes \bI_{k}) \tvec(\bE_{t}) 
&& = \bbE \tvec(\bE_{t} \bW \bE_{t}^\T)\cr
&&= \bbE (\tbE_{t} \otimes \bE_{t})\tvec(\bW) \cr
&& = \bbE \calR_2^{-1}(\bG_{0}^{-1}  \tvec(\bE_{t})   \tvec(\bE_{t}) ^\T)\tvec(W) \cr
&& =   \sigma_{0}^2 \calR_2^{-1}(\bG_{0}^{-1} )\tvec(\bW) \cr
&& = \sigma_{0}^2 \calR_1((\bG_{0}^{-1})^\T )^\T \tvec(\bW),
\eel
where the last equality holds by the relationship between $\calR_2$ and $\calR_1$ in (\ref{R12:2}).
In summary, the information matrix is $\Xi_{\btheta_{0},T} \coloneq - \bbE \left(\frac{1}{{T}} \frac{\partial^2 \calL_{T}(\btheta_{0})}{\partial  \btheta \partial \btheta^\T} \right) $. Specifically,
\begin{small}
\bel{A4-8}
&&\Xi_{\btheta_{0},T} = \nonumber \\
&& \begin{pmatrix}
	(\Xi_{\btheta_{0},T})_{11} ,&*, & *\\
	\frac{1}{\sigma_{0}^2 T}\sum\limits_{t = 1}^{T}\bbE (\tvec(\bX_{t-1}) \otimes \bI_{kn})(W\bZ_{t-1}^\T \otimes  \bI_{k} ), & \frac{1}{\sigma_{0}^2 T}\sum_{t = 1}^{T} \bbE \tvec(\bX_{t-1})\tvec(\bX_{t-1})^\T  \otimes \bI_{kn}  ,&\mathbf{0}_{k^2n^2 \times 1} \\
	\frac{1}{\sigma_{0}^2}\left( \calR_1(( \bG_{0}^{-1})^\T)^\T \tvec(\bW)  \right)^\T, &\mathbf{0}_{1\times k^2n^2},&\frac{kn}{2\sigma_{0}^4}
\end{pmatrix}, \nonumber \\ 
\eel
\end{small}
where
\begin{small}
\bel{A4-9}
(\Xi_{\btheta_{0},T})_{11}  =	\frac{1}{\sigma_{0}^2 T}\sum\limits_{t = 1}^{T}  \bbE \bZ_{t-1}\bW^\T \bW \bZ_{t-1} ^\T  \otimes \bI_{k} +
\bbE \tvec^{-1}\left( \calR_1( \bG_{0}^{-\T}\bG_{0}^{-1})^\T \tvec(\bW^\T \bW)\right) \otimes \bI_{k} + \sigma^2\tbW_{\tvec(\bC_{0})}. \nonumber \\
\eel
\end{small}
\subsection[The variance of the gradient at true value]{The variance of the gradient at $\mathbold{\theta}_0$}\label{proof:variance}
Recall that 
\begin{small}
\bel{deriv:3-cite1}
\frac{1}{\sqrt{T}} \frac{\partial \calL_{T}(\btheta_{0})}{\partial \btheta} = \begin{pmatrix}
	\frac{1}{\sigma_{0}^2}\frac{1}{\sqrt{T}}\sum_{t = 1}^{T} \left((\bZ_{t-1}\bW^\T \otimes \bI_{k}) \be_{t} +   (\tbE_{t}\bW^\T \otimes \bI_{k}) \be_{t} - \sigma_{0}^2 \calR_1(\bG_{0}^{-\T})^\T  \tvec(\bW)\right)\\
	\frac{1}{\sigma^2}\frac{1}{\sqrt{T}}\sum_{t = 1}^{T} (\tvec(\bX_{t-1}) \otimes \bI_{kn}) \be_{t} \\	\frac{1}{2\sigma^4}\frac{1}{\sqrt{T}}\sum_{t = 1}^{T}  \left(\be_{t}^\T\be_{t} - kn \sigma^2\right)
\end{pmatrix}. \nonumber \\
\eel
\end{small}
\noindent Define the variance  of the gradient at $\btheta_0$ as $\Psi_{\btheta_{0},T}$.
We first simplify the second term of the $(1,1)$-th block matrix of $\Psi_{\btheta_{0},T}$ as follows:
	\bel{psi-1}
	\bbE (\tbE_t W^\T \otimes \bI_{k})\be_{t}\be_{t}^\T && =  \bbE \tvec(\bE_{t} W \bE_{t}^\T)\be_{t}^\T\cr
	&&= \bbE (\tbE_{t} \otimes \bE_{t})\tvec(W)\be_{t}^\T \cr
	&& = \bbE \calR_2^{-1}(\bG_{0}^{-1}  \tvec(\bE_{t})   \tvec(\bE_{t}) ^\T)\tvec(W)\be_{t}^\T \cr
	&& = \bbE \calR_1(  \tvec(\bE_{t})   \tvec(\bE_{t}) ^\T \bG_{0}^{-\T})^\T\tvec(W)\be_{t}^\T \cr
	&& \overset{\Delta}{=} \mu_3 \bD_1,
	\eel
	where $\bD_1 \in \bbR^{k^2 \times kn}$ can be viewed as a collection of $k\times n$ arrays, each consisting of blocks of  size $k \times k$. For each $1 \leq i \leq k, 1 \leq j \leq n$, the $(i,j)$-th block of $\bD_1$ is diagonal, i.e., all off-diagonal entries are 0. That is 
	\bes
	[\bD_1]_{[i,j]} = \begin{pmatrix}
		\sum_{\tilde{j}= 1}^{n} W_{j,\tilde{j}} \left((\bG_{0}^{-\T})_{[j,\tilde{j}]}\right)_{1,i}	,&0,& \ldots, & 0\\
		0, & 	\sum_{\tilde{j}= 1}^{n} W_{j,\tilde{j}} \left((\bG_{0}^{-\T})_{[j,\tilde{j}]}\right)_{1,i}	 ,&\ldots,&0\\
		\vdots&\vdots&\ddots& \vdots\\
		0, &0,& \ldots,&\sum_{\tilde{j}= 1}^{n} W_{j,\tilde{j}} \left((\bG_{0}^{-\T})_{[j,\tilde{j}]}\right)_{k,i}
	\end{pmatrix}.
	\ees
Hence,  by algebra, we arrive at
\bel{psi:1-1}
(\Psi_{\btheta_{0},T})_{11}  &&=  \frac{1}{\sigma_{0}^2 T}\sum\limits_{t = 1}^{T}  \bbE (\bZ_{t-1}\bW^\T \bW \bZ_{t-1} ^\T  )\otimes \bI_{k} 
-   \calR_1(\bG_{0}^{-\T})^\T \tvec(\bW) \tvec(\bW)^\T \calR_1(\bG_{0}^{-\T})  \cr
&&  \quad + \frac{1}{\sigma_{0}^4 T}\sum\limits_{t = 1}^{T}  {\bbE \calR_1( \be_t  \be_t ^\T \bG_{0}^{-\T})^\T \tvec(\bW) \tvec(\bW)^\T \calR_1(\be_t  \be_t ^\T \bG_{0}^{-\T} )}  \cr
&& \quad + \frac{\mu_3}{\sigma_{0}^4}\frac{1}{T}\sum_{t = 1}^{T}  \bD_1 \left(\bW\bZ_{t-1}^\T \otimes \bI_{k}\right) +  \frac{\mu_3}{\sigma_{0}^4}\frac{1}{T}\sum_{t = 1}^{T} \left(\bZ_{t-1} \bW^\T \otimes \bI_{k}\right) \bD_1^\T,
\eel
where $\mu_3$ being the third moment of $\tvec(\bE_t)_{ij}$ for $1\leq i\leq n, 1\leq j\leq n$ and $1 \leq t \leq T$. 
Next, we examine the $(2,1)$-th block matrix of $\Psi_{\btheta_{0},T}$ as
\bel{psi:2-1}
&& (\Psi_{\btheta_0,T })_{21} \cr
&& = 	\frac{1}{\sigma_{0}^2 T}\sum\limits_{t = 1}^{T}\bbE (\tvec(\bX_{t-1}) \otimes \bI_{kn})(\bW\bZ_{t-1}^\T \otimes  \bI_{k} )  + \frac{1}{\sigma_{0}^4}\frac{1}{T}\sum_{t = 1}^{T} (\tbE_{t}\bW^\T \otimes \bI_{k}) \be_{t}  \be_{t}^\T  (\tvec(\bX_{t-1})^\T \otimes \bI_{kn}) \cr
&&= \frac{1}{\sigma_{0}^2 T}\sum\limits_{t = 1}^{T}\bbE (\tvec(\bX_{t-1}) \otimes \bI_{kn})(\bW\bZ_{t-1}^\T \otimes  \bI_{k} ) + 
\frac{\mu_3}{\sigma_{0}^4}\frac{1}{T}\sum_{t = 1}^{T} \bD_1 (\tvec(\bX_{t-1})^\T \otimes \bI_{kn}). 
\eel
Then, the $(3,1)$-th block of $(\Psi_{\btheta_{0}},T)$ is given  by 
\bel{psi:3-1_v1}
(\Psi_{\btheta_{0}},T)_{31} &&= {\frac{\mu_3}{2\sigma_{0}^6}\frac{1}{T}\sum_{t=1}^{T} (\bZ_{t-1}W^\T \otimes \bI_{k})\mathbf{1}_{kn}} 
+  	\frac{1}{2\sigma_{0}^6 T}\sum_{t = 1}^{T}  \bbE \be_{t}^\T \be_{t} \be_{t}^\T(\bW \tbE_{t}^\T \otimes \bI_{k}) \cr
&& \quad 	-\frac{kn}{2 \sigma_{0}^4 T}\sum_{t = 1}^{T}  \bbE\left(  (\tbE_{t}\bW^\T \otimes \bI_{k}) \be_t  \right) ^ \T ,
\eel
where $\mathbf{1}_{kn} \in \bbR^{kn}$ is a column vector with each entry is equal to 1. 
Among that, the second term of (\ref{psi:3-1_v1}) is given by 
\bel{psi:eq-1}
\bbE \be_{t}^\T \be_{t} \be_{t}^\T(W \tbE_{t}^\T \otimes \bI_{k}) &&= \bbE \left( ( \tbE_{t} W^\T \otimes \bI_{k}) \be_{t} \be_{t}^\T \be_{t} \right)^\T \cr
&& = \bbE \left( \tvec(\bE_{t}W\tbE_{t}^\T) \be_{t}^\T \be_{t} \right)^\T \cr
&& = \bbE \left( (\tbE_{t}\otimes \bE_{t}) \tvec(W) \be_{t}^\T \be_{t} \right)^\T \cr
&& = \bbE  \left( \calR_2^{-1}(\be_{t}^\T \be_{t} \bG_{0}^{-1}\be_{t}\be_{t}^\T) \tvec(W)  \right)^\T \cr
&& = ( kn+2 )\sigma_{0}^4 \left( \calR_2^{-1}( \bG_{0}^{-1}) \tvec(W)  \right)^\T \cr 
&& = ( \mu_4  + kn-1 )\sigma_{0}^4 \left( \calR_2^{-1}( \bG_{0}^{-1}) \tvec(\bW)  \right)^\T ,
\eel
where $\mu_4$ being the fourth moment of $\tvec(\bE_t)_{ij}$ for $1\leq i\leq n, 1\leq j\leq n$ and $1 \leq t \leq T$. Thus,
\bel{psi:3-1}
(\Psi_{\btheta_{0}},T)_{31} =  {\frac{\mu_3}{2\sigma_{0}^6}\frac{1}{T}\sum_{t=1}^{T} (\bZ_{t-1}\bW^\T \otimes \bI_{k})\mathbf{1}_{kn}}  + \frac{\mu_4 - \sigma_{0}^4}{\sigma_{0}^6}\left( \calR_2^{-1}( \bG_{0}^{-1})  \tvec(\bW)  \right)^\T. 
\eel
In summary, the variance of gradient at $\btheta_0$ is provided as following 
\bel{psi}
 \Psi_{\btheta_{0},T}  &&\coloneq \bbE \left(\frac{1}{\sqrt{T}} \frac{\partial \calL_{T}(\btheta_{0})}{\partial \btheta}  \cdot \frac{1}{\sqrt{T}} \frac{\partial \calL_{T}(\btheta_{0})}{\partial \btheta^\T} \right)\cr
&&= \begin{pmatrix}
	(\Psi_{\btheta_{0},T})_{11}	,&*, & *\\
	(\Psi_{\btheta_{0},T })_{21}, &  \frac{1}{\sigma_{0}^2 T}\sum_{t = 1}^{T} \bbE \tvec(\bX_{t-1})\tvec(\bX_{t-1})^\T  \otimes \bI_{kn} ,&* \\
	(\Psi_{\btheta_{0},T})_{31}, & {\frac{\mu_3}{2\sigma_{0}^6}\frac{1}{T}\sum_{t=2}^{T}\mathbf{1}_{kn}^\T(\tvec(\bX_{t-1})^\T \otimes \bI_{kn})  } ,&\frac{kn}{2\sigma_{0}^4} + \frac{\mu_4 - 3\sigma^ 4}{4\sigma_{0}^8}
\end{pmatrix} ,
\eel
where $(\Psi_{\btheta_{0},T})_{11}, (\Psi_{\btheta_{0},T})_{21}$ and $(\Psi_{\btheta_{0},T})_{31}$ are given in (\ref{psi:1-1}), (\ref{psi:2-1}) and (\ref{psi:3-1}), respectively.
	
\subsection{Some lemmas}
Recall that a  matrix  $\bM$ is considered bounded if both $\|\bM\|_1 < \infty$ and $\|\bM\|_{\infty} < \infty$.
\begin{lemma}\label{lemma:LLN}
	Under Assumption \ref{assum:3}-\ref{assum:4}, for any bounded matrix $\bM$ with compatible dimensions, we have
\bel{789}
	\frac{1}{T}\sum_{t = 1}^{T}\tvec(\bX_{t-1})^\T \bM \tvec(\bX_{t-1}) - 	\bbE \frac{1}{T}\sum_{t = 1}^{T}  \tvec(\bX_{t-1})^\T \bM \tvec(\bX_{t-1})  &=&  O_p(\frac{1}{\sqrt{T}}). \\
    \frac{1}{T}\sum_{t = 1}^{T}\tvec(\bX_{t-1})^\T \bM \tvec(\bE_{t}) - 	\bbE \frac{1}{T}\sum_{t = 1}^{T}  \tvec(\bX_{t-1})^\T \bM \tvec(\bE_{t})  &=& O_p(\frac{1}{\sqrt{T}}). \\
    \frac{1}{T}\sum_{t = 1}^{T}\tvec(\bE_{t})^\T \bM \tvec(\bE_{t}) - 	\bbE \frac{1}{T}\sum_{t = 1}^{T}  \tvec(\bE_{t})^\T \bM \tvec(\bE_{t})  &=& O_p(\frac{1}{\sqrt{T}}). 
\eel
where $\bbE \tvec(\bX_{t-1})^\T \bM \tvec(\bX_{t-1}) = O_p(1)$, $\bbE \tvec(\bX_{t-1})^\T \bM \tvec(\bE_{t})  = O_p(\frac{1}{T})$ and $\bbE \tvec(\bE_{t})^\T \bM \tvec(\bE_{t}) = O_p(1)$.
\end{lemma}
\begin{lemma}\label{lemma:6}
	For $\btheta$ lying in  the neighborhood of $\btheta_{0}$,  we have
	\bel{weaklemma:1}
	\left(	-\frac{1}{{T}} \frac{\partial^2 \ln \calL_{T}({\btheta})}{\partial {\btheta} \partial \btheta^\T} - \left(	-\frac{1}{{T}} \frac{\partial^2 \ln \calL_{T}({\btheta}_{0})}{\partial {\btheta} \partial \btheta^\T} \right)\right) = \|\btheta - \btheta_{0}\|_2 \cdot O_{p}(1).
	\eel
\end{lemma}

\begin{proof}
	This detailed difference of $\left(	-\frac{1}{{T}} \frac{\partial^2 \ln \calL_{T}({\btheta})}{\partial \btheta \partial {\btheta}^\T} - \left(	-\frac{1}{{T}} \frac{\partial^2 \ln \calL_{T}({\btheta}_{0})}{\partial {\btheta} \partial {\btheta}^\T} \right)\right)$ can be obtained by investigating the (\ref{deriv:2}). Then, based on equation (\ref{deriv:2}), we only need to examine three block matrices: $(1,1)$-th, $(3,1)$-th and $(3,3)$-th. 
	
	Firstly,when calculating  $
    \tbW_{\tvec(\bC)} - \tbW_{\tvec(\bC_{0})}$ of $(1,1)$-th block matrix, for fixed $i,j, \tilde{i},\tilde{j}$, we have 
	\bel{pf-lem-weak-eq-1}
	&& (\bG^{-1})_{[j,\tilde{j}]} \otimes [(\bG^{-1})_{[\tilde{i},i]}]^\T  - (\bG_{0}^{-1})_{[j,\tilde{j}]} \otimes [(\bG_{0}^{-1})_{[\tilde{i},i]}]^\T  \cr 
	&&=   (\bG^{-1})_{[j,\tilde{j}]} \otimes [(\bG^{-1} - \bG_{0}^{-1})_{[\tilde{i},i]} ]^\T + (\bG^{-1} - \bG_{0}^{-1})_{[j,\tilde{j}]}  \otimes [(\bG_{0}^{-1})_{[\tilde{i},i]}]^\T.
	\eel
	On the other hand, we note that 
	\bel{pf-lem-weak-eq-2}
	\bG^{-1} - \bG_{0}^{-1} = \bG_0^{-1}\left(W\otimes(\bC - \bC_{0})\right) \bG_{0}^{-1}.
	\eel
	Hence, as $\bW, \bG^{-1}$ and $\bG_{0}^{-1}$ are uniformly bounded, the expression (\ref{pf-lem-weak-eq-1})  has the same order of $\|\bC - \bC_{0}\|_F$. Therefore, the difference of $(1,1)$-th block matrix is of the order $\|\btheta - \btheta_{0}\|_2\cdot O(1)$.
	
	Secondly,  recall that $\bv_t = \bG \tvec(\bX_t) - \bPhi \tvec(\bX_{t-1})$ and $\be_{t} = \bG_{0} \tvec(\bX_t) - \bPhi_{0} \tvec(\bX_{t-1})$, we have 
	\bel{eq-34}
	\bv_t - \be_{t}  =  (\bG - \bG_{0})\tvec(\bX_t) - (\bPhi - \bPhi_{0})\tvec(\bX_{t-1}) .
	\eel
	Then for the difference of $(3,1)$-th block matrix, 
	\bel{eq-35}
	&&\frac{1}{T}\sum_{t = 1}^{T} (\bX_{t}W^\T \otimes \bI_{k}) \bV_t   - 	\frac{1}{T}\sum_{t = 1}^{T} (\bX_{t}W^\T \otimes \bI_{k}) \be_{t} \cr
	&&= \frac{1}{T}\sum_{t = 1}^{T} (\bX_{t}W^\T \otimes \bI_{k})(\bG - \bG_{0})\tvec(\bX_t) \!-\! \frac{1}{T}\sum_{t = 1}^{T} (\bX_{t}W^\T \otimes \bI_{k})(\bPhi - \bPhi_{0})\tvec(\bX_{t-1}).
	\eel
	Similarly, for $(3,3)$-th block matrix , 
	$
	\bv_t^\T \bv_t  -  \be_{t}^\T \be_{t} =  ( \bv_t +  \be_{t})^\T( \bv_t -  \be_{t})
	$. 
	Therefore, according to  Lemma \ref{lemma:LLN}, the difference is of the same order as $\|\btheta - \btheta_{0}\|_2$, multiplied by stochastic terms with order $O_p(1)$. Thus, we arrive at
	\bes
		-\frac{1}{{T}} \frac{\partial^2 \ln \calL_{T}({\btheta})}{\partial {\btheta} \partial \btheta^\T} - \left(	-\frac{1}{{T}} \frac{\partial^2 \ln \calL_{T}({\btheta}_{0})}{\partial {\btheta} \partial \btheta^\T} \right) = \|\btheta - \btheta_{0}\|_2 \cdot O_{p}(1).
	\ees
	
\end{proof}
\begin{lemma}\label{lemma:7}
	\bel{weaklemma:2}
	\left(	-\frac{1}{{T}} \frac{\partial^2 \ln \calL_{T}({\btheta}_{0})}{\partial {\btheta} \partial \btheta ^\T} \right) - \Xi_{\btheta_{0},T} = O_{p}(\frac{1}{T}).
	\eel
\end{lemma}
\begin{proof}
	Since $\Xi_{\btheta,T}$ defined as $- \bbE \left(\frac{1}{{T}} \frac{\partial^2 \calL_{T}(\btheta_{0})}{\partial \btheta \partial \btheta^\T} \right) $, all the entries of the difference $	\left(	-\frac{1}{{T}} \frac{\partial^2 \ln \calL_{T}({\btheta}_{0})}{\partial {\btheta} \partial \btheta ^\T} \right) - \Xi_{\btheta_{0},T} $ have zero means. The detailed procedures are provided in section \ref{proof:deriv2}. Moreover, according to lemma \ref{lemma:LLN}, all entries of the difference converge to zero at the rate of $O_p(\frac{1}{\sqrt{T}})$.
\end{proof}

\bibliography{sigmarRef}
\bibliographystyle{ims}